%% file: jfrpaper.tex
\documentclass[10pt,twocolumn,a4paper]{article} 
\usepackage{graphicx}
\usepackage{apalike}
\usepackage{setspace}

\usepackage{color,soul}

\usepackage{subfig}

\usepackage{textgreek}

\usepackage{upgreek}

\usepackage[]{algorithm,algcompatible,amsmath}

\usepackage{amssymb}

\usepackage{bm}

\usepackage{enumitem}

\usepackage{authblk}

\algnewcommand\INPUT{\item[\textbf{Input:}]}%
\algnewcommand\OUTPUT{\item[\textbf{Output:}]}%

\title{Using an Automated Heterogeneous Robotic System for Radiation Surveys}

\author[1]{Petr Gabrlik}
\author[1]{Tomas Lazna}
\author[1]{Tomas Jilek}
\author[2]{Petr Sladek}
\author[1]{Ludek Zalud}
\affil[1]{Central European Institute of Technology, Brno University of Technology, Brno, Czech Republic\\ 
	\textit{\{petr.gabrlik,tomas.lazna,tomas.jilek,ludek.zalud\}@ceitec.vutbr.cz}}
\affil[2]{\textit{sladek256@gmail.com}}

\providecommand{\keywords}[1]{\textbf{\textit{Keywords:}} #1}

\begin{document}

\maketitle

\begin{abstract}
During missions involving radiation exposure, unmanned robotic platforms may embody a valuable tool, especially thanks to their capability of replacing human operators in certain tasks to eliminate the health risks associated with such an environment. Moreover, rapid development of the technology allows us to increase the automation rate, making the human operator generally less important within the entire process. This article presents a multi-robotic system designed for highly automated radiation mapping and source localization. Our approach includes a three-phase procedure comprising sequential deployment of two diverse platforms, namely, an unmanned aircraft system (UAS) and an unmanned ground vehicle (UGV), to perform aerial photogrammetry, aerial radiation mapping, and terrestrial radiation mapping. The central idea is to produce a sparse dose rate map of the entire study site via the UAS and, subsequently, to perform detailed UGV-based mapping in limited radiation-contaminated regions. To accomplish these tasks, we designed numerous methods and data processing algorithms to facilitate, for example, digital elevation model (DEM)-based terrain following for the UAS, automatic selection of the regions of interest, obstacle map-based UGV trajectory planning, and source localization. The overall usability of the multi-robotic system was demonstrated by means of a one-day, authentic experiment, namely, a fictitious car accident including the loss of several radiation sources. The ability of the system to localize radiation hotspots and individual sources has been verified.
\end{abstract}



\section{Introduction}

\input{text/introduction}

\section{Previous Work}
\label{sec:previouswork}

\input{text/previous-work}

\section{Methods}

\input{text/methods-experiment}

\input{text/methods-overview}

\input{text/methods-uas}

\input{text/methods-ugv}

\input{text/methods-photogrammetry}

\input{text/methods-radiation-uas}

\input{text/methods-polygon}

\input{text/methods-trajectory-ugv}

\input{text/methods-sources-localization}

\section{Results}

\input{text/results-photogrammetry}

\input{text/results-radiation-uas}

\input{text/results-polygon}

\input{text/results-trajectory-ugv}

\input{text/results-radiation-ugv}


\section{Discussion}
\label{sec:discussion}

\input{text/discussion}

\section{Conclusion}

\input{text/conclusion}

\subsubsection*{Acknowledgments}
The research was supported by the European Regional Development Fund under the project Robotics 4 Industry 4.0 (reg. no. CZ.02.1.01/0.0/0.0/15\_003/0000470). Further funding was provided via the National Center of Competence 1 program of the Technology Agency of the Czech Republic, under the project TN01000024/15 "Robotic operations in a hazardous environment and intelligent maintenance".

We thank NUVIA, a.s. for the instrumentation support and the Air Force and Air Defence Military Technical Institute for the cooperation and provision of the UAS. We also acknowledge the consultation and assistance ensured by the University of Defence in Brno and the Fire Rescue Service of the South Moravian Region.

\bibliographystyle{apalike}
\bibliography{ref_petr_vlastni,ref_petr_radiace,ref_petr_disertace,ref_petr_photo,ref_petr_skupina_robotiky,ref_toml,ref_tomj}

\end{document}

%% file: text/introduction.tex
Any radiation mapping, namely, measurement that provides knowledge of the distribution of ionizing radiation in space and time, finds use in various applications related to common activities. In this context, we can mention, for example, geophysical surveys, environmental monitoring of nuclear sites, post-disaster responses, localization of lost radiation sources, and everyday operation of nuclear power plants (NPP). Advantageously, such tasks are often carried out by utilizing unmanned robotic systems, mainly to protect human health. Robots are also capable of reducing the time and increasing the accuracy thanks to semi- or fully autonomous operation. Unmanned systems were employed in resolving the most severe nuclear accidents, including the Chernobyl disaster in 1986 and the Fukuschima Dai-ichi NPP accident in 2011, when personnel safety was of major importance. To improve the efficiency, different assets and techniques may be combined. Thus, for example, one of the oldest and most commonly applied radiation survey methods is helicopter-based airborne spectrometry. The technique is able to quickly cover square kilometres of land, but only at the expense of inadequate accuracy and very high cost. Ground systems, by comparison, may ensure superb accuracy, but their operational ranges are mostly limited to several hundreds of square meters, albeit these are explorable in a reasonably short period. Unmanned aircraft systems (UAS) offer adequate accuracy and survey range; therefore, to recognize the radiological situation in medium-sized areas, a multi-robot system seems to be a promising option.
  
\subsection{Robot Deployment}
The necessity to employ remotely operated machines in radiation exposure environments appeared with the expansion of nuclear power plants during the second half of the 20th century. Such machines were mostly used to perform inspection, manipulation, and maintenance; however, nuclear accidents shifted the interest towards the development of terrestrial mobile robots intended for disaster response applications \cite{tsitsimpelis_review_2019}. These systems are principally applicable in reconnaissance, data gathering, and object manipulation; due to the complexity of the environment, remote control is generally employed as the most convenient approach. The main challenges and tasks characterizing the robot development stage involve radiation resistance and decontamination, environment traversability, and human-robot interaction \cite{qian_small_2012,guzman_rescuer_2016,ducros_rica_2017}. A teleoperated robot was successfully utilized, for example, to inspect the damage after the Fukushima Daiichi Nuclear Power Plant accident in 2011 \cite{nagatani_emergency_2013}.

The deployment of robots with autonomous functions in post-disaster environments, especially inside or close to collapsed buildings, remains a major challenge; however, various other applications comprising radiation exposure are available. Ground robots enabling autonomous or semi-autonomous operation can be employed in radioactive waste storage facilities; areas affected by radiation as a result of an accident; uranium mines; and to localize uncontrolled radiation sources. \cite{schneider_autonomous_2011} present a six-wheeled unmanned ground vehicle (UGV) specially designed for chemical, biological, radiological, nuclear, and explosive-related (CBRNE) tasks to solve some local navigation problems automatically. Laser scanner-based obstacle detection, together with a local navigation planning algorithm, is employed to find a collision-free path to the desired global positioning system (GPS) coordinate. Autonomous radiation mapping inside pre-defined polygons was discussed by \cite{hosmar_experimental_2017}. In this case, precise navigation is enabled thanks to a real-time kinematics (RTK) GPS receiver. The data collected by a four-wheeled robot, equipped with an Na(I) radiation detector, are further utilized for particle swarm optimization-based source localization. The presented solution is, however, suitable for obstacle-free and non-complex areas only. The same UGV platform was deployed in a nuclear storage facility to perform inspections \cite{wang_autonomous_2018}. In such a GPS-denied environment, localization represents the essential task. Within the research, a light detection and ranging (LiDAR)-based simultaneous localization and mapping (SLAM) algorithm facilitates navigation inside an unknown territory, while the data from an RGB-depth (RGB-D) sensor allow the localization of storage cylinders. The discussed problems principally concern localization, navigation, and mapping, and they thus pose fundamental challenges to general terrestrial mobile robotics, including radiation-related tasks.

The rapid growth of UASs applications in the last decade has manifested itself also within the discussed area. Compared to ground robots, there is an evident advantage: the operational environment can be considered obstacle-free and terrain shape-independent while flying outside, at a safe distance from the ground. The fact enables quick radiation data collection over a large area. A UAS as a means to assist in solving nuclear emergency cases was proposed already in 2008, when a 100~kg unmanned helicopter equipped with an 8~kg scintillating detector was employed to estimate dose-rate distribution automatically \cite{okuyama_remote_2008}. A similarly sized unmanned system proved to be beneficial after the Fukushima Daiichi accident, where it provided information  about radioactive cesium deposition inside a 5~km radius around the site \cite{sanada_aerial_2015,sanada_temporal_2016}. During the event, a detailed radiation map was compiled for the first time, considering the legal restrictions placed on manned aircraft operation in the area. Unlike ground robots, UASs operate at relevant distances from the source, typically situated at the ground level, and thus they require a sensitive radiation detection system, which embodies considerable payload. For this reason, micro-unmanned vehicles, a category popular thanks to its flexibility, low price, and safe operation, must operate as close to the ground as possible to collect radiation data even with less sensitive detectors \cite{martin_3d_2016,macfarlane_lightweight_2014}. Moreover, flying robots are applicable in producing digital elevation models (DEM) thanks to the LiDAR or photogrammetry techniques. The latter method was used by, for example, \cite{connor_application_2018}, resulting in a 3D surface model covered with a radiation data layer. \cite{vetter_advances_2019} present a complex, multi-sensor system for both UASs and UGVs, which integrates various sensors and approaches to present radiation data in 3D and real time. The drawback of low-altitude mapping rests in potential collisions with obstacles, a problem discussed within \cite{martin_use_2015}, where the flight height during legacy uranium mines mapping was manually adjusted according to the vegetation height.

In most cases, contrary to terrestrial robots, aerial platforms are enabled by the obstacle-free environment to operate in an automatic manner, utilizing predefined GPS waypoints. However, state-of-the-art technologies have opened up the opportunity to deploy such platforms even indoor, in GPS-denied environments \cite{mascarich_radiation_2018}, and novel, lightweight radiation sensors may allow the use of even smaller UASs, possibly operated in swarms \cite{baca_timepix_2019}.

The advantages of both ground and aerial robots may be combined within a multi-robotic radiation mapping system. Such an idea was introduced by \cite{kochersberger_post-disaster_2014}, whose unconventional solution comprises an unmanned helicopter carrying a small UGV. In this case, the UAS is intended to localize potential radiation-contaminated area via an onboard NaI detector and to produce a surface model by exploiting a stereovision system. A UGV, by contrast, is deployed with a winch system, facilitating comprehensive ground inspection and sample collection. The multi-robot system, however, was not subjected to thorough testing; thus, the vehicles' practical capabilities have not been confirmed sufficiently.

A promising approach rests in areal UAS-based mapping to facilitate gathering sparse radiation data within a large area and to produce a relevant map or a 3D model simultaneously, within a short time; these products then allow a UGV to be deployed at certain locations to perform detailed measurements, reconnaissance, and source localization. A similar method was introduced and verified by \cite{christie_radiation_2017}, whose aerial platform yielded a georeferenced orthophoto and a DEM, while also performing measurements with an onboard scintillation detector. A ground robot was then automatically navigated to locations exhibiting a maximal counts per second (CPS) value, and a classified map based on the orthophoto as well as the DEM enabled the choice of an energy-effective path; real-time obstacle avoidance was ensured by a LiDAR. The experiment verified the system's ability to localize an unknown source; however, the simple localization technique detects one maximum only, thus being unsuitable for multi-source or areal contamination scenarios. A promising concept to exploit different robotic platforms is described within the study \cite{peterson_experiments_2019}, where the key idea rests in using an aerial imagery-based DEM to divide the study site into sub-areas according to their suitability for individual robots. Ground radiation measurements are carried out in UGV-passable regions only; a UAS is employed in the rest of the target zone. Moreover, various algorithms exploiting radiation spectra are tested to find the sources, even in the real-time mode. Despite the advantages, the system has not yet been fully prepared to operate in real-world conditions without operator intervention.

\subsection{Source Localization Methods}
One of the common tasks addressed in the literature is the localization of radionuclides, which consists in identifying the parameters of the point sources present in the studied region of interest. The methods usually work with a series of discrete measurements that are assumed to have been taken at known positions; these measurements are performed by either a robotic platform or static sensors. In many cases, the methods are verified only by simulation.

The paper \cite{bai_maximum_2014} utilizes maximum likelihood estimation (MLE) to find a single source, reducing the problem to two dimensions to acquire a coarse estimate that is improved by using a gradient method. In \cite{lin_searching_2015}, the artificial potential field approach is adopted to localize one source, too; the attractive force is derived from the source's position estimated via the particle filter (PF) technique, while the repulsive one allows the robot to avoid obstacles. Another example of PF application can be found in \cite{chin_efficient_2011}; the advantage of the interpretation proposed within the article consists in that it is not necessary to know the number of sources a priori. The algorithm works with a network of detectors measuring at multiple places simultaneously and is thus unsuitable for single-robot scenarios. An array of directional sensitive detectors can be employed for tracking a moving source as well \cite{liu_wireless_2014}. Fast hotspot localization is characterized by \cite{newaz_fast_2016}, where the proposed algorithm dynamically adopts the UAS trajectory to move towards the hotspot. Two path planners, namely, HexTree and RIG-tree, are compared to demonstrate that the latter converges to the goal more quickly. Localization methods utilizing a UAS to collect data are examined in paper \cite{towler_radiation_2012}. The authors introduce an algorithm based on recursive Bayesian estimation, showing the disadvantages: The selected approach locates merely one source, requiring knowledge of the nuclide and its activity a priori. Then, a method exploiting the radiation contour is outlined; the related analysis managed via the Hough transform is able to find multiple sources, whose contours may overlap. Surveying the region of interest with more UGVs enabling us to localize multiple sources is covered in \cite{pinkam_exploration_2016}; the presented strategy prefers short paths having higher radiation intensity gradients. The parameter estimation utilizes the PF method with disperse resampling (to prevent particle degeneration). The radiation field model includes previously found sources. The studied area is divided into cells to be visited, with a focus on those that potentially contain a source. A drawback of the method consists in ignoring potential obstacles.

Over the last decade, the localization algorithms have been studied by B. Ristic and his research group, who partially verified the methods by using real data acquired during a field test. The paper \cite{gunatilaka_localisation_2007} compares three approaches to single source localization; the techniques are based on the MLE, the extended Kalman filter (EKF), and the unscented Kalman filter (UKF). The authors also analyzed the theoretical minimum estimation error with a Cramér-Rao bound, indicating that sequential Bayesian estimators (the EKF and the UKF) provide better performance than the MLE. The Bayesian approach is utilizable for a scenario including static detectors and a moving source. The models are fitted via partial Bayes factors, whose values are approximated by the Monte Carlo method denoted as importance sampling with progressive correction. The algorithm is capable of tracking up to four sources; however, their number needs to be known a priori \cite{morelande_radiological_2009}. The radiation field can be modeled as a weighted sum of 2D Gaussians, or a Gaussian mixture \cite{morelande_radiation_2009}. To find the Gaussians' parameters, two estimators, namely, a Gaussian and a Monte Carlo approximation, are employed, with the former yielding better results in both the simulations and the real data application. The algorithm is rather robust, and exact a priori knowledge of the number of sources is not required. In \cite{mendis_experimental_2009}, up to three sources are localized, with binary and continuous genetic algorithms constituting alternative implementations of the MLE algorithm and negative-log likelihood being the objective function. The number of sources present in the area can be found by applying the minimum description length (MDL) principle. This method is based on minimizing the function that takes the parameter matrix as the input; this matrix needs to be estimated for every considered number of sources \cite{gunatilaka_experimental_2010}. An information-driven search altering the measurement trajectory during the data acquisition process is outlined in the article \cite{ristic_information_2010}. The number of sources is assumed to be unknown; the source are tracked by one or more mobile observers, and their parameters are estimated via a multi-target, track-before-detect particle filter. The particles are initialized with different amounts of sources; at the end of each update step, some of the particles acquire a source while some others lose it. Particle degeneration is prevented via the progressive correction technique. The observer motion control is addressed as a partially observable Markov decision process; a control vector to maximize the estimation of the reward function is selected. The simulations have shown that the information-driven search yields results more accurate than those obtained from the survey along a pre-defined uniform trajectory. The method was also verified by using field data.

\subsection{Possible Applications}
Potential missions for multi-robot systems involve several applications that beneficially combine quick, flexible operation and a large range of aerial assets with the versatility and better radiation measurement conditions ensured by terrestrial robots. A combination of UASs and UGVs provides a synergy of benefits for radiological mapping, bringing both global information from the territory and accurate dosimetry or spectroscopy data from the points of interest. Nuclear safety, radiation and environmental protection, remediation, and decommissioning then embody some of the target fields. Regarding UGVs, a major advantage rests in the possibility of applying semiconductor high-purity germanium (HPGe) detectors with high resolution (radionuclide identification) capabilities; in UASs, conversely, the resolution is still limited by vibrations and the microphonic effect \cite{zimmermann_active_2013}, and the onboard heavy sensitive detectors restrain the operation time. In this context, a UGV is significantly more flexible and can carry diverse detection systems, including continuously working dose rate meters with high dynamic range coverages, accurate solid-state spectrometers, neutron detectors and beta contamination meters for occasional static measurements, and alpha contamination indicators. The devices mounted on a UGV may support the monitoring with measurements at a height of 1 m above the ground, which corresponds to the dosimetry standard for radiological mapping. In general terms, robotic systems functionally extend the set of regular nuclear monitoring options, where static station networks embody the most common approach, despite providing the evidence at several discrete points only. The area coverage could be improved by autonomous inspection tasks to enhance the applicability of the existing scenarios.

Comprehensive radiation surveys necessarily involve detailed, laboratory-based analyses of the samples, and the use of UASs/UGVs can improve the applied sampling strategy. Moreover, a ground robot is capable of assisting in remote sample collection if equipped appropriately. Importantly, UASs and UGVs also contribute to the remediation of legacy sites, which embodies a topical issue concerning diverse inactive locations that used to host nuclear technologies. In some cases, small, heavy-duty wireless monitoring stations are distributed across the investigated sites to provide data in a long-term horizon; this task can be also performed by ground-based assets at suitable points indicated via relevant aerial measurement. Finally, the localization of the lost or uncontrolled radiation sources discussed within this article constitutes a use case for robotic systems.

\subsection{Aims and Objectives} 
The paper aims to present the options and perspectives of using a multi-robot system to perform highly automated radiation mapping and source localization in an outdoor environment. The illustrative experiment, a car accident involving the loss of several sources, was designed to resemble a real-world scenario as much as possible. Moreover, to increase the authenticity, the entire field work comprised several UAS flights, UGV deployments, and data processing cycles within a single day.

The initial part of the article, namely, the 'Previous work' section, briefly discusses our long-term research activities within this domain. In the next chapter, 'Methods', we provide an overview of the proposed mapping process and experiment setup, followed by a thorough description of the applied equipment and designed algorithms. The acquired data and processing outputs of the radiation mapping and source localization tasks are presented within the 'Results' section, in a chronological order corresponding to reality. Finally, the 'Discussion' chapter compares the achieved results with both our originally planned targets and the outcomes outlined in the referenced literature. As this paper constitutes a part of a comprehensive research concept, we also address tasks to be potentially solved in the future.

%% file: text/previous-work.tex
\begin{figure}[!tbp]
	\centering
	\includegraphics[width=0.48\textwidth]{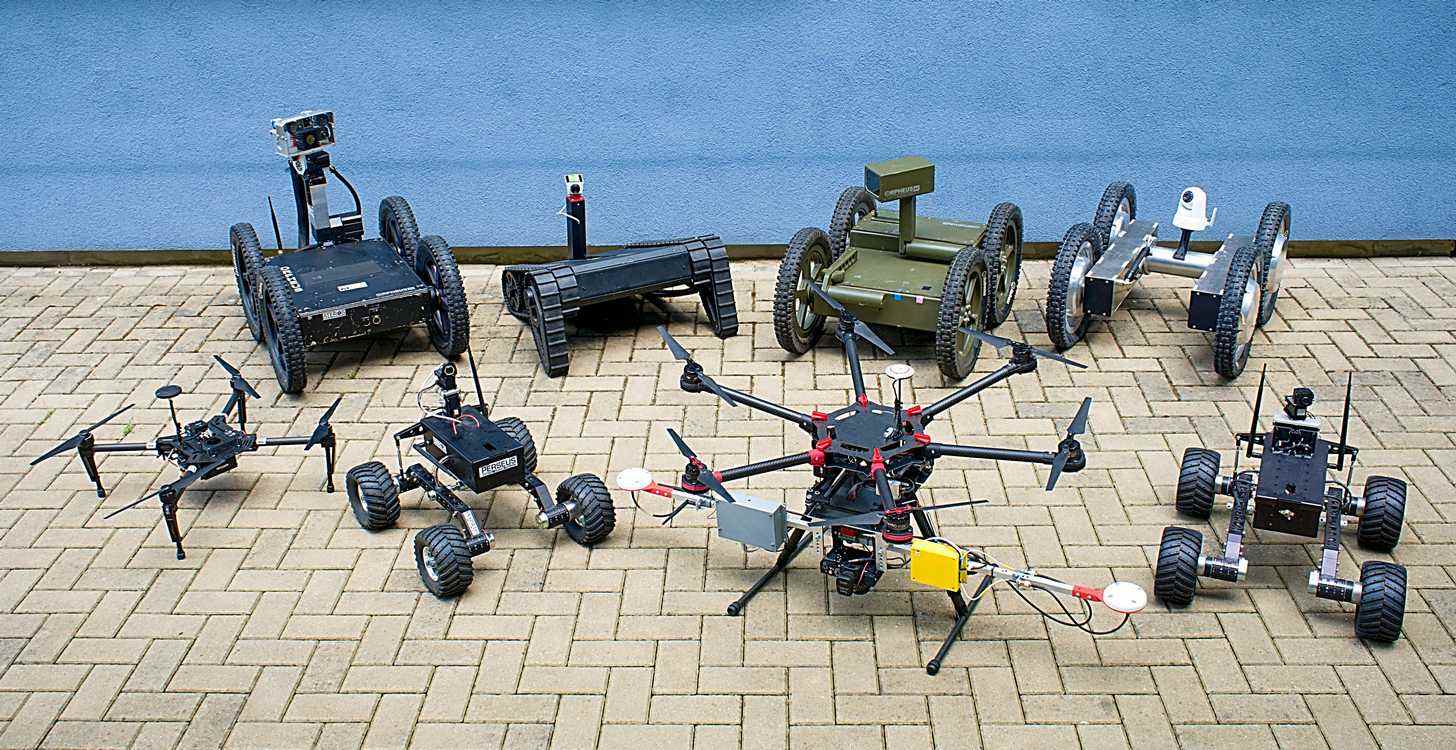}
	\caption{The four-wheel Orpheus robot family and other platforms of the heterogeneous reconnaissance mobile robot system ATEROS.  [1-column figure]}
	\label{fig:ateros}
\end{figure}

CBRNE robotics and multi-robot systems have for almost two decades embodied the research focus of the Robotics and AI group headed by Prof. Zalud at Brno University of Technology. The Orpheus reconnaissance robot family \cite{burian_multi-robot_2014}, a central project pursuing the development of four-wheel skid-driving portable CBRNE robots (Figure~\ref{fig:ateros}), is being continuously refined and has been employed in various experiments and missions, such as those devised to determine water contamination \cite{nejdl_remote-controlled_2015}. In the context of the topic, we have examined automatic radiation mapping thanks to the robot's built-in RTK global navigation satellite system (GNSS) -based navigation system, establishing that a UGV is capable of substituting for human-performed measurements effectively, more accurately, and without safety risks \cite{jilek_radiation_2015}. However, the approach was not subjected to comprehensive testing, including, for example, obstacle-accommodated environment.

To extend the usability of the terrestrial platforms, we developed a multi-sensor system for UAS photogrammetry to assemble high-resolution orthophotos and surface models \cite{gabrlik_calibration_2018}. Benefiting from the capability of operating without ground georeferencing targets, the solution is perfectly convenient for radiation-related tasks; the products are applicable in UGV trajectory planning under difficult conditions. Moreover, our simulations suggest that the surface model may find use in aerial radiation mapping, too \cite{gabrlik_simulation_2018}. In radiation detection system-equipped UASs, flying at a constant altitude above ground level (AGL) collects more consistent data compared to flying at a constant mean sea level (MSL) altitude, thus making source localization more accurate. All the above-mentioned equipment, methods, and experience enabled us to compose a comprehensive multi-platform system for automatic radiation source search. A first attempt in this field was published previously \cite{lazna_cooperation_2018}; however, numerous aspects and issues still remain to be addressed to increase the reliability, credibility, robot interoperability, and overall real-world usability, i.e., the main topics dealt with in this research.

%% file: text/methods-experiment.tex
\subsection{Experiment Setup}

\begin{figure*}[!tbp]
	\centering
	\includegraphics[width=0.9\textwidth]{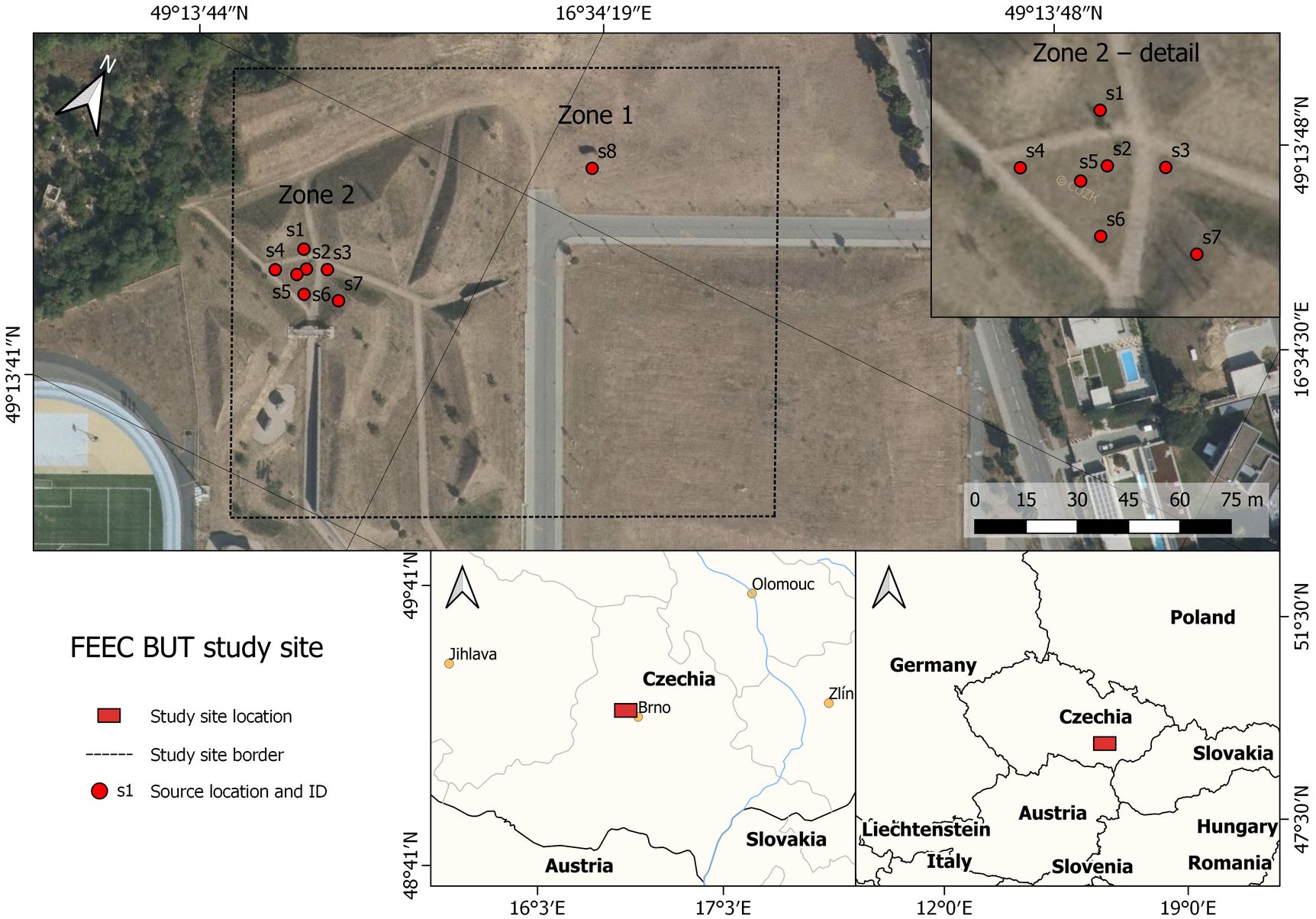}
	\caption{The location of the study site, and the spatial distribution of the radiation sources (orthophoto courtesy of the State Administration of Land Surveying and Cadastre \cite{cuzk_cuzk_2010}. FEEC BUT: Faculty of Electrical Engineering and Communication at Brno University of Technology [2-column figure]}
	\label{fig:overview-map}
\end{figure*}

The method for multi-robot radiation mapping and source localization presented in the paper was evaluated by utilizing a fictitious accident at a site in close proximity to the campus of Brno University of Technology, Brno, the Czech Republic, in August 2018 (Figure~\ref{fig:overview-map}). The goal was to arrange authentic conditions corresponding to a scenario with several gamma radiation sources lost in a certain area after a car accident. Regarding the parameters known to the tested method, the exact location, number, and activity were undefined; we can nevertheless assume that the sources belong to the class utilized in the civil sector, and the application options thus involve, for example, the calibration of devices for nondestructive testing, flow meters, level measurement systems, nuclear densometers, and density well-logging probes. 

The experiment site occupies an area of 20,000~m$^2$, comprising mainly grassy terrain with various man-made objects such as a road, paths, climbing walls, and several vehicles involved in the car accident. While one half of the location is relatively flat ($<4^{\circ}$), the other includes hills with slopes up to $30^{\circ}$ and other UGV-impassable zones.

Within the experiment site, we planted eight gamma radiation sources, namely, Co-60 and Cs-137 isotopes exhibiting the activity of 2.9--123.8 MBq (Table~\ref{tab:sources}). As is evident from Figure~\ref{fig:overview-map}, the sources are scattered inside two locations: Zone 1, containing a single, high-activity source, and zone 2, which includes seven sources representing the areal contamination. To ensure safety, the relevant area was closed to common access during the experiment.

\begin{table}[!tbp]
	\caption{Radiation sources used in the experiment.}
	\begin{center}
		{\def\arraystretch{1.3}
			\begin{tabular}{lccc} 
				\hline
				\bf{Source} & \bf{Zone} & \bf{Isotope} & \bf{Activity [MBq]}  \\ \hline
				s1 & 2 & Co-60 & 2.85 \\
				s2 & 2 & Cs-137 & 7.53 \\
				s3 & 2 & Co-60 & 2.95 \\
				s4 & 2 & Cs-137 & 7.53 \\
				s5 & 2 & Cs-137 & 79.82 \\
				s6 & 2 & Co-60 & 24.56 \\ 
				s7 & 2 & Co-60 & 24.76 \\
				s8 & 1 & Co-60 & 123.78 \\
				\hline
		\end{tabular}}
	\end{center}
	\label{tab:sources}
\end{table}



%% file: text/methods-overview.tex
\subsection{Method Overview}

Robot-based environmental mapping in an outdoor environment generally embodies a challenging task due to the largely variable conditions that may be encountered, especially in terms of the terrain, vegetation diversity, and weather conditions. Moreover, further special requirements may arise as regards the measuring equipment and time constraints. In this context, choosing the proper robotic platform is crucial to achieve the desired results.

To perform the radiation mapping and source localization tasks, we designed a method operating two different robots, namely, a hexacopter UAS and a four-wheel, skid-steering UGV. The former platform, described in detail within section~\ref{ssec:uas}, enables us to cover a large area within a reasonable time, regardless of the terrain nature; however, the distance from the surface may limit the applicability of some sensors. Advantageously, at the initial stages of the procedure, the vehicle is employed to carry out the aerial photogrammetry and sparse radiation mapping. The latter platform (section~\ref{ssec:ugv}) is suitable for the reconnaissance and mapping of small areas (hundreds of square meters) only, due to its low operation speed; another limiting factor rests in the reduced terrain negotiability, depending on the slope pattern. Thus, the UGV finds use in precise radiation mapping and source localization, namely, at the final stages.

\begin{figure*}[!tbp]
	\centering
	\includegraphics[width=1\textwidth]{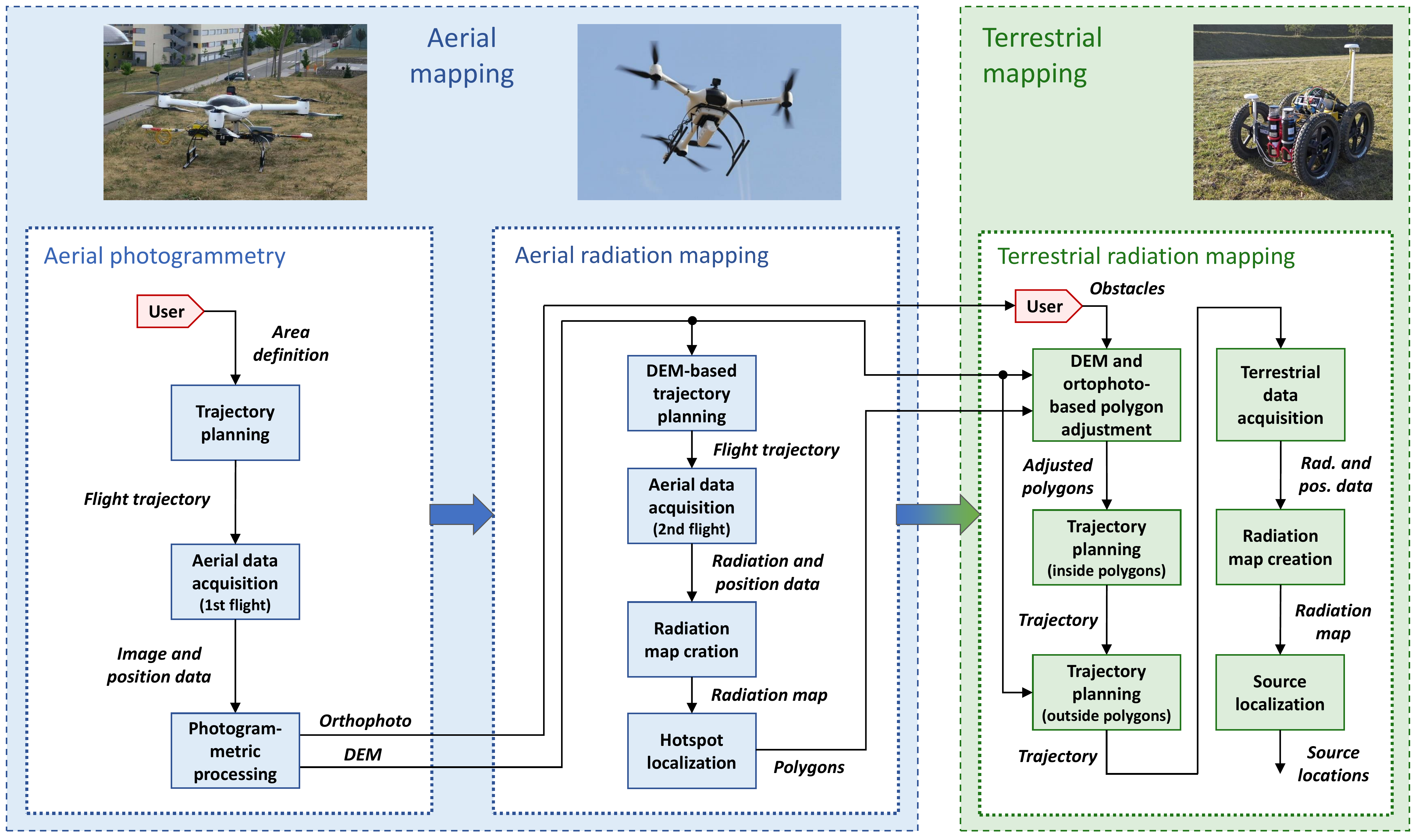}
	\caption{The sequence of the operations that form the entire process. The actual mapping comprises the aerial (blue) and terrestrial (green) branches; the user interventions are highlighted in red. DEM: digital elevation model. [2-column figure]}
	\label{fig:dataflow}
\end{figure*}

As is evident from Figure~\ref{fig:dataflow}, our approach comprises the following three phases: aerial photogrammetry, aerial radiation mapping, and terrestrial radiation mapping. The first phase aims to create the actual orthophoto and 3D model of the area, meaning products to be utilized later for the trajectory planning and to help operators orientate themselves in the unknown environment. The initial step, namely, defining the area of interest, must be executed by a user considering the current situation; however, the following operations, such as the actual flight, are already fully automatic, with the UAS's trajectory designed according to the photogrammetric requirements. Yet, from the general perspective, the entire operation must still be supervised by a pilot, especially due to safety and legal concerns. The outcomes of the photogrammetric processing (section~\ref{ssec:aerial-photo}) and the first phase as a whole embody a georeferenced orthophoto and a DEM.

The second phase is intended to localize potential radiation hotspots by means of aerial radiation mapping of the entire area. In order to obtain credible results, the UAS trajectory design encompasses the DEM acquired within the previous phase to allow us to operate at a constant height above ground level (AGL). This procedure is described thoroughly in section~\ref{ssec:aerial-rad-map}. Once the sparse radiation map is available, our algorithm selects the sub-areas that exhibit increased radiation levels (section~\ref{sec:roiselection}).

The goal of the final phase consists in building detailed radiation maps of the hotspots by using the UGV; this step facilitates the potential localization of individual sources. To perform such a task, we must consider the degree limited terrain negotiability limitation in the relevant platform, and thus the selected regions are adjusted via both the DEM-based obstacle map and the orthophoto, where other possible obstacles and impassable locations are selected by the user. The aforementioned mechanisms are addressed in section~\ref{sec:roiselection}. The UGV trajectory planning problem can then be divided into two tasks, namely, covering the pre-specified polygons (hotspots) and executing A*-based robot navigation between the polygons (section~\ref{ssec:terr-rad-map}). The collected data are employed to generate a detailed radiation map and to allow the source localization. This stage, described in section~\ref{ssec:source-localization}, involves utilizing the least-square method to estimate both the precise location of the individual sources and their approximate activity.

%% file: text/methods-uas.tex
\subsection{Unmanned Aerial System}
\label{ssec:uas}

In our method, the UAS embodies a platform to ensure aerial data acquisition during the initial stage of the mapping process. The system must be capable of carrying various sensors (to perform photogrammetry and ionizing radiation measurement) and flying according to the preprogrammed trajectories. For these tasks, UASs were already employed previously. The discussed approach to aerial photogrammetry is one of the most common applications of UASs in the scientific domain and the commercial sector; the technique finds use in, for example, agriculture, to monitor crop height \cite{torres-sanchez_assessing_2017,belton_crop_2019}; forestry, to collect tree inventory data \cite{mikita_forest_2016}; geodesy, to investigate topographic changes  \cite{james_3-d_2017}; and archaeology, to deliver spatial site reconstruction \cite{waagen_new_2019}. As described in the Introduction, unmanned aircraft have also been deployed for ionizing radiation mapping. At present, there are science-based efforts to utilize the systems in legacy mine mapping, nuclear waste storage inspection, and disaster response scenarios; compared to aerial photogrammetry, however, these domains offer only marginal options to use UASs. The choice of a suitable unmanned platform depends on multiple factors, of which the most substantial ones are the payload capacity and operation range. UASs with rotary wings (helicopters and multicopters) provide excellent payload capacities, with another significant advantage being the ability to operate at low speeds and to hover. Conversely, fixed-wing unmanned aircraft of similar size and weight exhibit better ranges and endurances, but the load capability is typically lower.

\begin{figure}[!tbp]
	\centering
	\subfloat[]{\includegraphics[width=0.4\textwidth]{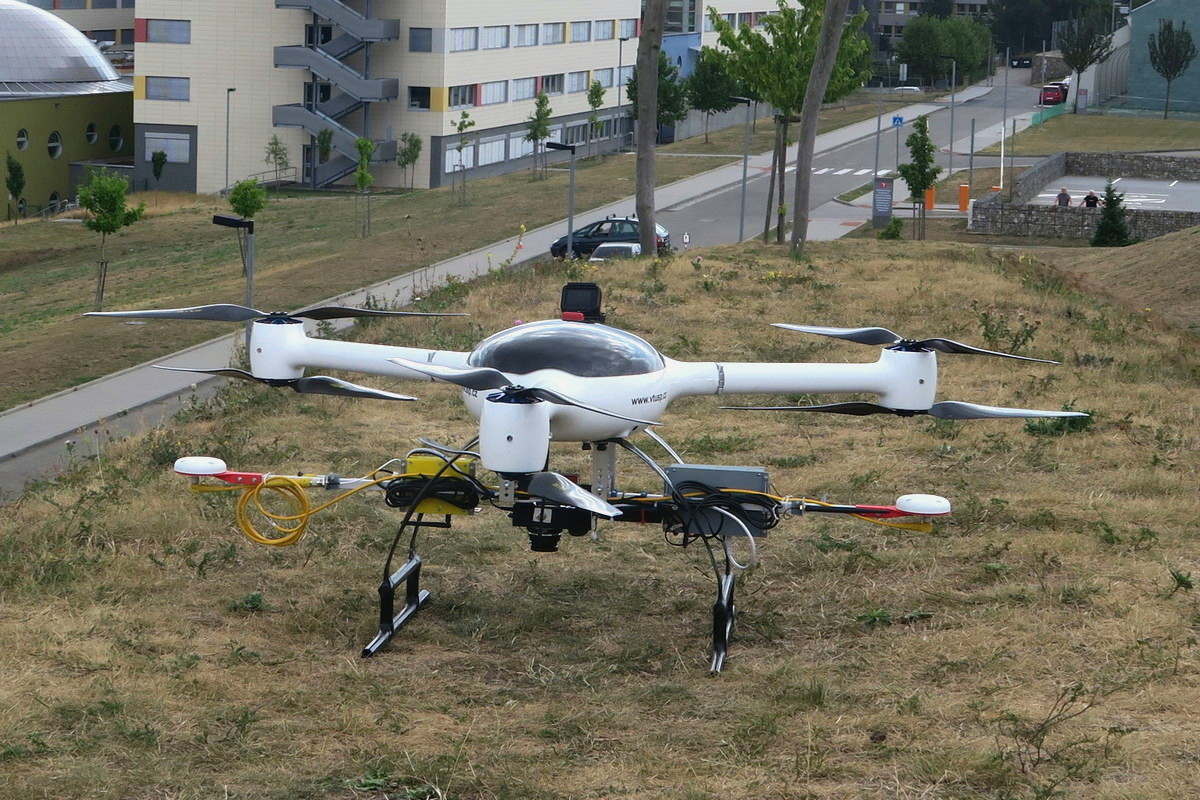}\label{fig:brus_foto}}
	\quad
	\subfloat[]{\includegraphics[width=0.4\textwidth]{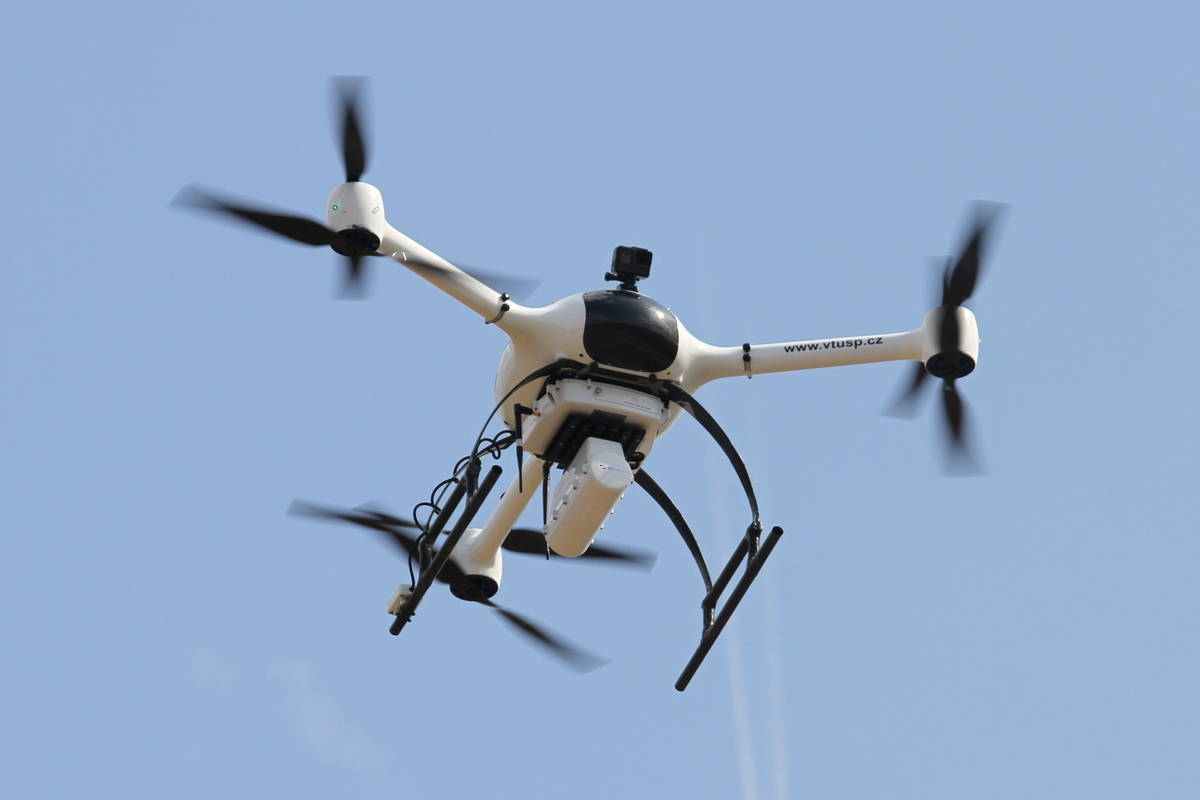}\label{fig:brus_rad}}
	\caption{ The BRUS UAS configured for the experiment: with a multi-sensor system to perform the photogrammetry (a), and carrying a gamma radiation detection setup (b). UAS: unmanned aircraft system [1-column figure]}
	\label{fig:brus}
\end{figure}

The proposed mapping and search method involves low-altitude radiation measurement, which requires high maneuverability and flying at low speeds. For this reason, and also due to the equipment weight, we chose the BRUS rotary-wing UAS (Figure.~\ref{fig:brus}), a six-rotor aircraft developed and produced by the Military Technical Institute of the Czech Republic \cite{military_technical_institute_brus_2018}. The UAS was designed to perform various tasks, such as those allowing the reconnaissance, monitoring, and rescue missions during natural disasters. The device's in-house control system supports both manual and waypoint-based automatic flight, whereas the automatic mode offers wide configurability. The payload mounting system at the bottom of the fuselage facilitates carrying equipment of up to 4 kg in weight, and the standard endurance is 45 minutes; more parameters are indicated in Table~\ref{tab:robots}.

\begin{table*}[!tbp]
	\caption{The parameters of the BRUS and Orpheus-X3 unmanned platforms. UAS: unmanned aircraft system; UGV: unmanned ground vehicle.}
	\label{tab:robots}
	\begin{center}
		{\def\arraystretch{1.3}
			\begin{tabular}{lcc}
				\hline
				{\bf Parameter} & {\bf UAS} & {\bf UGV} \\
				\hline
				Dimensions & 1.2 $\times$ 1.2 $\times$ 0.5 m & 1.0 $\times$ 0.6 $\times$ 0.4 m \\ 
				Weight & 4.5 kg & 51 kg \\ 	
				Max payload weight & 4 kg & 30 kg \\ 
				Standard operational time & 45 mins & 120 mins \\ 			
				Drive type & multi-rotor & wheel, differential \\ 			
				Operating speed & 2 -- 5 m/s & 0.6 m/s \\ 			
				Max speed & 16.7 m/s & 4.2 m/s \\
				\hline
		\end{tabular}}
	\end{center}
\end{table*}
At the initial stage of the actual mapping, the UAS is fitted with a custom-built multi-sensor system for aerial photogrammetry, illustrated in Figure~\ref{fig:brus_foto}. This setup enables us to create georeferenced photogrammetric products, namely, an orthophoto or a DEM, without requiring ground control points (GCPs). The sensors are a consumer-grade Sony Alpha A7 digital camera, a Trimble BD982 dual-antenna GNSS receiver supporting the RTK technology and vector measurement, and an SBG Ellipse-E inertial navigation system (INS). The last-named component performs real-time data fusion to estimate the position and orientation; thanks to precise time synchronization, the exterior orientation parameters of each captured image are collected. To attain high positioning accuracy, a base station transmitting real-time kinematic correction data must be deployed; normally, the station is positioned close to the study site. The system weighs 2.8 kg and is completely independent from the applied UAS platform. The setup was previously described in more detail within article \cite{gabrlik_calibration_2018} and subsequently found use in, for example, UAS-based aerial snow depth mapping \cite{gabrlik_towards_2019}. The existing results indicate that the system is capable of reaching centimeter-level object accuracy. Even though GCP-free UAS photogrammetry is markedly less common than the approach that relies on ground targets, the concept was already utilized in previous research \cite{eling_development_2015,bliakharskii_modelling_2019,hasheminasab_gnssins-assisted_2020}.

The second phase of the mapping cycle comprises ionizing radiation measurement; for this purpose, the UAS is fitted with a NUVIA DRONES-G radiation detection system (Figure~\ref{fig:brus_rad}). The setup was specially designed for light airborne radiation monitoring via UASs, and it consists of two modules: a base one and a detection one. The former module contains a processing unit, a data storage device, an RF datalink, a GNSS module, a laser altimeter, a battery source, and other relevant electronic components. The latter module is equipped with a 2$\times$2" NaI(Tl) detector operating in the 50 keV -- 3 MeV energy range, allowing us to measure the dose rates in the range of 50 nGy -- 100 \textmu Gy. The one-second radiation data (spectra recordings) are georeferenced and saved, enabling later analysis. The entire detection system weighs 3 kg and, similarly to the photogrammetry equipment, is independent of the applied airborne platform.

%% file: text/methods-ugv.tex
\subsection{Unmanned Ground Vehicle}
\label{ssec:ugv}

An Orpheus-X4 UGV is utilized as the experimentation platform. The robot is a mid-size four-wheeled UGV developed by the Robotics and Artificial Intelligence Group at the Department of Control and Instrumentation, FEEC BUT. The Orpheus-X4 is well-suited for outdoor environments thanks to its tank-like differential drive with four independent actuators; conversely, this type of drive is not convenient for specific types of indoor obstacles, such as stairs. However, we can assume that a radiation accident area resembles a post-disaster site; moreover, the current navigation module is based on the GNSS technology, which requires outdoor operation, and the Orpheus-X4 structure suits the needs of radiation mapping missions. Besides the drive, the robot is equipped with a camera head carried by a 3-degrees of freedom (DOF) manipulator. The head finds use in remote control of the UGV by an operator. The robot contains two computers to run various control algorithms; its parameters are summarized in Table~\ref{tab:robots}. A more detailed description of Orpheus robots is proposed in articles \cite{kocmanova_effective_2015} and \cite{zalud_orpheus_2008}.

The Orpheus-X4 offers automatic navigation along the planned trajectory, which is represented by a sequence of waypoints. The navigation is designed and implemented at our department, thus allowing fine-tuning for a specific mobile robot and application. The emphasis on parameterizability during the design and implementation stages enables the configuration to be represented by the chassis coefficients and required motion characteristics. The attainable accuracy of the Orpheus-X4 automatic navigation relies on self-localization accuracy. In the case of a good and stable RTK solution, it is possible to reach 3 cm ($1 \sigma$) in stable flat surfaces \cite{phdtomj}; however, the accuracy is generally much worse in unstable traction terrains.

The self-localization of a mobile robot employs a GNSS, an INS, and wheel odometry. The system relies mainly on the RTK GNSS to solve the 2D position and heading. The dead reckoning solutions from the micro-electro-mechanical systems (MEMS) -based INS and wheel odometry are used to bypass insufficient GNSS solutions. The short-term GNSS position and heading outages in cases of dynamic robot motion are bypassed by the INS. Long-term GNSS outages during low motion stages are bridged by a wheel odometry-based solution that employs INS heading. We use a Trimble BD982 GNSS receiver in a dual antenna setup to obtain the RTK solution of the position and heading. Our configuration exploits dual-frequency tracking of the GPS, GLONASS, and Galileo signals to achieve reasonable position accuracy (better than 10 cm). All data are updated at 50 Hz. The correction data for the RTK processing are typically obtained from our own local GNSS base station, which utilizes the same BD982 module as the rover. Although we need not know the accurate position of the base station antenna, it is beneficial to use such a position to synchronize our work with other systems that do not employ the base station with a mission-specific coordinate system. Alternatively, it is possible to use data from commercial permanent GNSS stations or networks accessible via global system for mobile communications (GSM) network connection.

To perform the robotic mapping of gamma radiation, scintillation detectors seem to make a trade-off for the desired features. The detectors provide a high density and volume, thus being beneficial in maximizing the probability of interaction with gamma rays for good sensitivity. To detect a radiation source from a greater distance, the sensitivity embodies a crucial parameter. Moreover, common inorganic scintillators possess spectrometric abilities; knowledge of the spectra enables us to identify different radionuclides and can facilitate separating useful information from the radiation background. To develop the mapping system, we chose the combination of a 2$\times$2$^{\prime\prime}$ NaI(Tl) scintillator and a photomultiplier tube, mainly thanks to its accessibility, conventionality, and previous experience. The detection system exhibits an energy resolution of more than 7 \% at 662 keV \cite{ahmed_physics_2007}. The detected values are read out by a NUVIA MCB3 multichannel analyzer  comprising a high voltage source, a preamplifier, an analog-to-digital converter, a processing unit, and an Ethernet communication interface. The analyzer provides a resolution of up to 4,096 channels and is usually calibrated to the 0.03--3 MeV energy range. If not stated otherwise, the presented algorithms work with the total count (TC) value, i.e., the sum of counts in all channels. 

\begin{figure}[!tbp]
	\centering
	\includegraphics[width=0.45\textwidth]{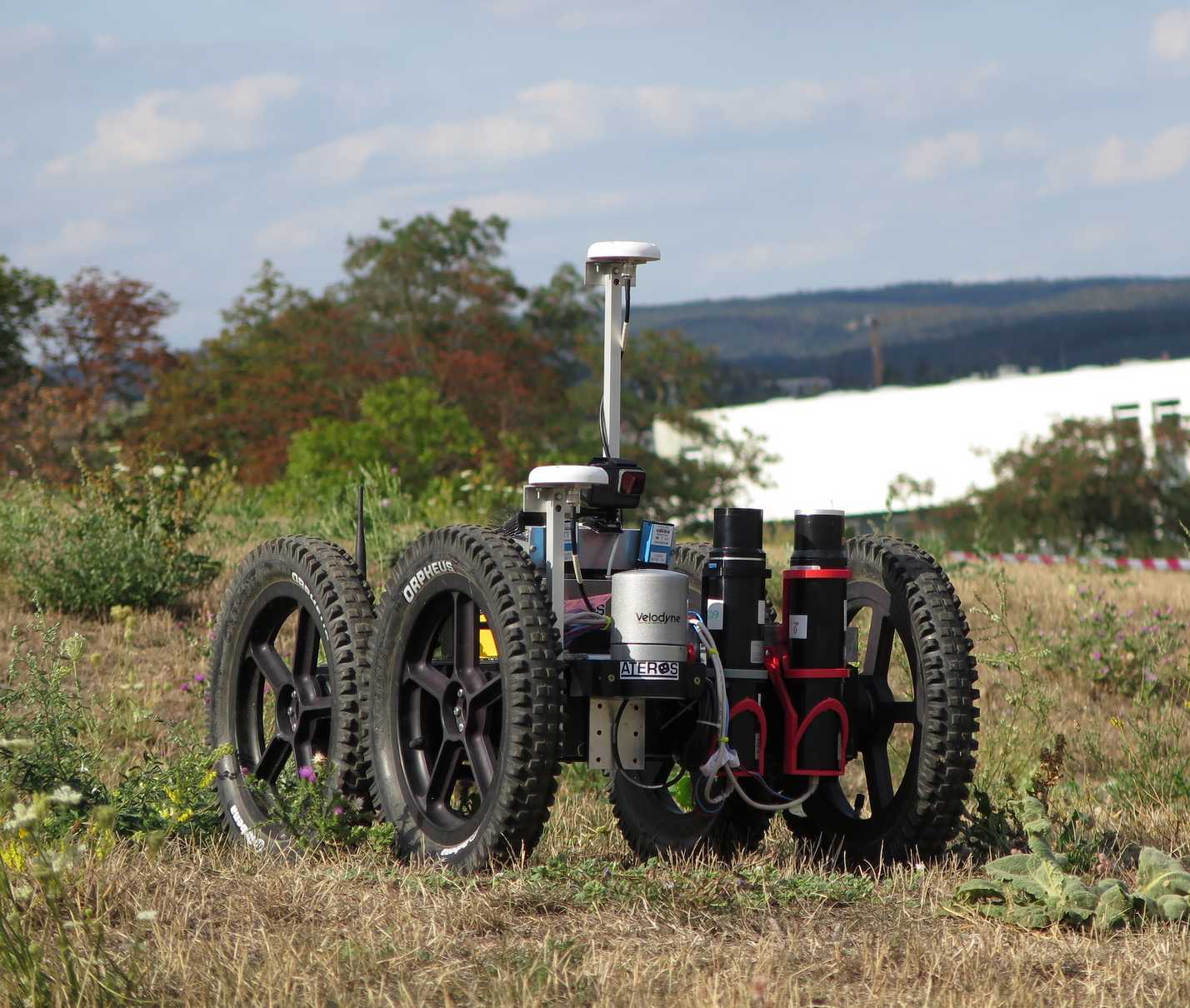}
	\caption{The Orpheus-X4 UGV equipped with a GNSS receiver and a pair of NaI(Tl) radiation detectors. UGV: unmanned ground vehicle; GNSS: global navigation satellite system. [1-column figure]}
	\label{fig:orpheus}
\end{figure}

In the applied system, the UGV is equipped with a pair of NaI(Tl) detectors. An advantage of utilizing  multiple detectors consists in higher sensitivity of the measurement system and better cumulative statistical attributes of the measurements. Even though the detectors influence each other in terms of shading, they still embody two statistically independent sources of information regarding the stochastic nature of radioactive decay. A cumulative spectrum provides information with a statistical significance higher than that of single spectra \cite{knoll_radiation_2010}. The Orpheus-X4 carrying all the presented equipment is shown in Figure~\ref{fig:orpheus}.

%% file: text/methods-photogrammetry.tex
\subsection{Aerial Photogrammetry}
\label{ssec:aerial-photo}

Aerial photogrammetry embodies the first phase of the mapping method, and its goal is to map an unknown area, yielding an orthophoto and a DEM. These products are of vital importance for the next stages of the technique, trajectory planning to support the UAS- and UGV-based radiation mapping in particular. As our experiment simulates a situation where the relevant area is potentially contaminated and dangerous for humans, the aerial mapping process may not require prior field work. Thus, we employed a custom-built multi-sensor system for aerial photogrammetry to facilitate direct georeferencing (section~\ref{ssec:uas}).

\begin{figure}[!tbp]
	\centering
	\includegraphics[width=0.4\textwidth]{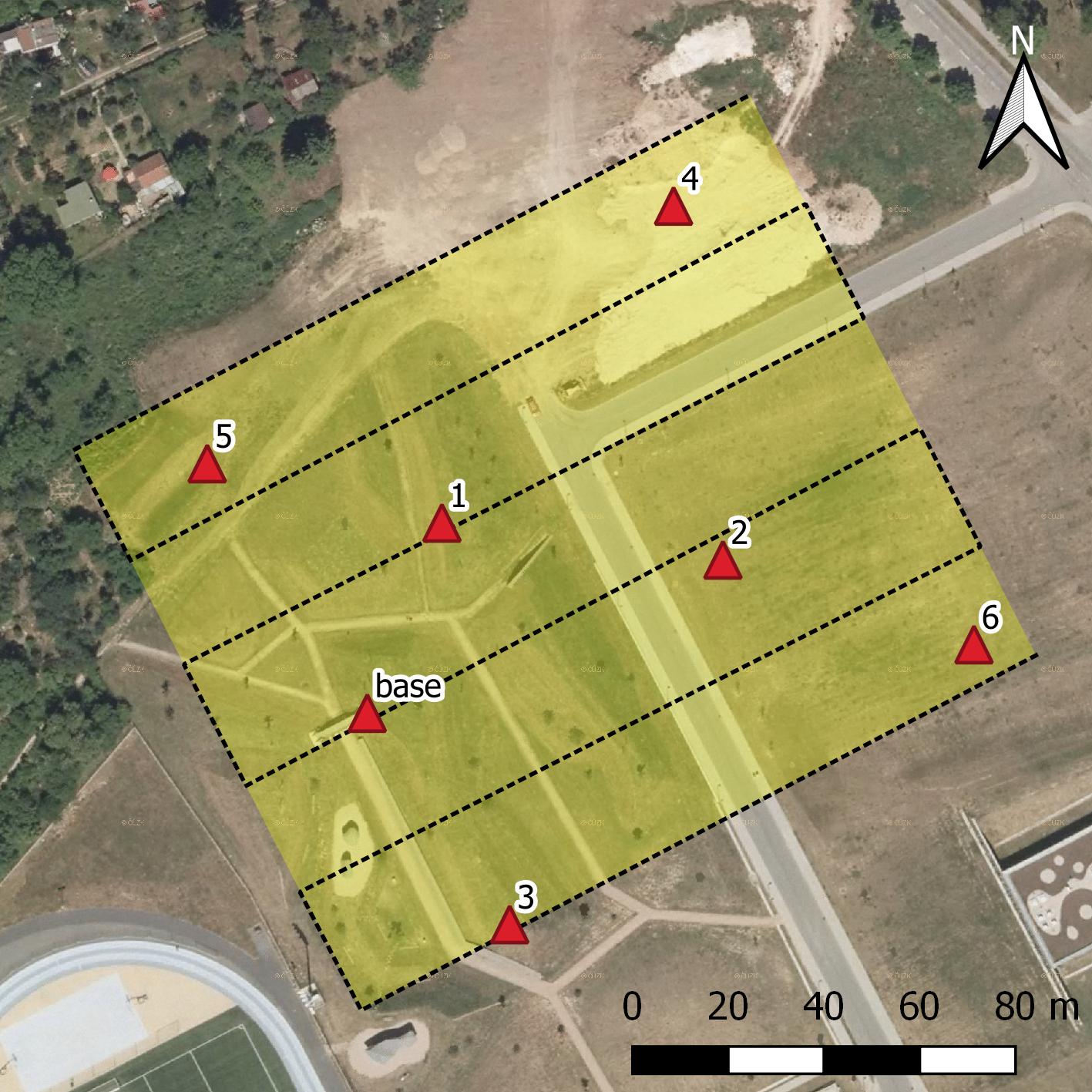}
	\caption{The UAS trajectory planned for the photogrammetry flight. The yellow rectangle represents the study site (having an area of 20,000~m$^2$), and the red triangles indicate the positions of the ground targets. UAS: unmanned aircraft system. [1-column figure]}
	\label{fig:uas-traj-photo-plan}
\end{figure}

The process starts with trajectory planning across the user-defined region. We utilize the common parallel strips flight pattern known from both manned and unmanned aerial photogrammetry \cite{kraus_photogrammetry:_2007,cabreira_survey_2019}, as shown in Figure~\ref{fig:uas-traj-photo-plan}. The trajectory and data acquisition parameters, such as the flying altitude and distance between the strips and images, must satisfy certain photogrammetric rules, including the desired forward and side image overlaps, ground resolution, and altitude restrictions. The trajectory planning also depends on the intrinsic parameters of the applied camera. The relevant values applied in the presented experiment are summarized in Table~\ref{tab:uav-trajectory}. In general terms, the image overlapping rate reaches approximately 90\% to achieve high object accuracy, and the ground resolution is selected to range within the centimeter level to capture even the smallest details relative to the dimensions of the UGV which will carry out the terrestrial mapping.

\begin{table}[!tbp]
	\caption{The parameters of the flight trajectories and data acquisition for both flights (one enabling the photogrammetry and the other facilitating the radiation mapping). ATOP: above take-off point; AGL: above ground level; ISO: international organization for standardization.}
	\begin{center}
		{\def\arraystretch{1.3}
			\begin{tabular}{lcc} 
				\hline
				\bf{Parameter} & \bf{1st flight} & \bf{2nd flight}  \\ \hline
				Number of strips & 6 & 14 \\
				Strip length & 160 m & 140 m \\
				Distance between strips &  26 m & 10 m \\
				Flying altitude &  60 m ATOP & 15 m AGL  \\
				Flying speed &  5 m/s &  2 m/s \\
				Sampling period &  2 s &  1 s  \\ 
				Base & 10 m & 2 m \\ \hline
				Photo scale &  1:5,200 & ---  \\
				Forward overlap &  92 \% & ---  \\
				Side overlap &  84 \% & ---  \\
				Image footprint &   190 $\times$ 125 m & ---  \\
				Ground resolution &  3.1 cm/px & --- \\ 
				Shutter speed &  1/1,000 s & --- \\
				Aperture &  5.6 & --- \\
				ISO &  Auto (100--400) & --- \\ 
				\hline
		\end{tabular}}
	\end{center}
	\label{tab:uav-trajectory}
\end{table}

The photogrammetric dataset comprises two types of data: images and georeferencing-related information (positions, orientations, and accuracies). These data are processed with Agisoft Photoscan Professional (version~1.4.2), a comprehensive software package to execute all the photogrammetric processing stages. The workflow starts with the align phase, where the exterior and interior orientations of the camera \cite{hartley_multiple_2004} are estimated based on the feature points detected in the overlapping images. Further, the locations of the feature points are determined via the structure-from-motion procedure, resulting in a sparse point cloud \cite{szeliski_computer_2011}. Within the following step, a dense point cloud can be generated by means of multi-view stereo (MVS) reconstruction. As the operation is performed at the level of pixels, even small details are reconstructed, yielding a point cloud containing millions of points. This product is then employed to generate the orthophoto and DEM.

Photoscan can transform the camera poses and points into a geographic coordinate system in two ways: via data measured by onboard sensors, a method known as direct georeferencing, or with GCPs, namely, the indirect georeferencing approach. In general terms, the latter technique is utilized more frequently, as it allows us to reach high georeferencing accuracy even with inexpensive, consumer-grade onboard sensors; moreover, the results are reliable, and inter-sensor calibration is not required \cite{verhoeven_mapping_2012,oniga_determining_2020}. The indispensability of well-distributed ground targets, however, makes the method unsuitable for CBRNE tasks. The direct technique, conversely, relies on onboard sensors only, leading to higher requirements on payload capacity and the calibration process; in this context, it should also be stressed that the georeferencing results are typically less accurate, as indicated in relevant studies. \cite{shahbazi_development_2015,fazeli_evaluating_2016,gabrlik_calibration_2018}.

Although we utilized a direct georeferencing system, six ground targets were deployed in the area prior to the experiment (Figure~\ref{fig:uas-traj-photo-plan}). This step enabled us to determine the accuracy of the photogrammetric products, i.e., the orthophoto and DEM, which will be necessary during the subsequent stages of the radiation mapping method. Moreover, the targets would allow us to perform the georeferencing even under a failure of the onboard GNSS/INS. We used 20 cm-sized, squared, black and white patterned paper targets having clearly defined centers. All the targets were glued onto a solid support and fixed to the ground with iron nails. The position of every single target was acquired by a Trimble BX982 RTK GNSS receiver obtaining the correction data from CZEPOS (Czech provider of correction data). With the same equipment, we determined the position of the GNSS base station, indicated in Figure~\ref{fig:uas-traj-photo-plan}; the station provided the correction data for the GNSS receiver aboard the UAS (and the UGV), allowing us to carry out the mission even when the correction data provider was not available.

%% file: text/methods-radiation-uas.tex
\subsection{Aerial Radiation Mapping}
\label{ssec:aerial-rad-map}

\begin{figure}[!tbp]
	\centering
	\includegraphics[width=0.35\textwidth]{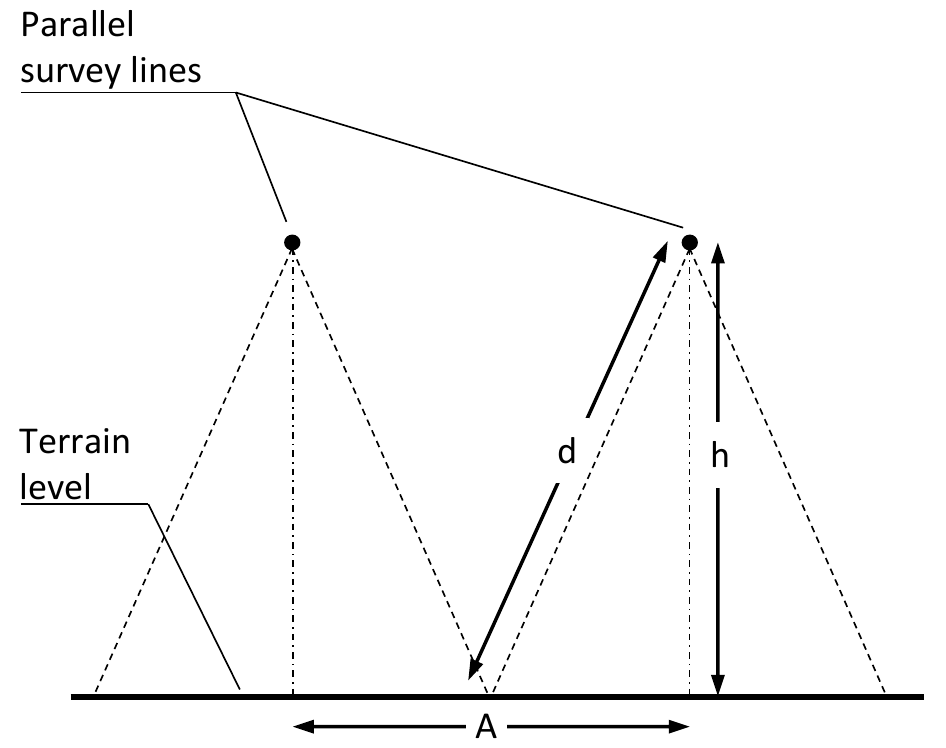}
	\caption{The basic parameters of the UAS trajectory for the radiation mapping procedure. UAS: unmanned aircraft system. [1-column figure]}
	\label{fig:planning-uas-rad}
\end{figure}

Aerial radiation mapping, described within the diagram in Figure~\ref{fig:dataflow}, embodies the second phase of the mapping method. The procedure aims to create a sparse ionizing radiation map of the entire study site to localize possible hotspots to be mapped with the UGV. Without any prior knowledge of the hotspots, and lacking a detection system with directional sensitivity, the straightforward flight strategy comprises parallel survey lines, similarly to the previous photogrammetry flight. The main difference rests in the setting of major parameters, including flight altitude, distance between strips, and speed. As the dose rate decreases with the square of the distance, the AGL altitude must be as low as possible to detect even weak sources. In practice, the minimum flight altitude is always limited by the actual precision of the UAS navigation system. Regarding the terrain shape and obstacles, the lowest possible value for the applied UAS equals 15~m AGL. However, the distance $d$ between the source and the detector is, in addition to the vertical component $h$, formed also by the horizontal distance. The condition $d=h$ applies when the UAS is directly above the source, and the formula (\ref{eq:distd}) describes the marginal situation when the source is located exactly between the survey lines being $A$ meters apart (Figure~\ref{fig:planning-uas-rad}).

\begin{equation} 
d = \sqrt{\left(\frac{A}{2}\right)^2+h^2}
\label{eq:distd}
\end{equation}

\begin{figure}[!tbp]
	\centering
	\includegraphics[width=0.4\textwidth]{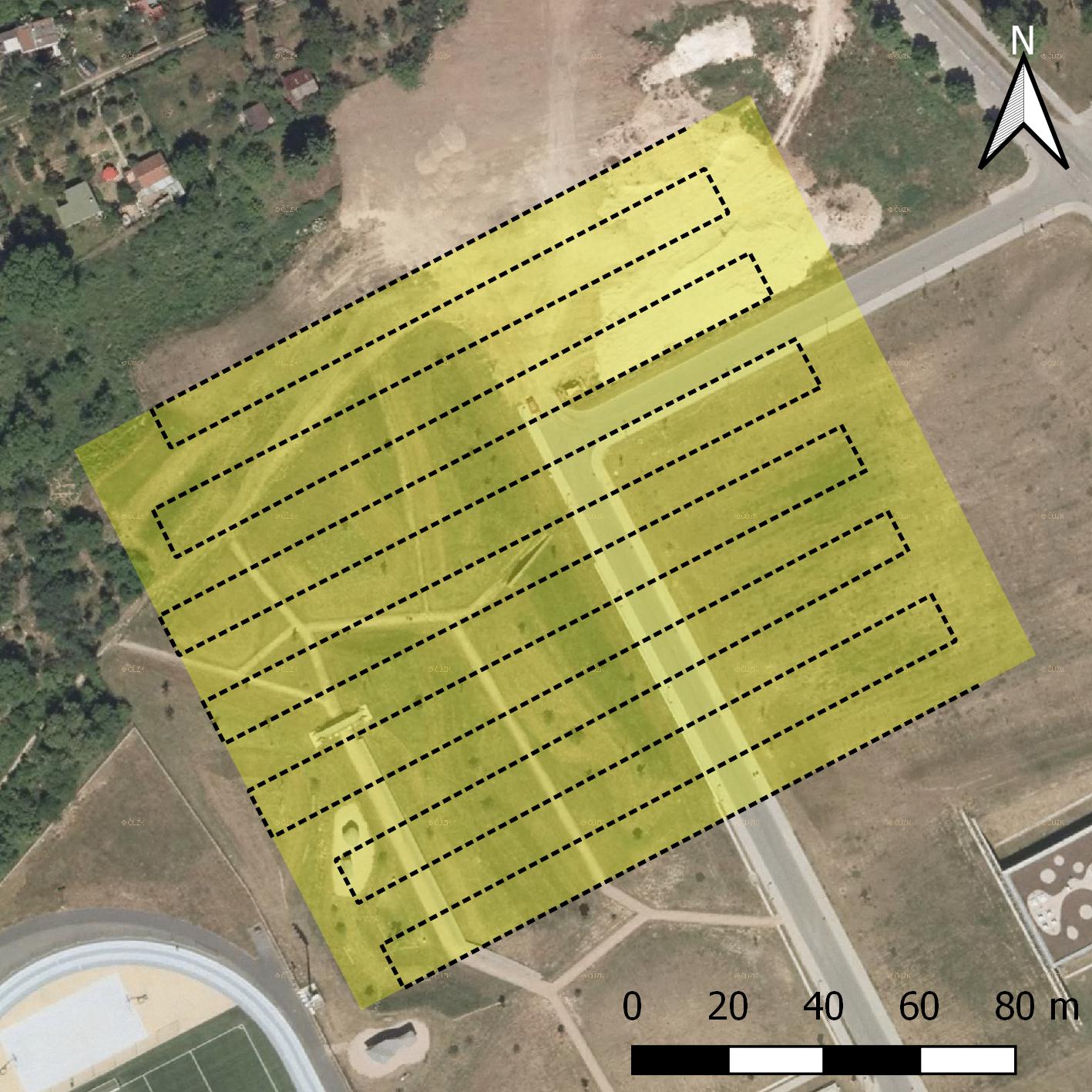}
	\caption{The UAS trajectory planned for the radiation measurement flight (the yellow rectangle represents the study site). UAS: unmanned aircraft system. [1-column figure]}
	\label{fig:uas-traj-rad-plan}
\end{figure}

The distance $A$ must be chosen with respect to the following factors, attributes, and scenarios: A low value yields high spatial resolution and the possibility of localizing a hotspot in a precise manner, even though it can result in a very long flight trajectory (operation time) and the related necessity to replace the battery frequently. Contrariwise, a high distance between the flight lines is time-saving but can negatively affect the ability to detect low-intensity sources. Considering all the circumstances, the value $A=10$~m constitutes optimal setting in our experiment, namely, the UAS can cover the relevant area without battery replacement, and the maximum possible distance between the source and the detector ($d$) is only 5~$\%$ greater than the flight height. In this case, the intensity decreases by the acceptable value of 11~$\%$ against the $d=h$ condition. The resulting trajectory comprises 14 parallel lines and is approximately 2 kilometers long (Table~\ref{tab:uav-trajectory} and Figure~\ref{fig:uas-traj-rad-plan}).

\begin{figure}[!tbp]
	\centering
	\includegraphics[width=0.45\textwidth]{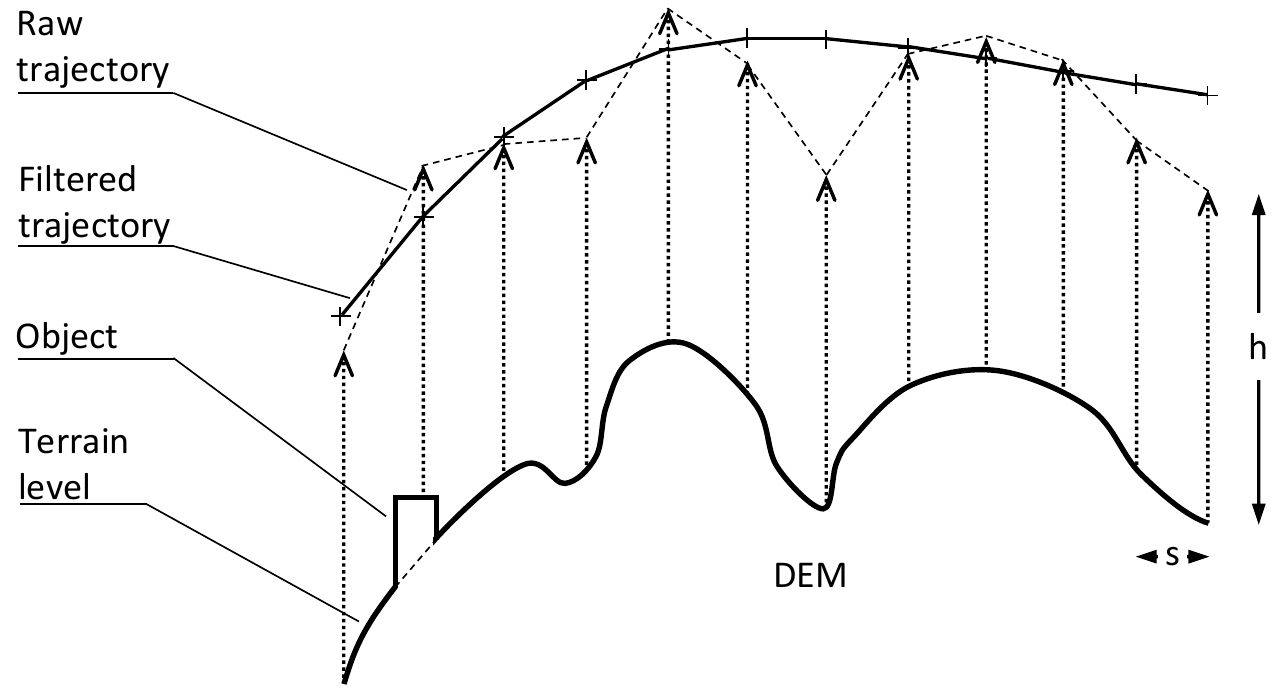}
	\caption{The principle of the terrain-based trajectory adjustment. DEM: digital elevation model. [1-column figure]}
	\label{fig:traj-adj}
\end{figure}

In UAS-based radiation mapping, the common approach involves operating at a constant MSL altitude \cite{christie_radiation_2017,martin_use_2015}, an option applicable at locations that lack significant height differences. As indicated within one of our previous papers \cite{gabrlik_simulation_2018}, major variations in the flight height above ground level produce non-homogeneous and unreliable data; thus, a means to secure a constant AGL height is essential in hilly sites. The proposed method utilizes a photogrammetry-based DEM, the output of the initial UAS flight (section~\ref{ssec:aerial-photo}), to adjust the radiation mapping trajectory (Algorithm~\ref{alg:dem-traj}). The vertical components of the trajectory are computed as the sum of the DEM heights at the given waypoints and the desired AGL flight height $h$. To follow the terrain precisely, the trajectory is split into smaller segments having a size $s$, whose value is chosen with respect to the character of the terrain. Such a trajectory, however, may contain multiple height changes; this variability, then, is not energy-efficient and can increase the operation time. To avoid the behavior, we apply a low-pass filter to obtain a smooth trajectory and to prevent sudden UAS altitude variations. The principle is illustrated in Figure~\ref{fig:traj-adj}. A drawback to the approach rests in the extensive amount of waypoints to be stored in the UAS memory. Within the presented experiment, we use $s=10$~m, the appropriate distance for the discussed study site, leading to approximately 200 waypoints.

\begin{algorithm}
	\caption{The DEM-based trajectory adjustment.}
	\begin{algorithmic}[1]
		\INPUT The horizontal trajectory $\mathrm{T}$, digital elevation model $\mathrm{D}$, AGL height $h$, and segment size $s$.
		\OUTPUT The terrain-adjusted spatial trajectory $\mathrm{T}_t$.
		\STATE \textbf{Trajectory segmentation:} Splitting the lines defined by the points $\mathrm{T}$ into smaller segments having a maximum size $s$ to obtain dense trajectory points $\mathrm{T}_s$.
		\STATE \textbf{Find the corresponding DEM points:} For every point defined in $\mathrm{T}_s$ find the nearest horizontal point of $\mathrm{D}$.
		\STATE \textbf{Compose the 3D trajectory:} Use the height values of the obtained DEM points as the height coordinates for the trajectory $\mathrm{T}_s$.
		\STATE \textbf{Compute a new altitude:} Increase the altitude of every point in $\mathrm{T}_s$ by the height $h$.
		\STATE \textbf{Filter the trajectory:} Apply the low-pass filter to the $\mathrm{T}_s$ point sequence to obtain the filtered trajectory $\mathrm{T}_t$.
	\end{algorithmic}
	\label{alg:dem-traj}
\end{algorithm}

The last parameter associated with the aerial data acquisition is flight speed. As the UAS is intended to fly close to the terrain and to change altitude markedly, we apply the rather low horizontal speed of 2~m/s. Considering the detection system sampling period of 1~s, we obtain the sampling distance of 2~m .

%% file: text/methods-polygon.tex
\subsection{Automatic Selection of the Terrestrial Mapping Areas}
\label{sec:roiselection} 

Aerial radiation mapping yields a set of scattered data points, each comprising the coordinates and spectrum acquired by the detection system during a measurement period. The data points are not very suitable for visualization and further processing of the map, namely, finding hotspots to be applied in the terrestrial survey. Thus, calculating the radiation intensity (dose rate) at points in a regular grid is required; this step can be carried out through the Delaunay triangulation \cite{amidror_scattered_2002}. The density of the data points in the axis parallel to the flight direction is approximately five times higher than that in the perpendicular axis, due to the chosen flight speed, sampling period, and distance between the strips. Regrettably, such point distribution is not convenient for the interpolation, and each four subsequent spectra are thus summed and assigned to a single position in order to achieve an even distance of points in both axes. The statistics of the cumulative spectra are significantly better, although information on areal distribution of the dose rate is partially lost; this is, however, not a cardinal issue in the given case, as more detailed terrestrial measurement follows.

Once the interpolated radiation map is available, the operator can manually mark the hotspots, or regions of interest (ROIs); this step is nevertheless best performed automatically. A possible approach is to employ a two-dimensional peak detector; such an option is unsuitable for the general case, as the data do not always represent a clear sharp peak, e.g., if
\begin{itemize}
	\item the peak comprises contributions by multiple radiation sources;
	\item the magnitude of the peak is comparable to the radiation background, as the data are very noisy due to statistical laws;
	\item the magnitude of the peak exceeds the capacity of the detector, and the dead time is over 50 \%, causing the data to exhibit a rapid decrease.
\end{itemize}

The first two cases can be certainly expected during aerial radiation mapping; thus, we adopt a different method. The basic idea is as follows: By eliminating the radiation background, a connected set will be left for each significant peak. The problem is in identifying the background, as it not only depends on the geographical location but, generally, can be increased by strong artificial sources. The unnecessary data may be assumed to lie within the three-sigma band around their mean value. To find such an {\it adaptive threshold}, the statistical parameters of the background must be estimated. An analytical solution to the described problem is not feasible, because we cannot anticipate the number of radiation sources or their activity relative to natural radionuclides and cosmic rays. Instead, an empirical threshold $T_{bg}$ is derived from the statistical parameters of the complete dataset as a sum of the dataset's mean value and a half of its standard deviation: 
\begin{equation}
	T_{bg} = \mu + \frac{\sigma}{2}
\end{equation}

From points having an intensity lower than $T_{bg}$, the threshold of the hotspots is derived:

\begin{equation}
	T_{hotspots} = \mu_{bg} + 3\cdot\sigma_{bg}
\end{equation}

\begin{figure*}[!tbp]
	\centering
	\subfloat[]{\includegraphics[width=0.45\textwidth]{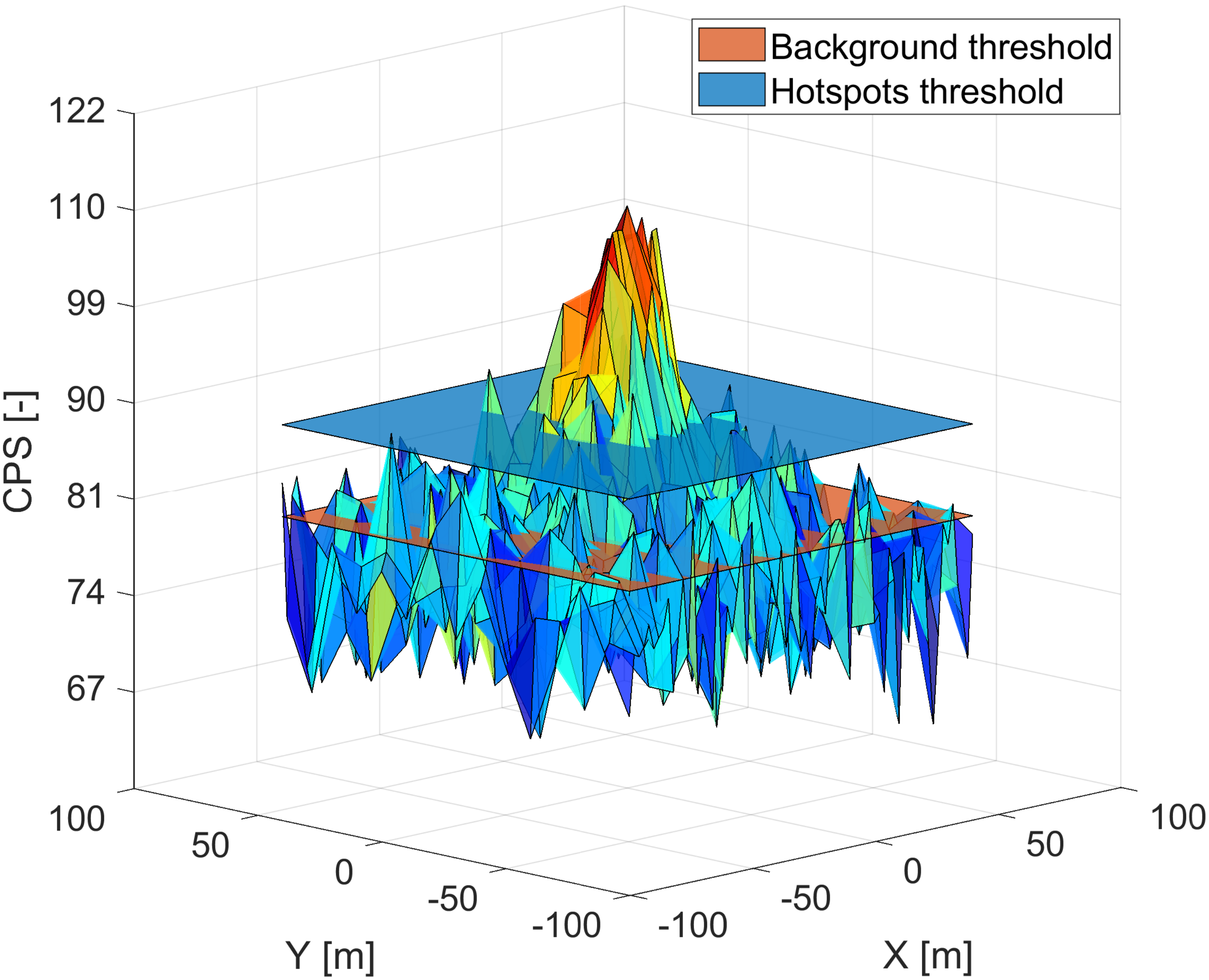}\label{fig:threshUav}} \hspace{0.05\textwidth}
	\subfloat[]{\includegraphics[width=0.45\textwidth]{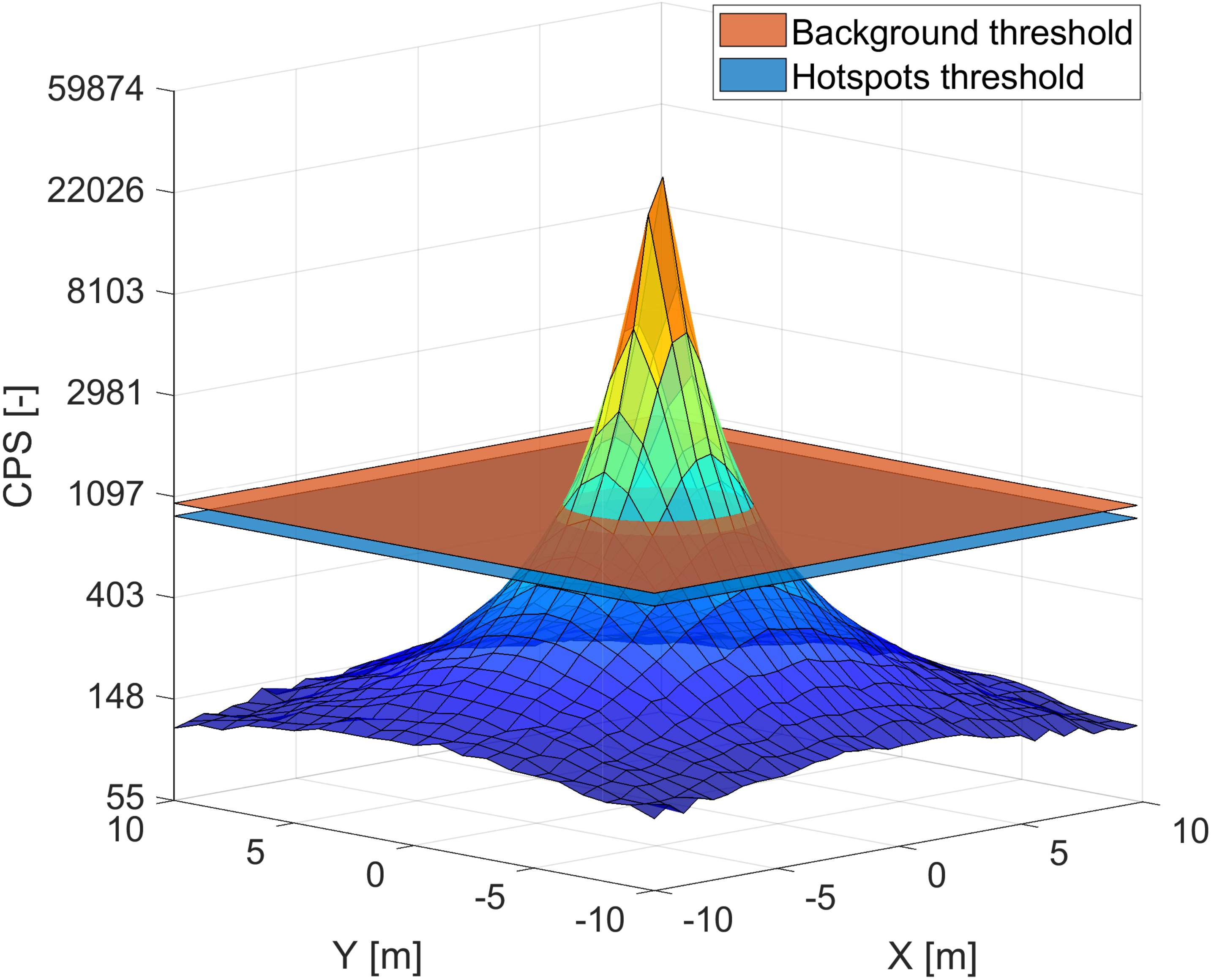}\label{fig:threshUgv}}
	\caption{An example of adaptive thresholding related to UAS-based (a) and UGV-based (b) measurement. CPS: counts per second; UAS: unmanned aircraft system; UGV: unmanned ground vehicle. [2-column figure]}
	\label{fig:thresholding}
\end{figure*}

The adaptive thresholding method was verified with both simulated \cite{gabrlik_simulation_2018} and previously measured terrestrial data. An example relating to a single source in which the emission at the distance of 1 meter reaches ten times higher than in the background is shown in Figure \ref{fig:thresholding} for both aerial and terrestrial survey.

Once the thresholding is applied to the interpolated points arranged in a regular grid, the remaining connected sets are enclosed by contours using the marching squares algorithm \cite{maple_geometric_2003}. Apparently, only contours having a certain minimal length should be accepted in order to eliminate random noise-induced peaks; we suggest that a valid contour should encircle at least four aerial samples. Finally, the regions are smoothed and optionally enlarged via the Minkowski addition \cite{lien_covering_2008} with a circle-shaped structuring element. The hotspots are eroded at first to suppress the noise; subsequently, they may be dilated again to adjust their sizes. The resulting ROIs are passed, as connected sets of points in a regular grid, on to the next stage for further processing; such a grid is then denoted as the {\it ROI map}.

%% file: text/methods-trajectory-ugv.tex
\subsection{Terrestrial Radiation Mapping}
\label{ssec:terr-rad-map}

The first task for a UGV is to move from a safe zone to the first detected contaminated area. A system user selects in the map suitable places where the robot can be potentially unloaded. This task requires the knowledge of obstacles in the area of interest. The required obstacle map is computed from the previously created DEM.

Further utilization (path planning in particular) of the obstacle map does not require a resolution as high as DEMs typically have. Satisfactory path planning accuracy can be achieved even with a reduced pixel size. Our approach towards creating the obstacle map also involves reducing the pixel count; this operation, however, is not implemented as separate downscaling. The input parameters for the obstacle map creation process are as follows:

\begin{itemize}
	\item the maximum operation angle of the mobile robot,
	\item the maximum height of a negotiable obstacle perpendicular to the terrain,
	\item the pixel size of the obstacle map.
\end{itemize}

\begin{figure}[!tbp]
	\centering
	\includegraphics[width=0.48\textwidth]{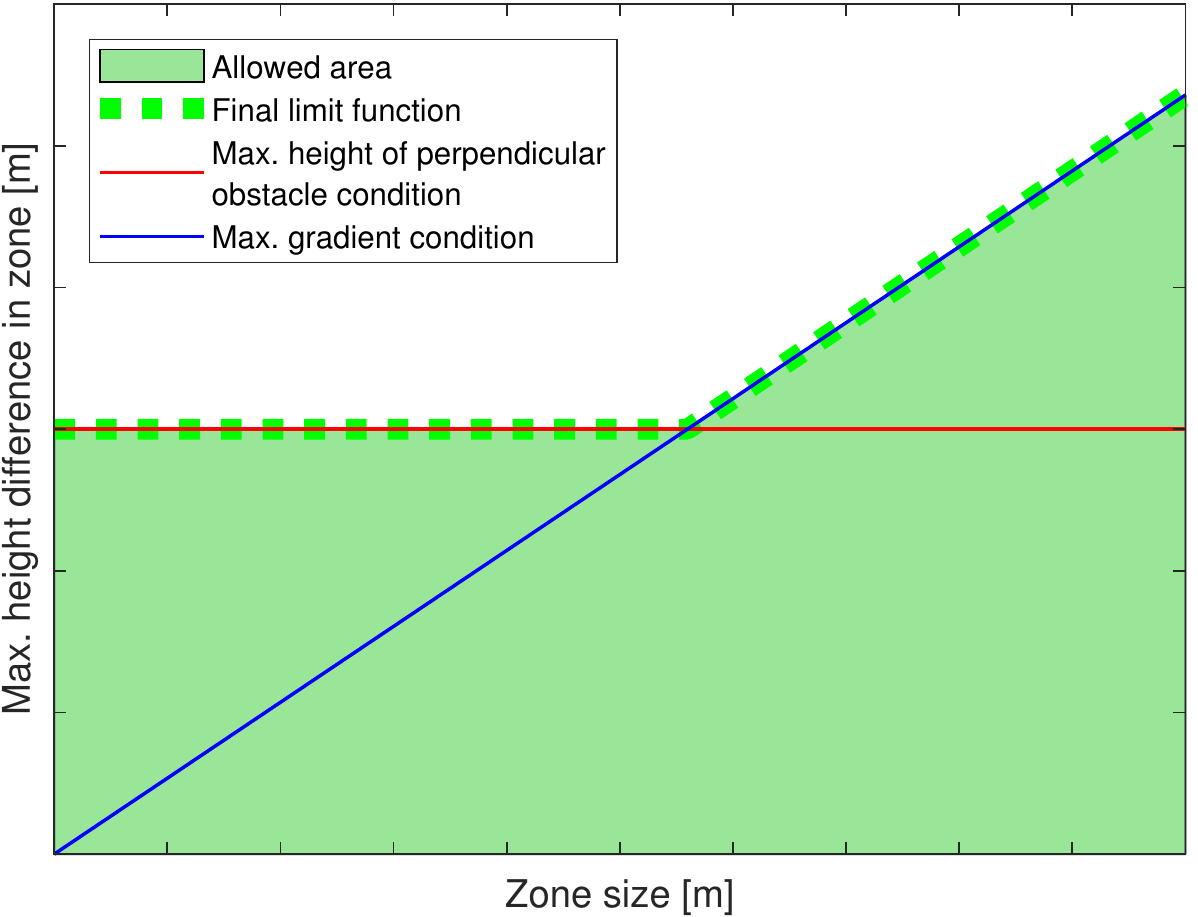}
	\caption{The obstacle function. [1-column figure]}
	\label{fig:obstacle-function}
\end{figure}

\begin{figure*}[!tbp]
	\centering
	\subfloat[]{\includegraphics[width=0.4\textwidth]{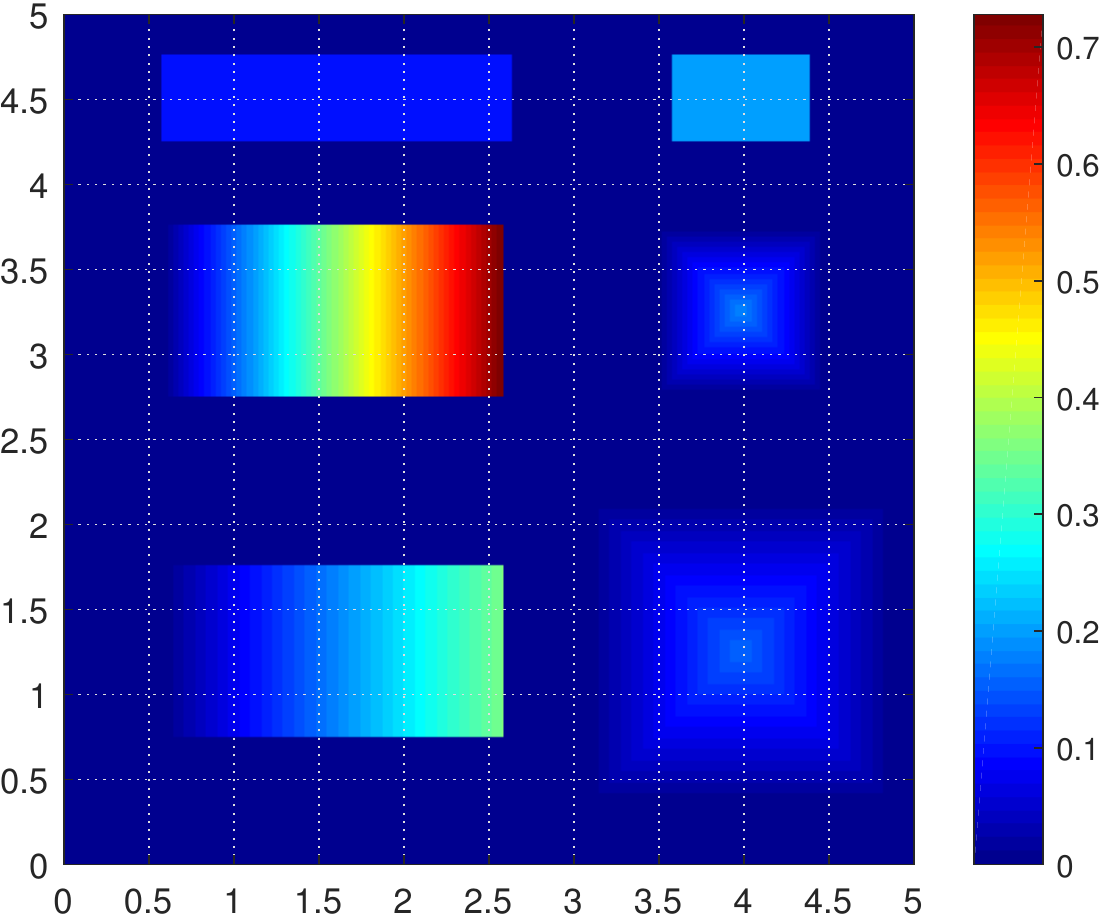}\label{fig:dem-sample}} \hspace{0.05\textwidth}
	\subfloat[]{\includegraphics[width=0.4\textwidth]{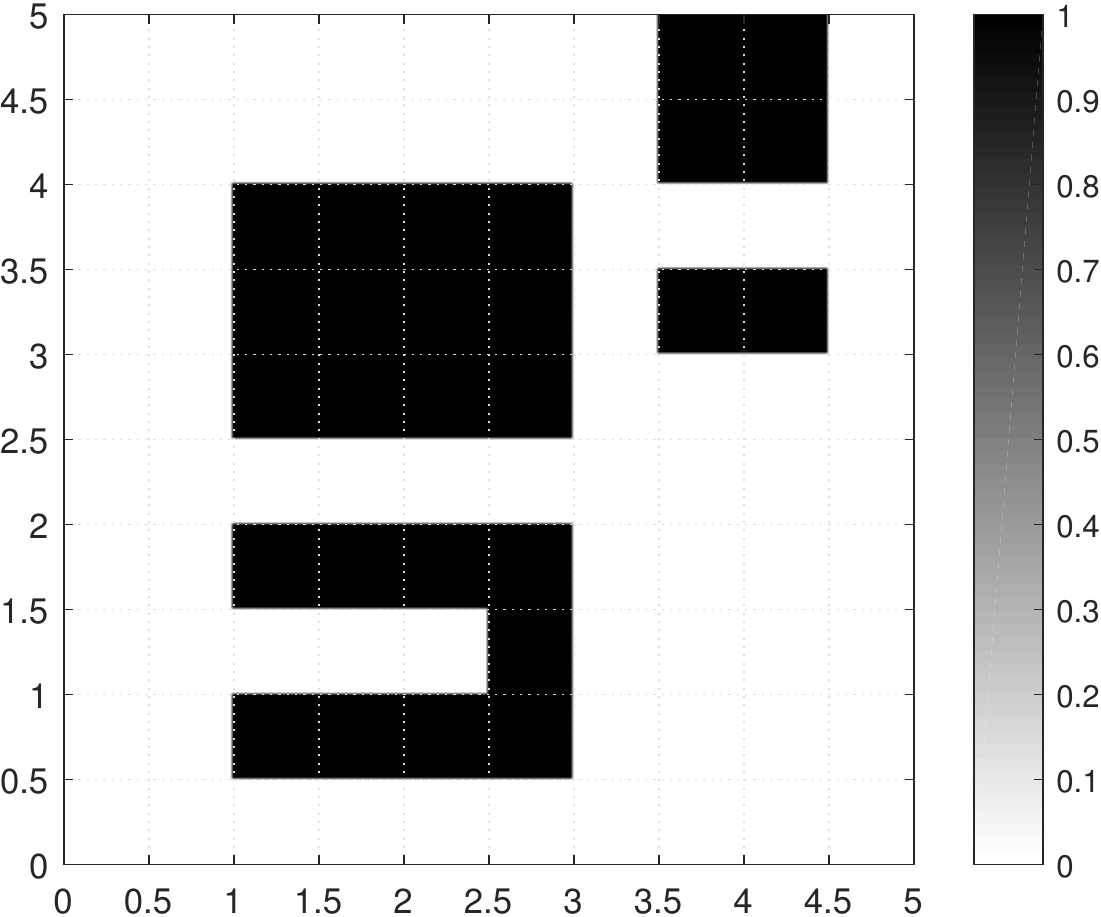}\label{fig:obstacle-map-sample}}
	\caption{A sample DEM (a) with a color scale representing the height in meters; and the resulting obstacle map (b) with a gray scale to indicate the obstacle probability. DEM: digital elevation model. [2-column figure]}
	\label{fig:sample-dem-and-obstacle-map}
\end{figure*}

From these parameters, we can define the obstacle function (Figure~\ref{fig:obstacle-function}) of the employed UGV. The function is used for detecting the impassable area in the group of DEM pixels that forms one pixel of the obstacle map. Each square sub-groups of the DEM pixels for every obstacle map pixel is checked by passing the obstacle function. In the case that a sub-group of DEM pixels is found that does not meet the obstacle function, the corresponding pixel of the obstacle map is marked as the obstacle. The process produces a binary map whose pixel size equals the integer multiple of the DEM pixel size. The principle of the operation is demonstrated for a sample DEM (Figure~\ref{fig:sample-dem-and-obstacle-map}), including terrain gradients that are below (10 deg) or above (20 deg) the robot's limit (15 deg). The sample DEM also includes perpendicular obstacles with a height of 10 or 20 cm; the robot's limit for this type of obstacles is 15 cm.

To find the optimal scenario of moving a UGV to the contaminated areas, the system operator must manually select the places where the robot can be potentially unloaded. From these starting points, we plan three types of trajectories: towards the detected contaminated areas; between these zones; and back from the last area to the unloading place. To move between the contaminated areas, the starting point for the path planning is the waypoint at the end of the trajectory inside the current area, while the final point is marked by one of the endpoints of the trajectory inside the next area. These path planning tasks can be generally solved by any A* based algorithm \cite{astar}; the shortest sequence of paths from the set of all possible solutions is used.

To plan a trajectory inside the regions of interest, we have to describe each such region with a set of polygons, one 'envelope' representing the outer limits of the area; optionally, the description can be expanded to include multiple 'holes' that characterize obstacles not traversable by the UGV. At this point, both the coarse characterization of the terrestrial-mapped hotspots (Section~\ref{sec:roiselection}) and the obstacle map are available and need to be fused. This is also the moment when the operator should intervene to validate if all of the actual obstacles are contained in the map; alternatively, the operator inserts the missing objects manually. Note that this step can utilize the earlier acquired orthophoto to identify restrictions.

\begin{figure*}[!tbp]
	\centering
	\includegraphics[width=0.85\textwidth]{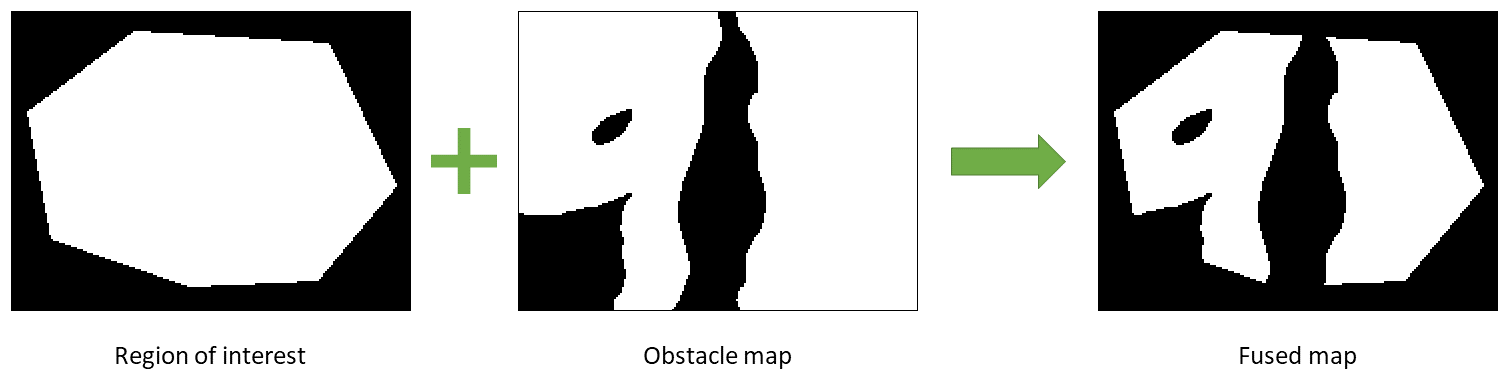}
	\caption{An example of how a fused map is generated; the white color represents the free space. [2-column figure]}
	\label{fig:mapfusion}
\end{figure*}

First, the obstacle map is resampled to exhibit the same resolution, or cell size, as the ROI map. Both maps are composed of binary value cells, which can be either empty or occupied. In the ROI map, the empty cells represent the areas where the terrestrial mapping is to be performed. The maps are fused through a relatively simple intersection: If corresponding cells in the maps are empty, then the cell is empty; conversely, it is occupied when the occupancy condition has been met in at least one of the maps. An example of the fusion producing a {\it fused map} is shown in Figure \ref{fig:mapfusion}.

Generally, a single region of interest may be divided by obstacles into multiple subregions; thus, the fused map is subjected to connected-component labeling \cite{he_fast_2009} to distinguish individual areas enclosed by the envelopes. Subsequently, each area greater than the rationally chosen threshold (the criterion being applied to exclude miniature portions of the region) is searched for contours in order to identify its envelope and holes. For practical reasons, namely, to reduce the computational requirements for the trajectory planning, the contours are reduced to polygons with a limited number of edges. The reduction is carried out incrementally, by successive removal of the least important vertices; the importance $i$ depends on the lengths of adjacent edges and the angle between them. We have

\begin{equation}
i = \left|\arccos\left(\frac{\bm{x_1}\cdot\bm{x_2}}{\|\bm{x_1}\|\cdot\|\bm{x_2}\|}\right)\right|\cdot\|\bm{x_1}\|\cdot\|\bm{x_2}\| 
\end{equation}
where $\bm{x_1}$ and $\bm{x_2}$ are the vectors of adjacent vertices.

Then, each mapped region is divided into a set of disjoint obstacle-free subregions by using the Boustrophedon decomposition \cite{choset_coverage_2000}, a procedure suitable for problems where obstacles are defined by polygons. The principle of this algorithm is to acquire subregions, or cells, that can be completely covered by a uniform 'zig-zag' trajectory; each cell has two edges parallel to the {\it sweep line}, which, in turn, is parallel to the survey direction. The result of the decomposition  depends on the selected sweep line orientation (relative to the ROI); in general terms, it is desirable to minimize the number of cells. Finally, the region is described by a graph whose nodes represent the subregions and edges define their adjacency. 

\begin{figure}[!tbp]
	\centering
	\includegraphics[width=0.45\textwidth]{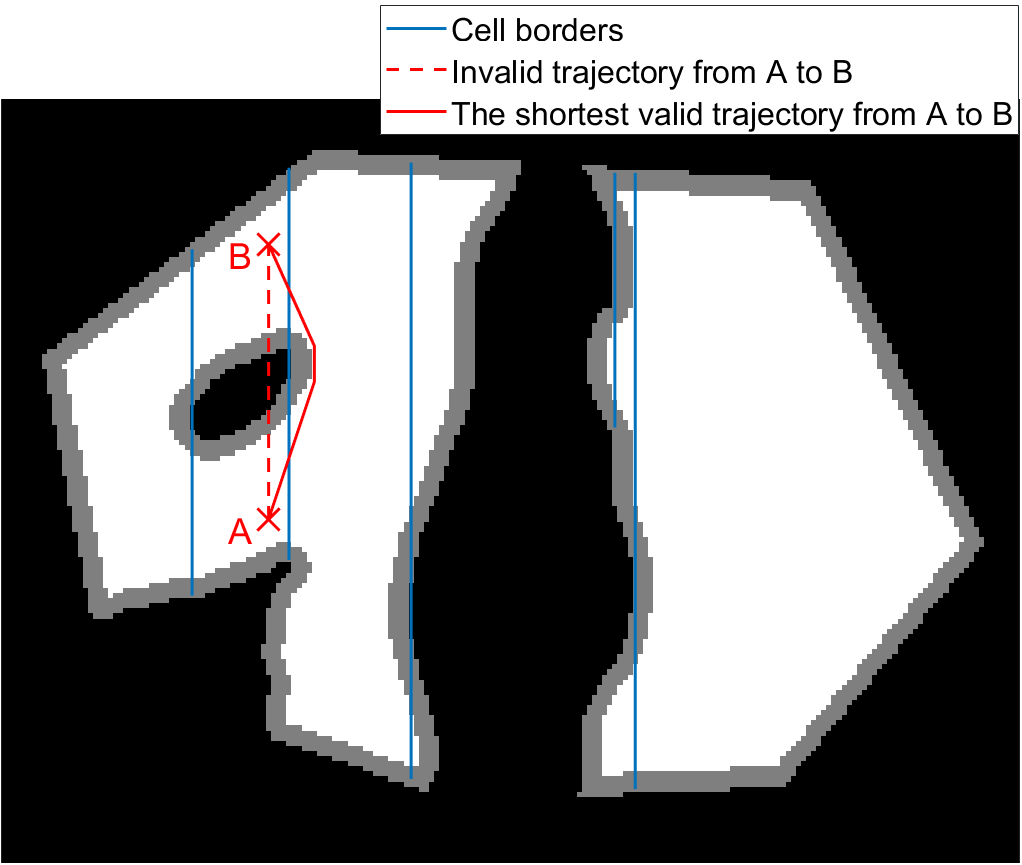}
	\caption{An example of the Boustrophedon decomposition and interconnection of subsequent cells; the black color represents the obstacles, while their dilation is in grey. [1-column figure]}
	\label{fig:cells}
\end{figure}

To determine the order in which the subregions are explored, the depth-first search algorithm is applied; this method guarantees that all nodes (cells) are visited and prefers transitions between adjacent ones. Trajectory planning inside the cells is rather straightforward, as the survey lines are parallel to the sweep line, their mutual distance remains constant, and the orientation varies in order to achieve the 'zig-zag' shape of the trajectory. In some cases, when moving from one cell to another, a direct connecting path may collide with an obstacle. Since the non-traversable zones are already described by the polygons, the visibility graph algorithm \cite{han-pang_huang_dynamic_2004} is utilized to find the shortest non-colliding path; to preserve a clearance from the obstacles, the corresponding polygons are dilated. The situation is illustrated in Figure \ref{fig:cells}.

Eventually, the whole region of interest is covered with evenly distributed parallel portions of the survey trajectory, along with connecting paths between consecutive subregions, and the obstacles are avoided.

%% file: text/methods-sources-localization.tex
\subsection{Radiation Data Processing and Source Localization}
\label{ssec:source-localization}

Handling the terrestrial data is largely similar to the aerial radiation data processing presented in Section \ref{sec:roiselection}. The localization of the sources can be characterized by three steps:
\begin{enumerate}
	\item Estimating the number of sources, $R$.
	\item Estimating the initial coarse parameters of $R$ sources.
	\item Increasing the accuracy of the parameters in accordance with the measured data.
\end{enumerate}

The first step utilizes the adaptive thresholding algorithm. Although multiple sources in a single region form a sole hotspot within the primary map, they may yield more peaks inside the detailed secondary map built from the UGV data, which are acquired in a finer grid and from a closer distance than the aerial dataset. To perform the estimation, the following steps are applied:
\begin{enumerate}[label=1.\alph*]
	\item Compute the peak threshold.
	\item Interpolate the data into a regular grid.
	\item Eliminate the radiation background.
	\item Find valid contours in the map; their count equals the number of sources.
\end{enumerate}

Regarding the source parameters, three items are sought for each source; these items include the emission intensity and coordinates in two axes. Let us have a source~$i$ with the vector $\bm{\theta}_i = (\alpha_i, x_i, y_i)$; all of the sources are then characterized by the parameter matrix $\bm{\theta} = (\bm{\theta}_1, \bm{\theta}_2, \ldots, \bm{\theta}_R)^{\intercal}$. To initiate the matrix, we suggest choosing a central point within each contour to define the coordinates and taking the greatest corresponding total count value to estimate the intensity. By filling in the matrix, the second localization step is completed.

Finally, the accuracy of the parameters is iteratively improved via the Gauss-Newton (G-N) method \cite{deuflhard_least_2011}, which finds use in solving non-linear least squares problems. Given a matrix of $M$ measurements, $\bm{z} = (\bm{z}_1, \bm{z}_2, \ldots, \bm{z}_M)^{\intercal}$, where $\bm{z}_i = (c_i, x_i, y_i)$ to denote the total count obtained and the coordinates where the measurement has been taken, the G-N algorithm minimizes the sum of residuals (the differences between the expected and the measured values); the residual $m$ is expressed as:

\begin{equation}
r_m = c_m - \sum_{r=1}^R\frac{\alpha_r}{(x_m-x_r)^2 + (y_m-y_r)^2 + h^2},
\end{equation}
where $h$ is the height of the detectors above the terrain. The parameter matrix is updated in each step according to the equation
\begin{equation}
\bm{\theta}_{k+1} = \bm{\theta}_k - (\bm{J}^\intercal\bm{J})^{-1}\bm{J}^\intercal\bm{r}(\bm{\theta}_k),
\end{equation}
where $\bm{J}$ is the $M \times 3R$ Jacobi matrix of the partial derivatives of the residuals. The iterations continue until the sum of the squared residuals stops  decreasing significantly.

%% file: text/results-photogrammetry.tex
\subsection{Aerial Photogrammetry}
\label{ssec:res-aerial-photo}

\begin{figure}[!tbp]
	\centering
	\includegraphics[width=0.45\textwidth]{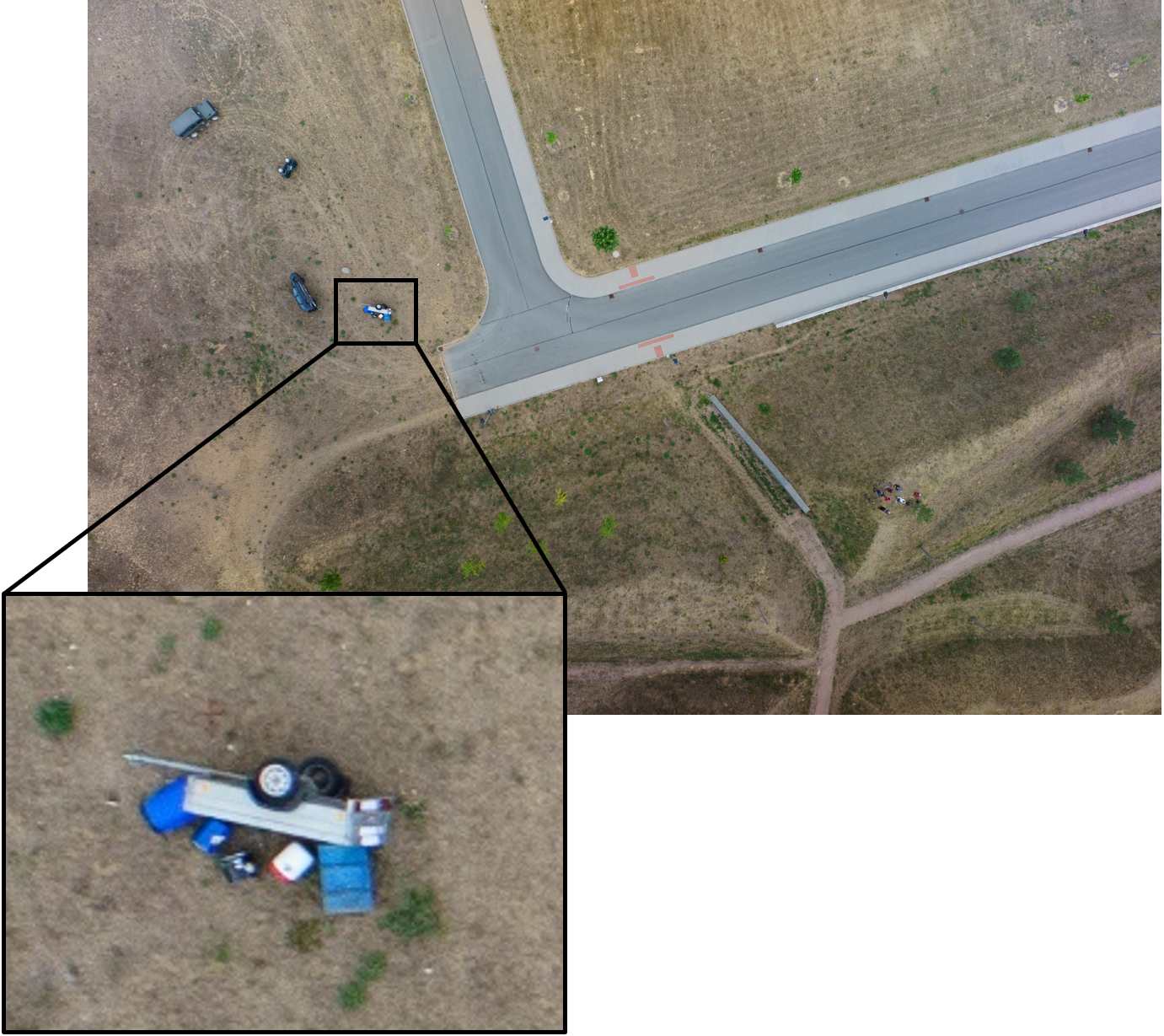}
	\caption{A sample image captured at 60~m AGL during the initial flight. The detail shows the car accident simulated in zone 2. AGL: above ground level. [1-column figure]}
	\label{fig:aerial-image}
\end{figure}

The data collected during the first UAS flight were processed immediately after landing, as the photogrammetry products are essential for the subsequent mapping phases. From the total amount of 211 images, we utilized only 124 items; those captured during the take-off and landing procedures were excluded because they are not relevant to the actual processing. As expected from the designed trajectory, the automatic waypoint-based flight lasted approximately 10~minutes and was 1~kilometer long. One of the images, together with the maximum zoom available, is shown in Figure~\ref{fig:aerial-image}.

\begin{figure}[!tbp]
	\centering
	\includegraphics[width=0.5\textwidth]{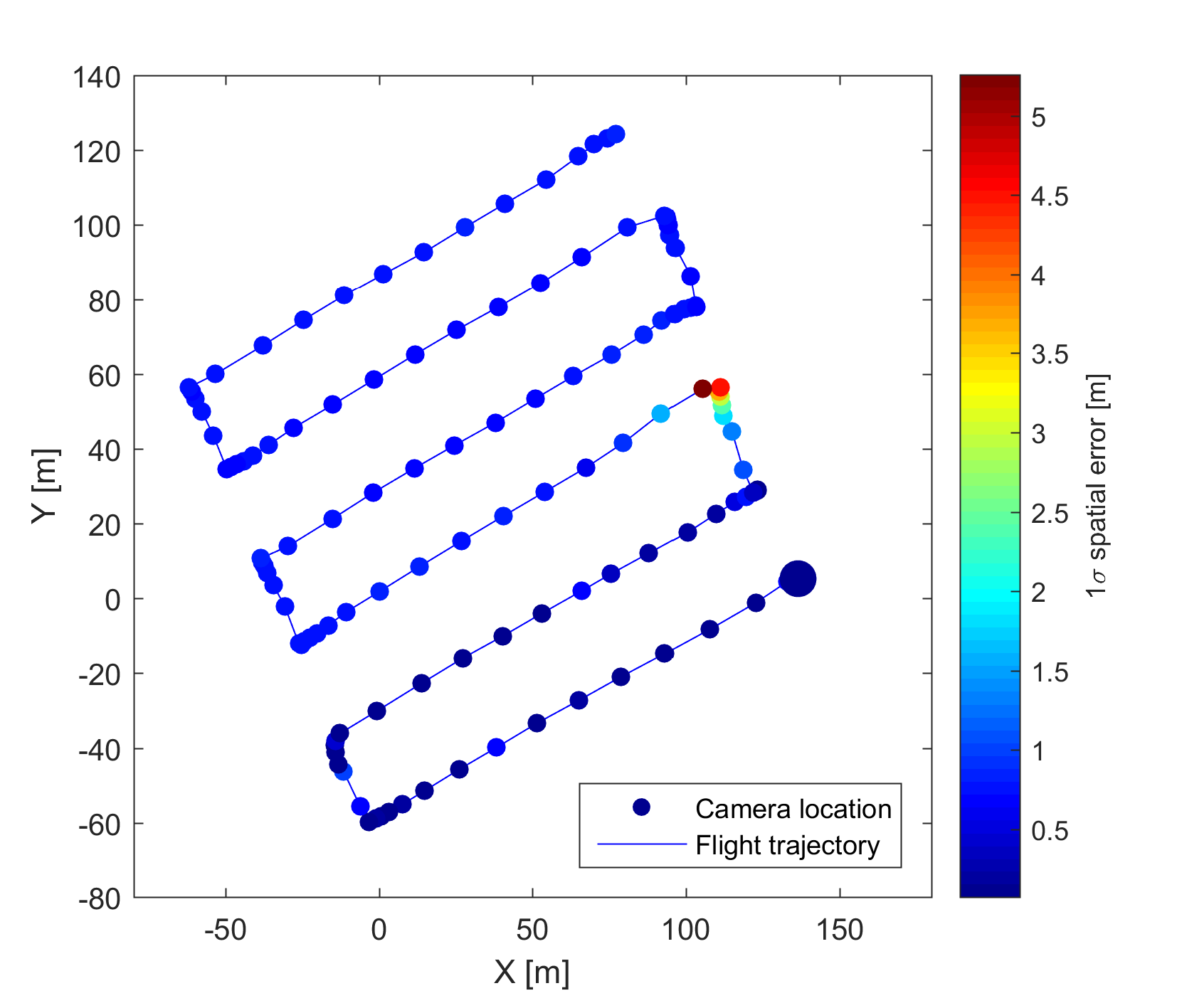}
	\caption{The flight trajectory and the INS spatial error estimation at the camera locations, with the starting point highlighted. The zero coordinates correspond to the position of the base station. INS: inertial navigation system. [1-column figure]}
	\label{fig:rtk-quality}
\end{figure}

The photographs were directly georeferenced by using the onboard multi-sensor system, making the exterior orientation parameters accessible immediately after the flight, regardless of the GCPs' availability. In addition to the position and orientation, the system offers various indicators facilitating analyses of its operation quality. Figure~\ref{fig:rtk-quality} summarizes the estimation of the GNSS-aided INS 1$\sigma$ spatial error at image taking. Based on the presented data, the average and maximum errors equalled 0.74~m and 5.3~m, respectively. The reason for the conspicuous accuracy decrease consisted in the GNSS' RTK fix solution outages caused by insufficient quality of the signal necessary for the carrier-phase tracking. At one moment, the INS excluded the inaccurate GNSS-produced aiding data from the computation, and the position error increased sharply to up to five meters.

\begin{figure*}[!tbp]
	\centering
	\subfloat[]{\includegraphics[width=0.4\textwidth]{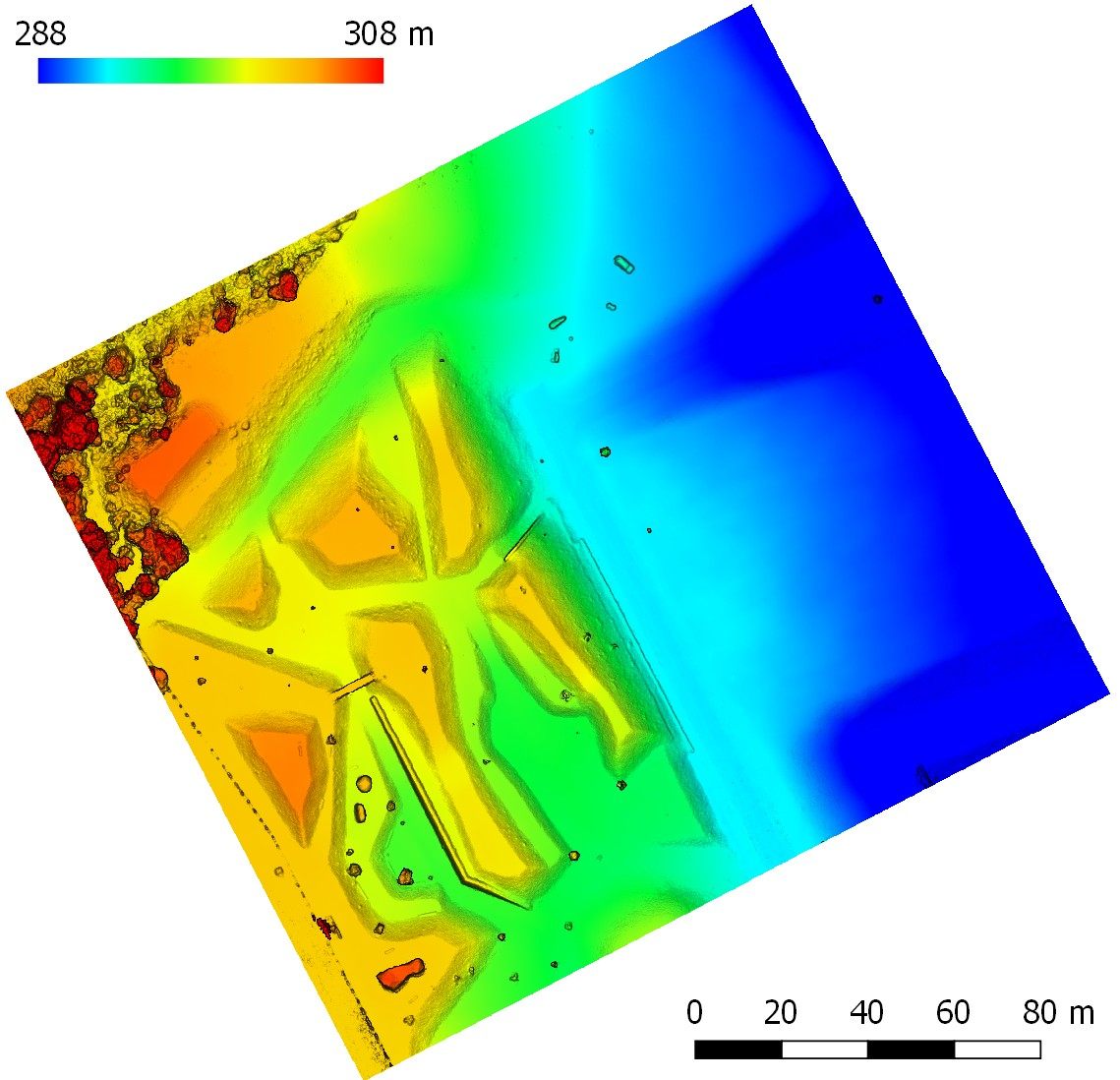}\label{fig:dem}} \hspace{0.05\textwidth}
	\subfloat[]{\includegraphics[width=0.4\textwidth]{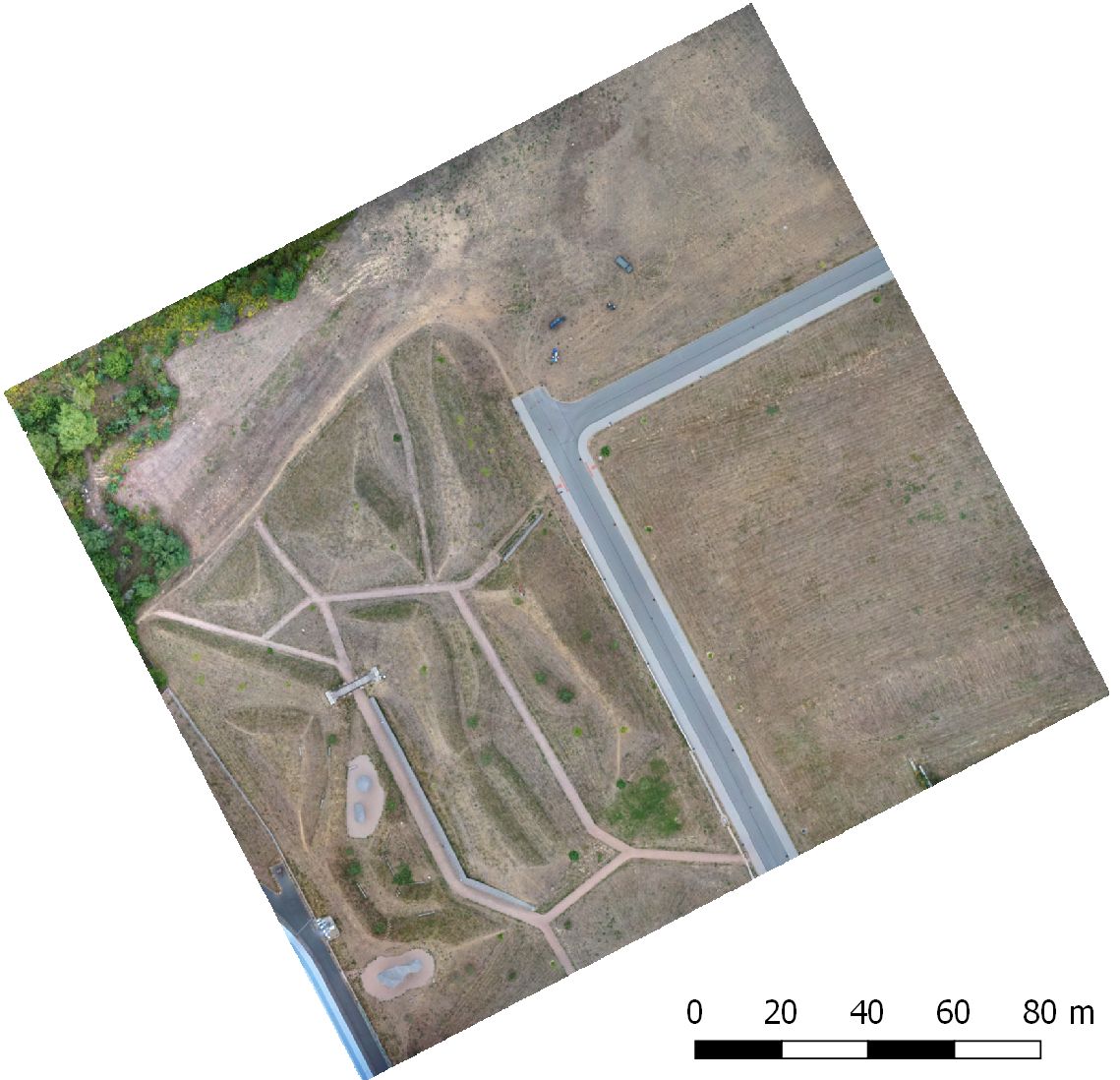}\label{fig:ortho}}
	\caption{The UAS photogrammetry-based DEM (indicating the spectral color-scaled elevation and black-marked slopes) (a) and orthophoto (b). DEM: digital elevation model [2-column figure]}
	\label{fig:photo-products}
\end{figure*}

The dataset was processed in Agisoft Photoscan at a low quality in order to reduce the processing time. We used images downscaled by a factor of four to perform the alignment stage, and 16-fold downscaled images were employed in the dense point cloud generation to yield 7.6~million points ($\sim$200~points/m$^2$). The digital elevation model constructed from the point cloud delivered the resolution of 7.4~cm/pix; the subsequent product, namely, the orthophoto, exhibited 1.9~cm/pix (Figure~\ref{fig:dem} and \ref{fig:ortho}). The entire photogrammetry processing lasted approximately 45~minutes.

%

To determine the georeferencing quality, namely, to assess the method, we utilized six GCPs (Figure~\ref{fig:uas-traj-photo-plan}). Despite the issues with the GNSS/INS, we obtained the georeferencing root mean square error (RMSE) values of 0.55, 0.34, and 1.13~m for the latitude, longitude, and altitude, respectively; the errors in the individual targets were computed in Photoscan as the distances between the measured and the estimated positions. The resulting accuracy is below expectations and does not correspond to the capabilities of the system; however, the level remains acceptable for the subsequent mapping phases.

%% file: text/results-radiation-uas.tex
\subsection{Aerial Radiation Mapping}

\begin{figure*}[!tbp]
	\centering
	\includegraphics[width=0.85\textwidth]{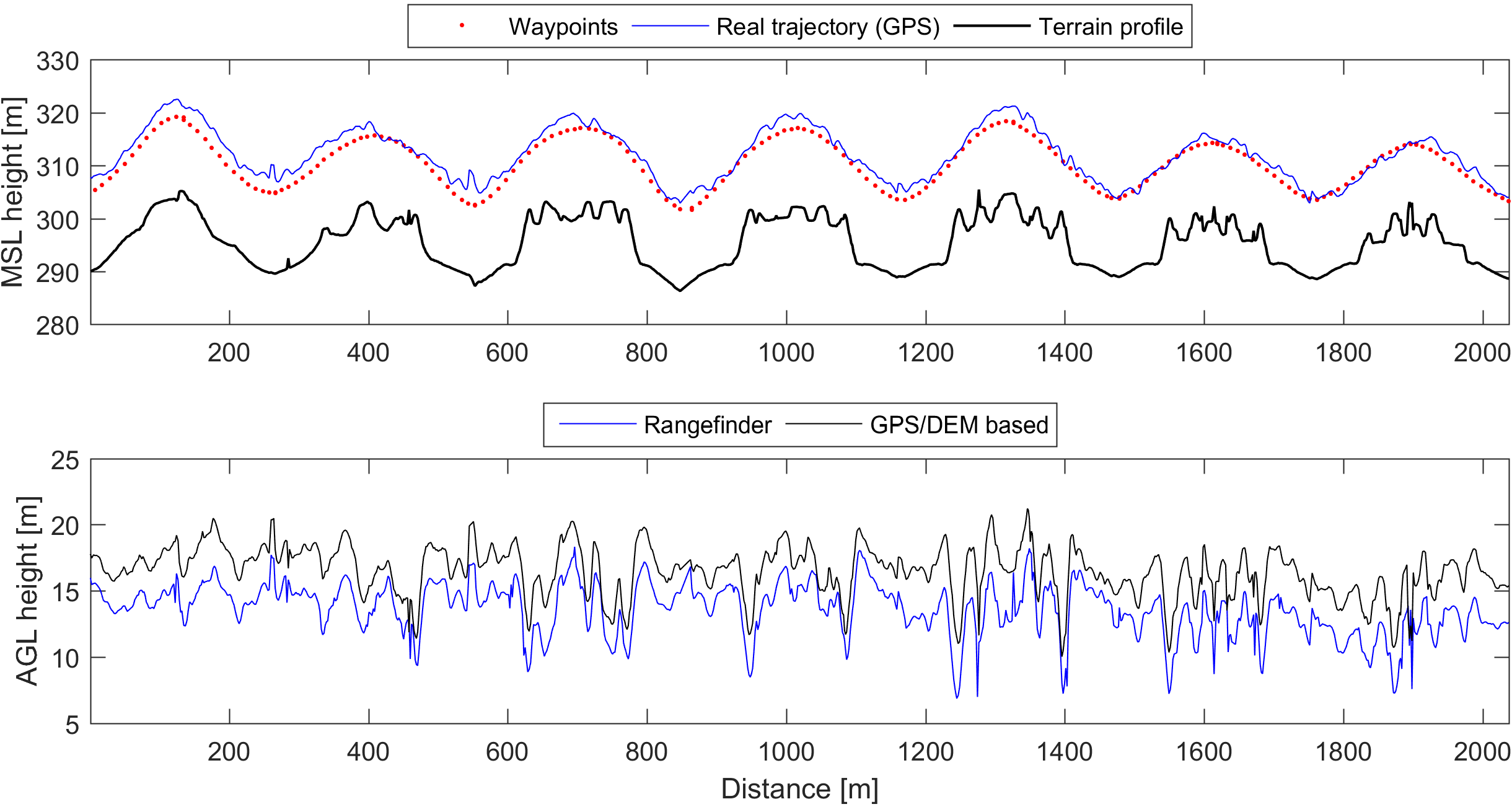}
	\caption{The vertical profile of the UAS trajectory during the radiation-mapping phase, completed with the trajectory waypoints and underlying terrain (upper graph). The AGL flight height recorded by the rangefinger, represented together with the related GPS/DEM-based estimation (bottom graph). MSL: mean sea level; AGL: above ground level; DEM: digital elevation model; GPS: global positioning system. [2-column figure]}
	\label{fig:rad-traj-profile}
\end{figure*}

Using the algorithm described in section~\ref{ssec:aerial-rad-map}, and utilizing the DEM obtained within the previous step, we automatically adjusted the vertical components of the trajectory for the aerial radiation mapping. At the inital stage, the trajectory (Figure~\ref{fig:uas-traj-rad-plan}) was split into 10~m segments, resulting in 209 waypoints; subsequently, the MSL height for every waypoint was computed (the red points in Figure~\ref{fig:rad-traj-profile}).

Compared to the photogrammetry flight, the second UAS mission lasted longer (approximately 20~minutes) due to the trajectory length and lower operating speed. Thanks to the GPS receiver and laser rangefinder, both integrated in the onboard radiation detection system, we can analyze the 'terrain-following' algorithm performance. The upper part of Figure~\ref{fig:rad-traj-profile} displays the vertical coordinate of the actual GPS trajectory and the underlying terrain; the bottom graph shows the height above ground level. The presented data refer to the UAS flown at a relatively constant distance from the surface; at some moments, however, deviation from the desired value of 15~m is obvious. In this context, the rangefinder reports the height of 13.7~m RMSE, while the GPS-DEM derived value (GPS height minus surface height) is slightly higher, reaching 16.6~m RMSE. It should be noted that none of the sources is accurate enough for detailed assessment: The distance value measured by the rangefinder is biased due to the UAS tilting, and both the GPS and the DEM errors lie within the order of meters. Despite these facts, the presented data clearly indicate that the terrain following method has met the expectations, allowing us to collect radiation data at a sufficiently constant distance from the ground, regardless of the surface shape.

\begin{figure*}[!tbp]
	\centering
	\subfloat[]{\includegraphics[width=0.4\textwidth]{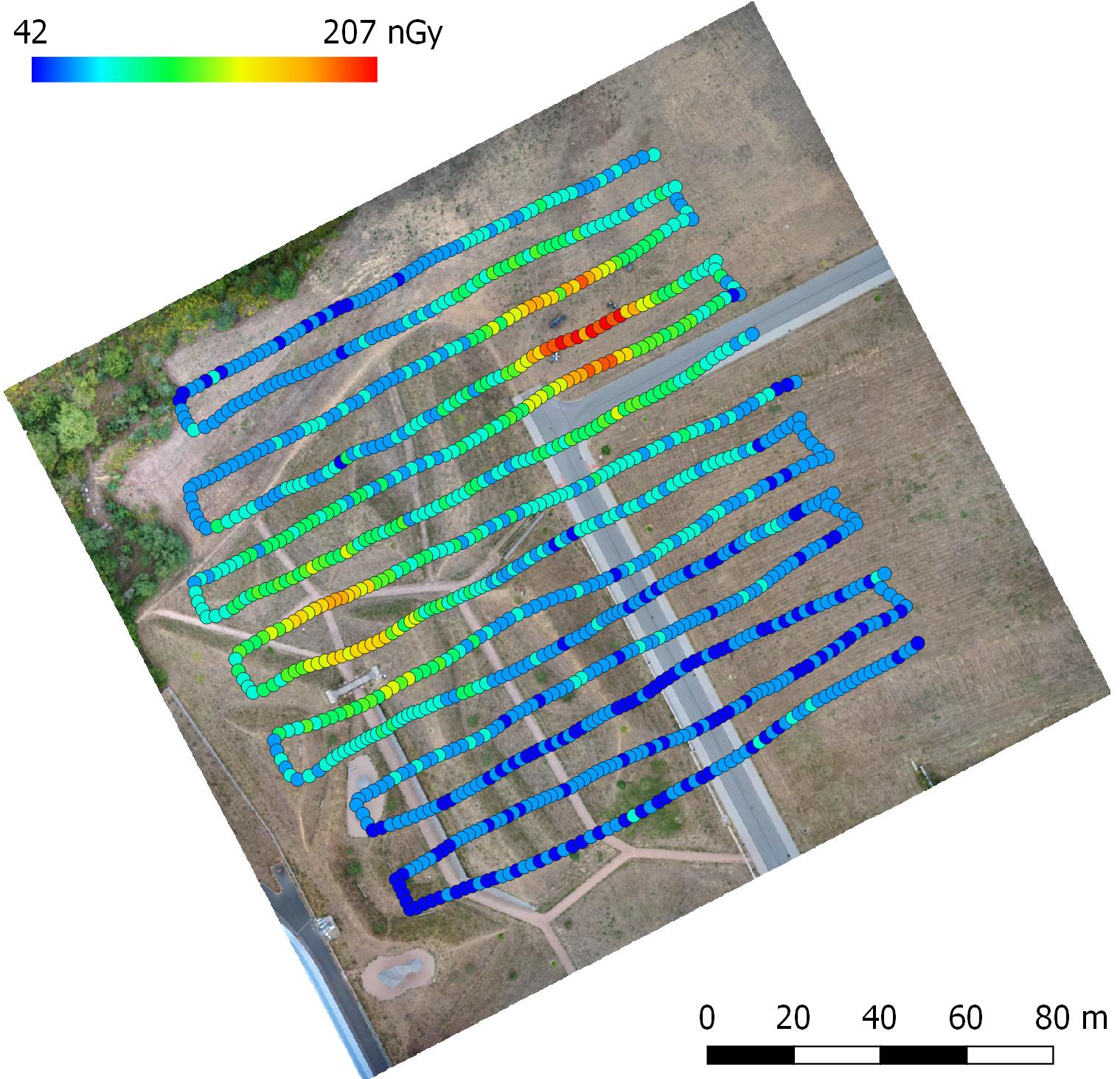}\label{fig:rad-uas-scatter}} \hspace{0.05\textwidth}
	\subfloat[]{\includegraphics[width=0.4\textwidth]{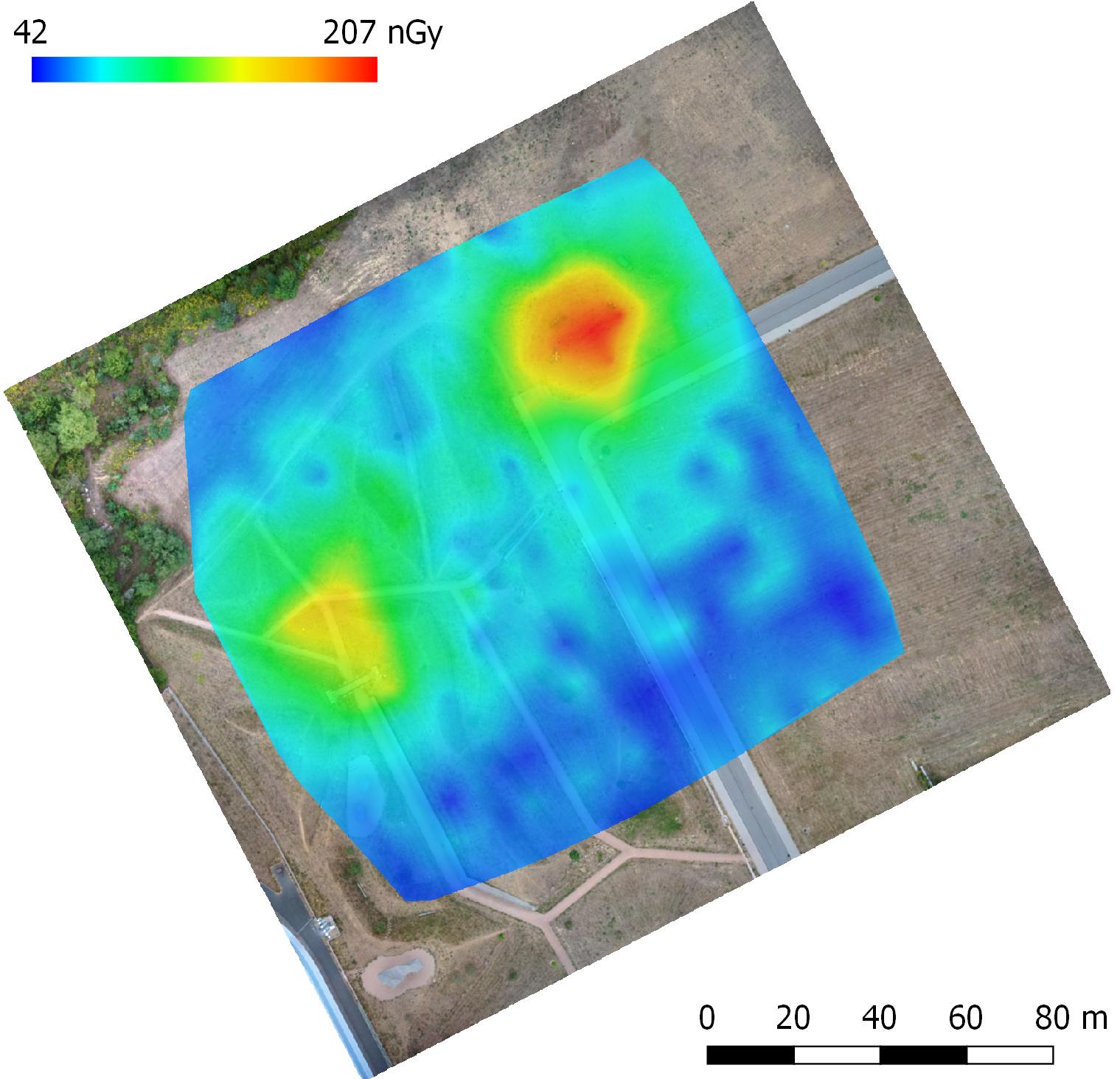}\label{fig:rad-uas-interp}}
	\caption{The dose rates obtained during the UAS-based radiation mapping procedure (a); the acquired data were downsampled and interpolated for the subsequent processing (b). UAS: unmanned aircraft system. [2-column figure]}
	\label{fig:rad-uas}
\end{figure*}

Given the one-second sampling period of the radiation detection system, the dataset comprises 1,100 measurements (excluding the data gathered during the take-off and landing procedures), with the minimum and maximum values of 0.042 and 0.207~\textmu Gy$\cdot$h$^{-1}$, respectively. Note that the mean radiation background intensity approximately equalled 0.07~\textmu Gy$\cdot$h$^{-1}$. Considering the horizontal coordinates of the individual measurements (Figure~\ref{fig:rad-uas-scatter}), we can clearly identify two separated hotspots, namely, areas with increased radiation intensity. As already discussed within section~\ref{sec:roiselection}, the scattered data had to be downsampled prior to the actual processing in order to achieve comparable data density in both axes, parallel and perpendicular to the flight direction. The processed and interpolated data presented in Figure~\ref{fig:rad-uas-interp} clearly define the situation at the location and, above all, facilitate automatic selection of the ROI (section~\ref{ssec:roi-sel}). The radiation data were interpolated to a 0.1 m regular grid.


The collected data contains, in addition to the dose rate values, also raw data allowing spectral analysis and radionuclide identification; this step, however, is not necessary for the subsequent stage (hotspot localization) and was thus not performed during the experiment. The spectral analysis potential is outlined in section~\ref{sec:discussion}.

%% file: text/results-polygon.tex
\subsection{Areas Selected for the Terrestrial Mapping}
\label{ssec:roi-sel}


\begin{figure*}[!tbp]
	\centering
	\subfloat[]{\includegraphics[width=0.45\textwidth]{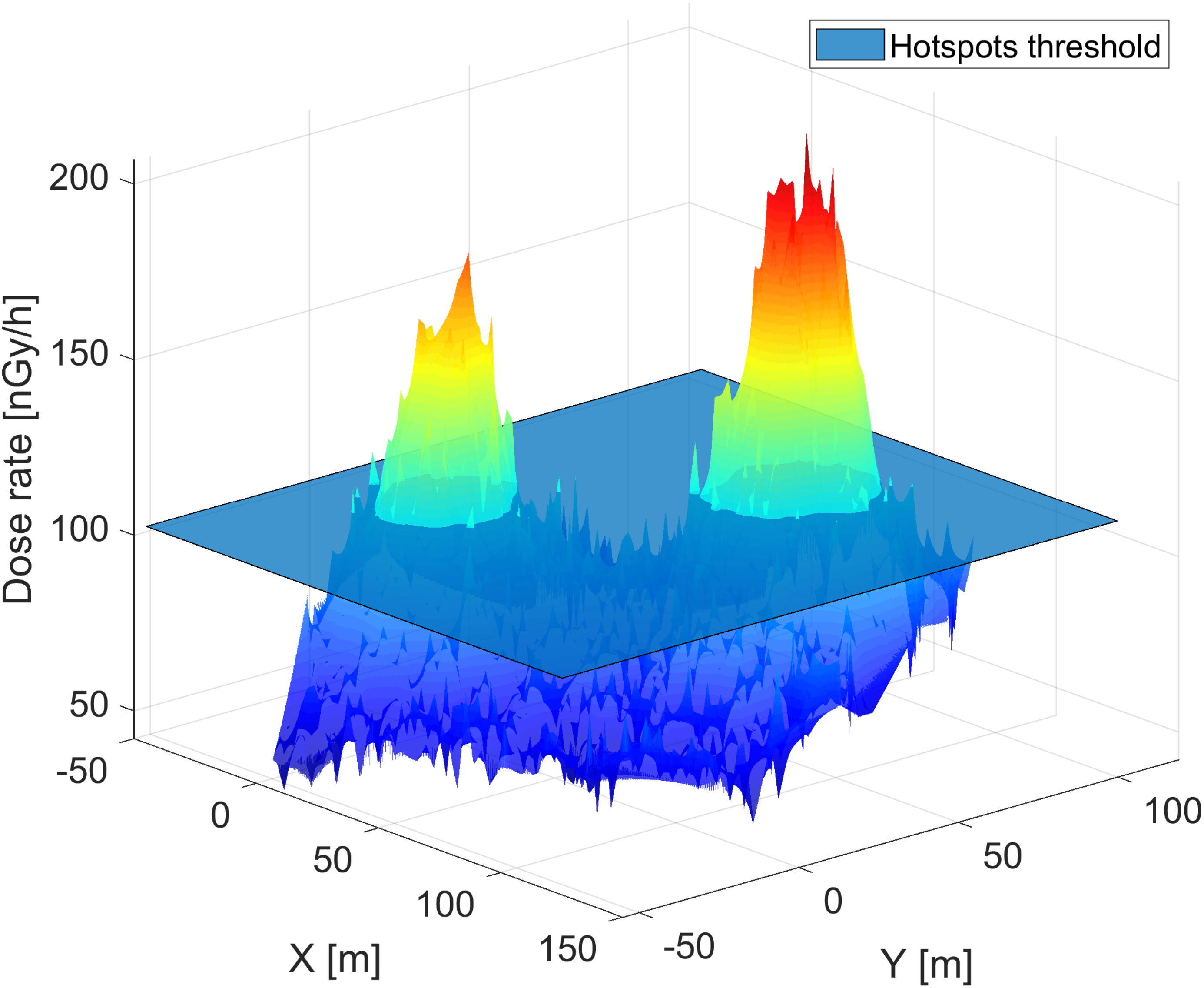}\label{fig:hotspots3d}} \hspace{0.05\textwidth}
	\subfloat[]{\includegraphics[width=0.45\textwidth]{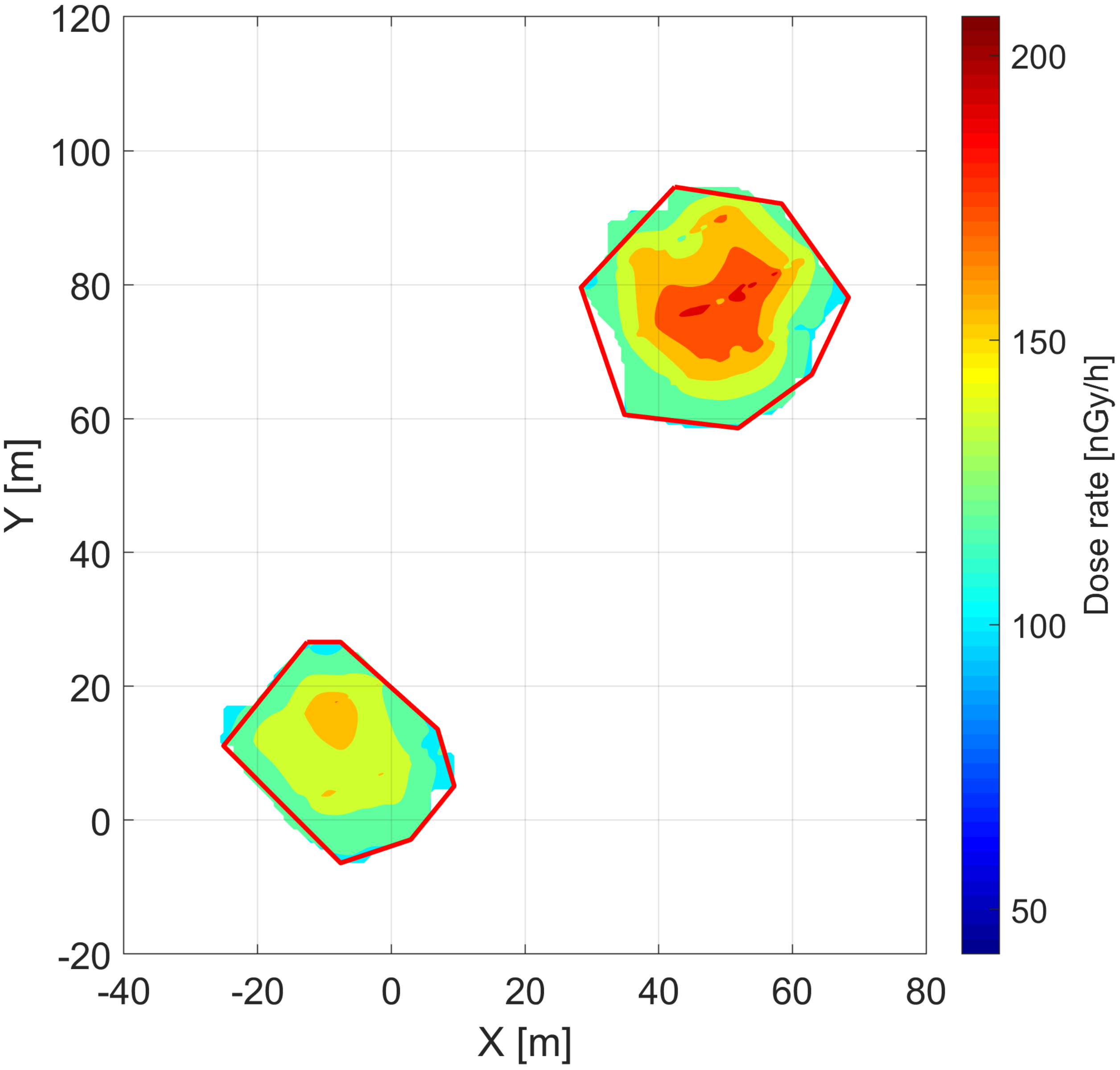}\label{fig:rois}}
	\caption{The adaptive thresholding applied to the aerial radiation data (a), and polygonal approximation of the hotspots (b). [2-column figure]}
	\label{fig:areaselection}
\end{figure*}

The interpolated radiation map has been subjected to the automatic ROI selection algorithm. First, the background threshold was computed, equaling 0.090~\textmu Gy$\cdot$h$^{-1}$; such a result is in good accordance with the actual background intensity, which reached up to about 0.095~\textmu Gy$\cdot$h$^{-1}$. Subsequently, the script was able to determine the hotspot separation threshold, attaining 0.103~\textmu Gy$\cdot$h$^{-1}$. A 3D visualization of the thresholding process is shown in Figure~\ref{fig:hotspots3d}. Note the small 'spikes' around the two major radiation intensity peaks, induced by the measurement noise. In order to eliminate these spikes and to smoothen the region's borders slightly, the imprint of the hotspots was morphologically eroded by a structuring element of a size corresponding to 3 meters. Finally, both of the remaining regions were roughly approximated by polygons with 7 vertices (Figure~\ref{fig:rois}).

%% file: text/results-trajectory-ugv.tex
\subsection{Terrestrial Radiation Mapping}

\begin{figure}[!tbp]
	\centering
	\includegraphics[width=0.45\textwidth]{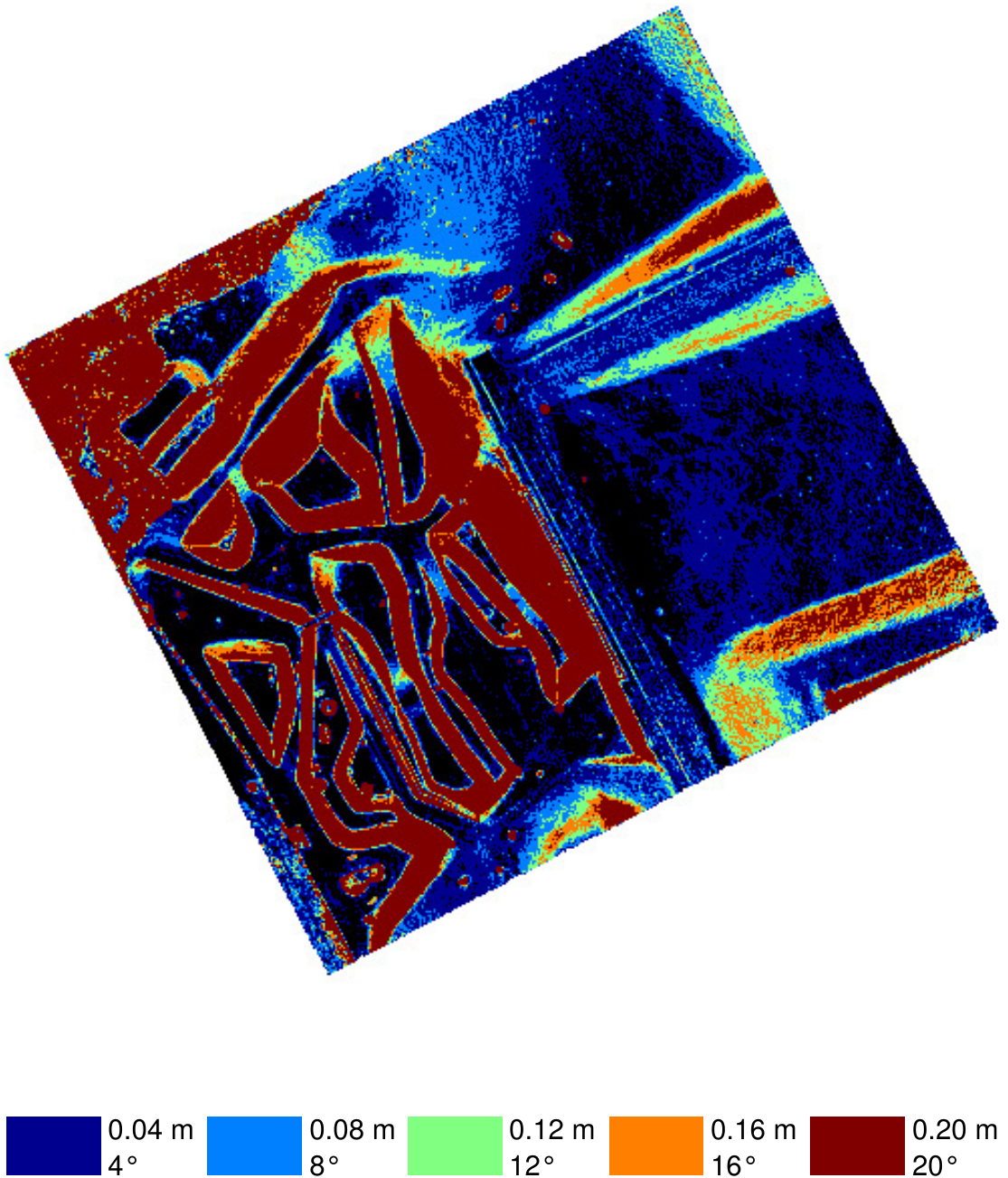}
	\caption{The obstacle maps computed for the different UGV limits. UGV: unmanned ground vehicle [1-column figure]}
	\label{fig:ugv-obstacle-map-layered}
\end{figure}

The obstacle map for the path planning is computed from all pixels of the source DEM. Generally, the size of a DEM pixel (74 mm) is too small to be applied in the obstacle map suited for outdoor path planning tasks; in this context, we can point out that larger pixels reduce the time required for the subsequent processing operations. At such stages, the pixel size approaches the actual width of the employed UGV. To carry out the planned mission, we selected the value of 0.518 m, namely, the integer multiple of the DEM pixel size. The resulting obstacle maps (Figure~\ref{fig:ugv-obstacle-map-layered}) computed for five different terrain limits show the terrain negotiability differences. When in the automatic navigation mode, our UGV can safely pass a terrain characterized by a gradient of 16 degrees or surmount obstacles having 0.16 m; if operated manually, however, the vehicle is capable of managing 20 degrees and 0.2 m. The results described below are based on the obstacle map computed for the 0.16 m and 16 deg limits (the orange layer in Figure~\ref{fig:ugv-obstacle-map-layered}).

\begin{figure}[!tbp]
	\centering
	\includegraphics[width=0.45\textwidth]{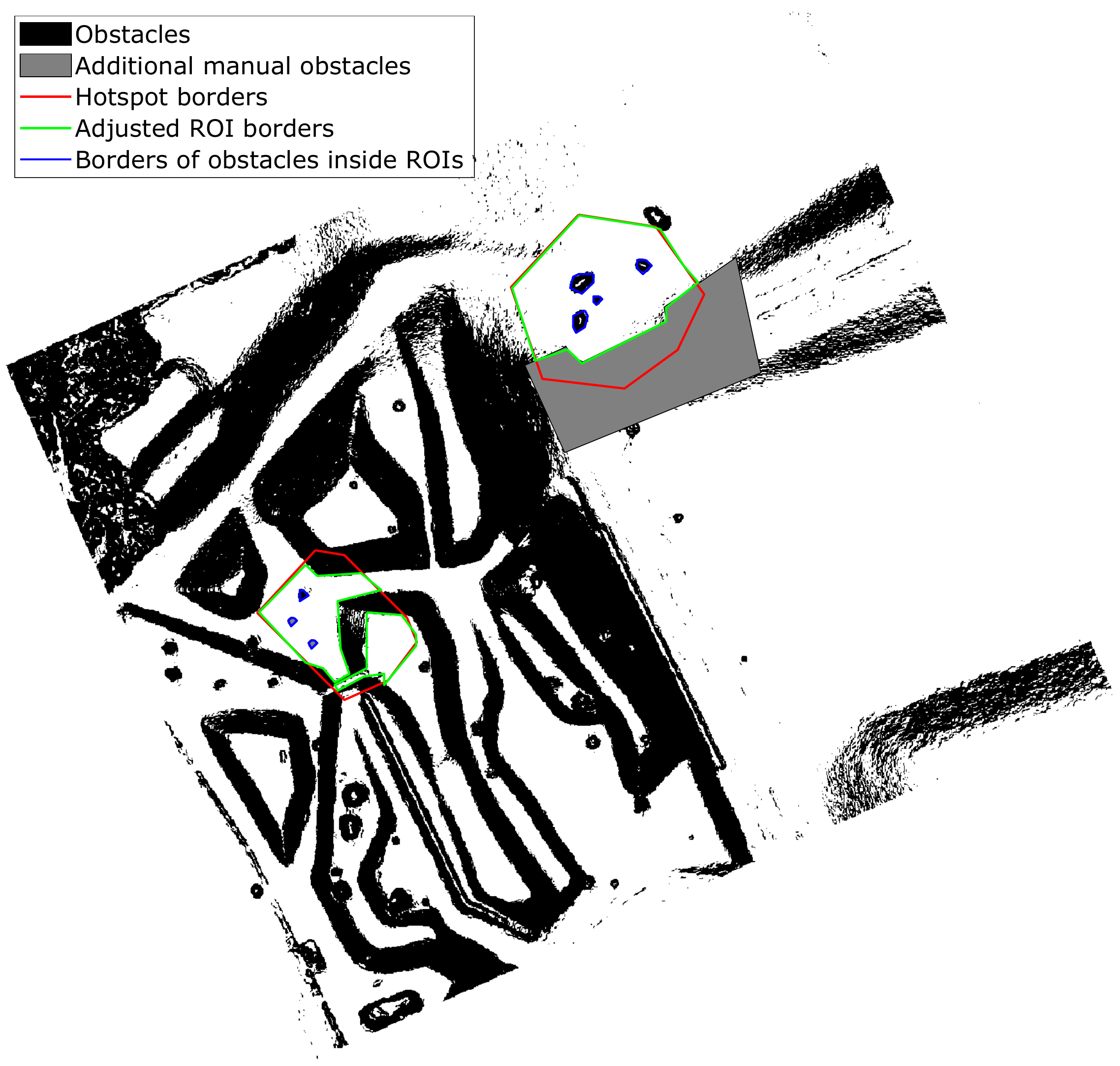}
	\caption{The adjustment of the regions of interest via the obstacle map. [1-column figure]}
	\label{fig:roi-adjust}
\end{figure}

Subsequently, the obstacle map is fused with the hotspot polygons. Prior to the processing by the automatic script, several additional obstacles, in particular a curb and plants that formed a new boundary limiting the southern side of the upper-right ROI (corresponding to zone 1), had to be defined manually. Moreover, two minor obstacles, namely, a small barrel and the remains of a tree, were added inside the lower-left ROI (zone 2). In Figure~\ref{fig:roi-adjust}, these adjustments are marked in gray. The rough hotspot borders, modified in accordance with the obstacle map, form the 'envelopes' of the regions to be mapped and are visualized as the green polygons; the blue polygons inside the green ones then represent the 'holes' to be avoided. Note that the algorithm yielded two distinct subregions within the lower-left ROI; one of these areas, however, is inaccessible to the UGV (as can be proved via the path-planning algorithm) and will not be further examined within the article.

\begin{figure*}[!tbp]
	\centering
	\subfloat[]{\includegraphics[width=0.45\textwidth]{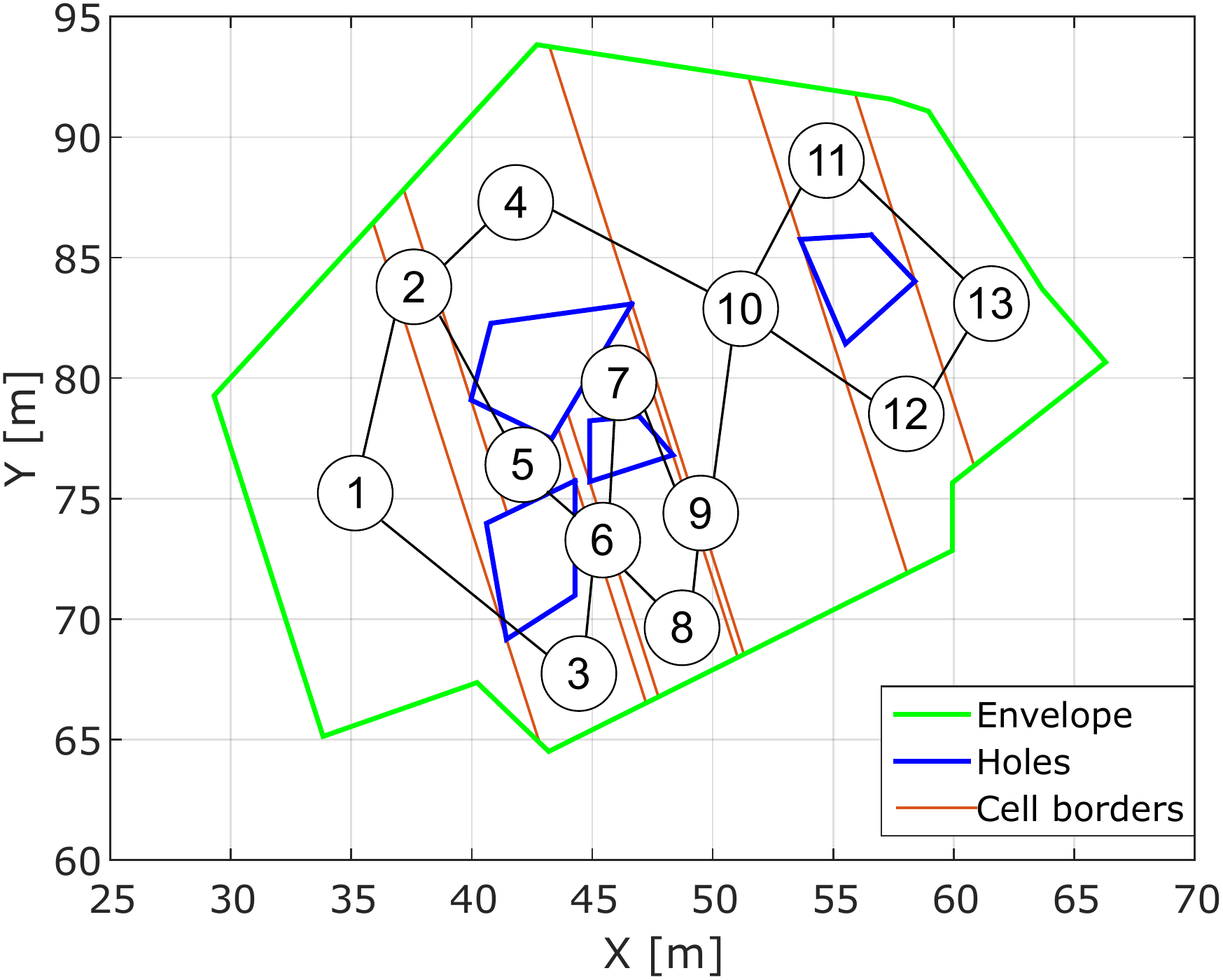}\label{fig:decomposition1}} \hspace{0.05\textwidth}
	\subfloat[]{\includegraphics[width=0.45\textwidth]{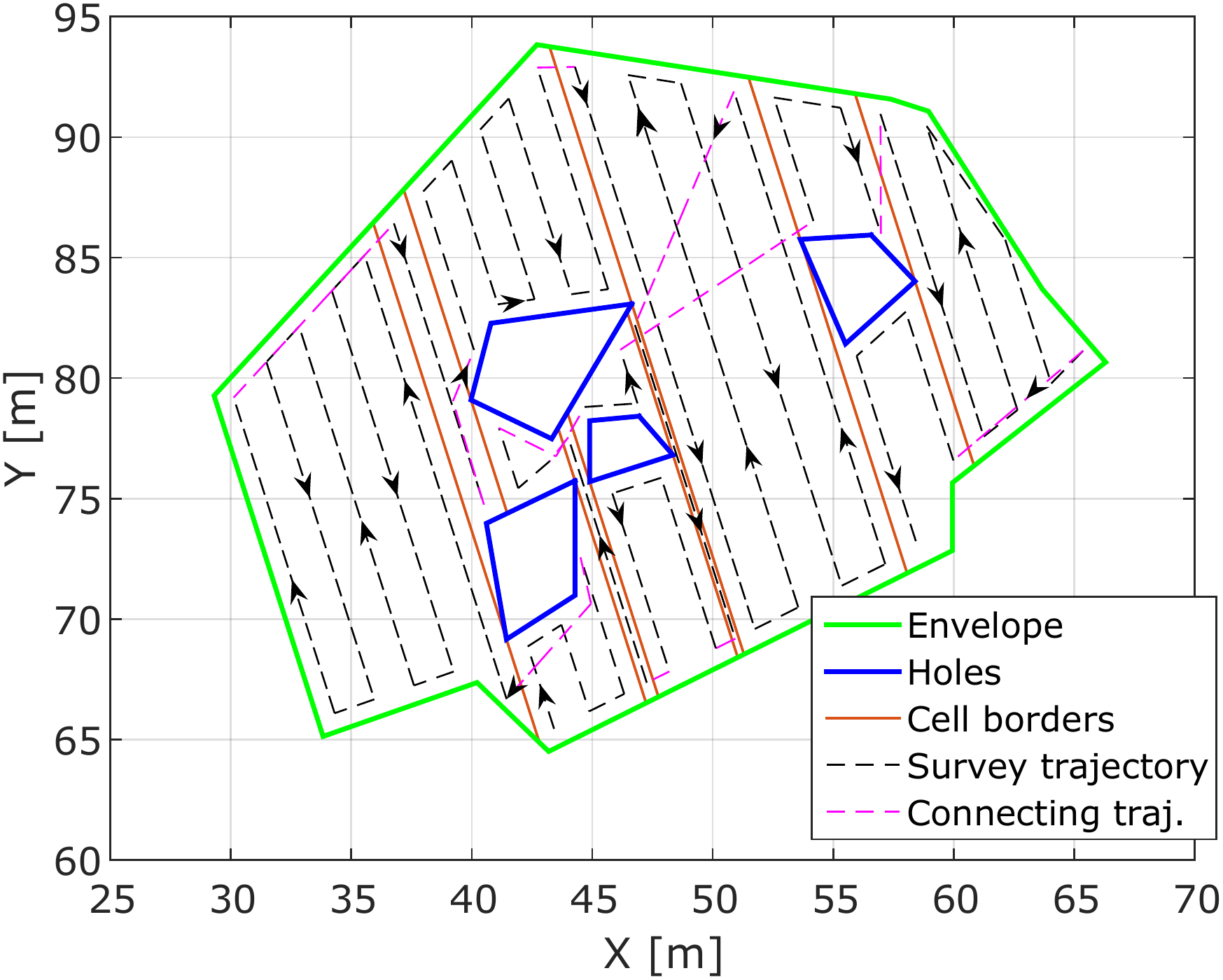}\label{fig:trajectory1}}
	\caption{The result of the Boustrophedon cell decomposition for the first ROI, complemented with a cell adjacency graph (a); the planned trajectory within the first ROI (b). ROI: region of interest [2-column figure]}
	\label{fig:traj-area1}
\end{figure*}

Both envelopes and their corresponding holes are passed to the algorithm responsible for the decomposition. Zone 1, whose area corresponded to approximately 750 m$^2$, was divided into 13 cells, as shown in Figure~\ref{fig:decomposition1}. The sweep line orientation was eventually chosen manually because the implementation had not been robust enough to handle an arbitrary case. In the trajectory planning, the first phase consists in selecting the initial point to start the survey; in our case, this step was performed manually. The cells were automatically ordered as follows: 
$3 \rightarrow 1 \rightarrow 2 \rightarrow 4 \rightarrow 10 \rightarrow 9 \rightarrow 8 \rightarrow 6 \rightarrow 5 \rightarrow 7 \rightarrow 11 \rightarrow 13 \rightarrow 12$.
The resulting trajectory is plotted in Figure~\ref{fig:trajectory1}, and it consists of two elements: a) 'zig-zag' shaped paths covering the whole area (except for the holes), and b) links connecting portions of the inspection routes to ensure that the holes are avoided safely. The theoretical length of the complete trajectory equals 448 m; note that this value applies only to holonomic robots without kinematic constraints. 

\begin{figure*}[!tbp]
	\centering
	\subfloat[]{\includegraphics[width=0.42\textwidth]{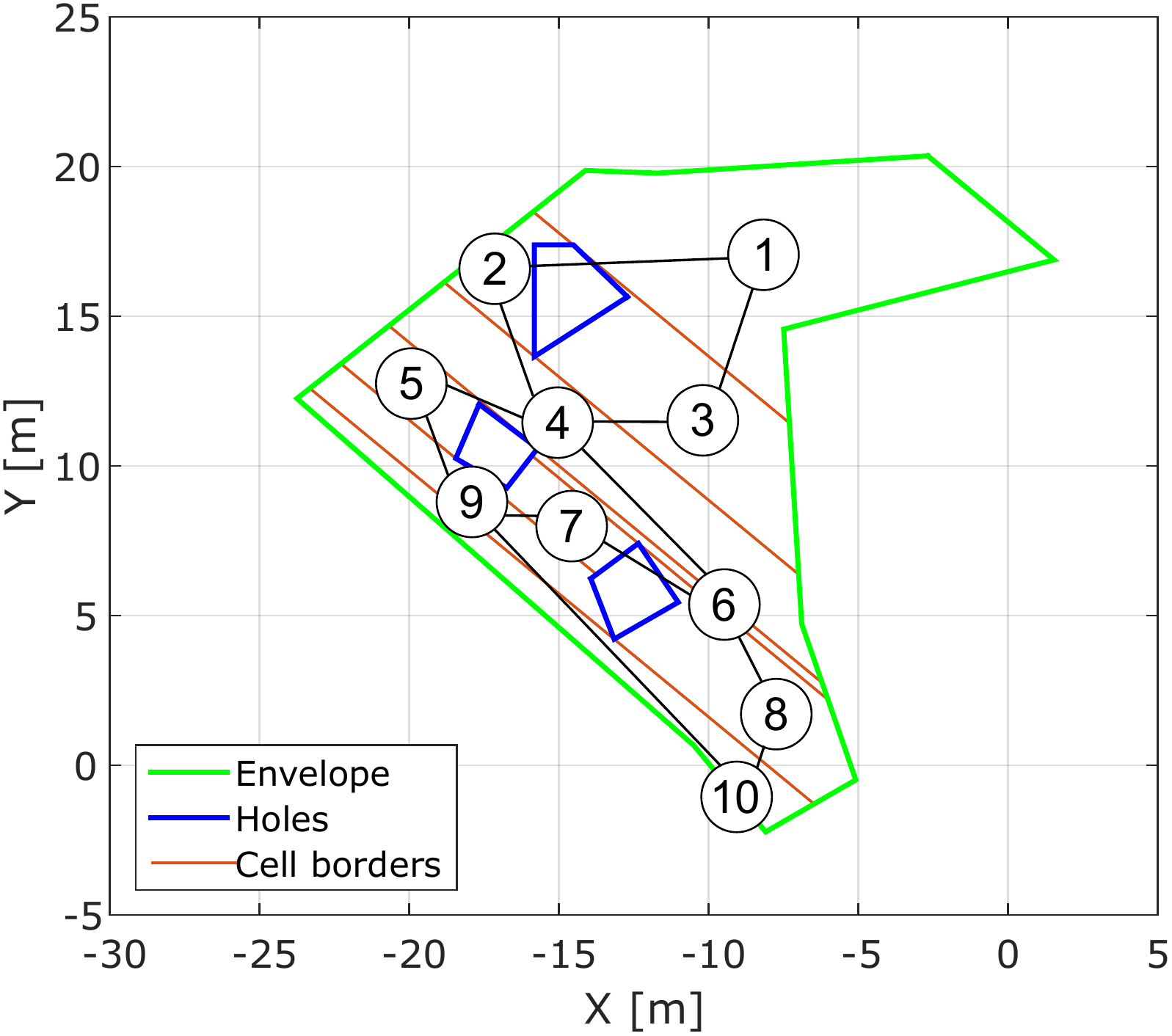}\label{fig:decomposition2}} \hspace{0.05\textwidth}
	\subfloat[]{\includegraphics[width=0.42\textwidth]{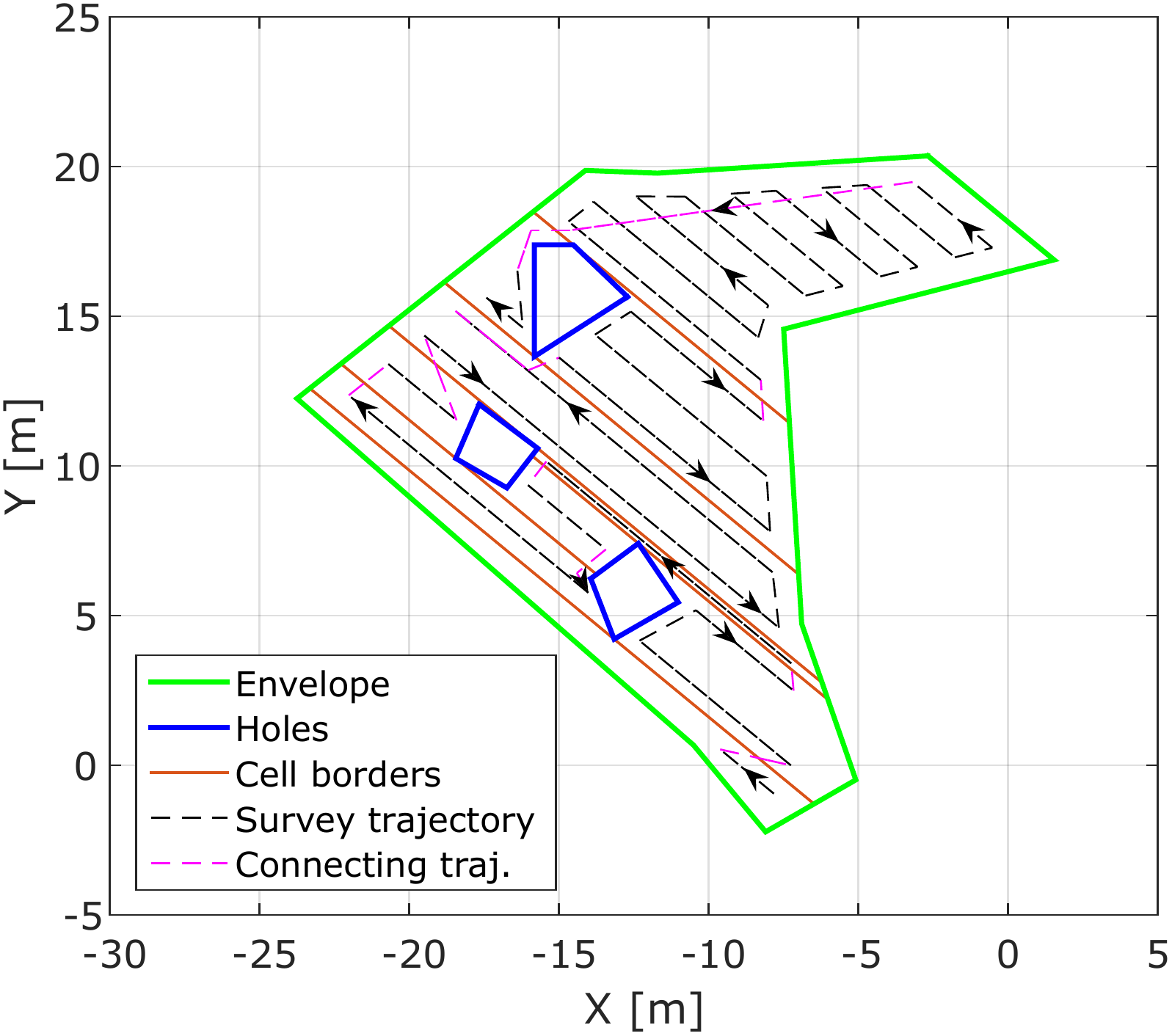}\label{fig:trajectory2}}
	\caption{The result of the Boustrophedon cell decomposition for the second ROI, complemented with a cell adjacency graph (a); the planned trajectory within the second ROI (b). ROI: region of interest [2-column figure]}
	\label{fig:traj-area2}
\end{figure*}

The same procedure was utilized also in zone 2, where the Boustrophedon algorithm split the area of 250 m$^2$ into 10 distinct partitions (Figure~\ref{fig:decomposition2}). The trajectory planning was initiated in the last cell (10) to yield the following survey sequence:
$10 \rightarrow 8 \rightarrow 6 \rightarrow 7 \rightarrow 9 \rightarrow 5 \rightarrow 4 \rightarrow 3 \rightarrow 1 \rightarrow 2$.
In zone 2, the complete trajectory has the length of 192~m and is shown in Figure~\ref{fig:trajectory2}.


\begin{figure}[!tbp]
	\centering
	\includegraphics[width=0.45\textwidth]{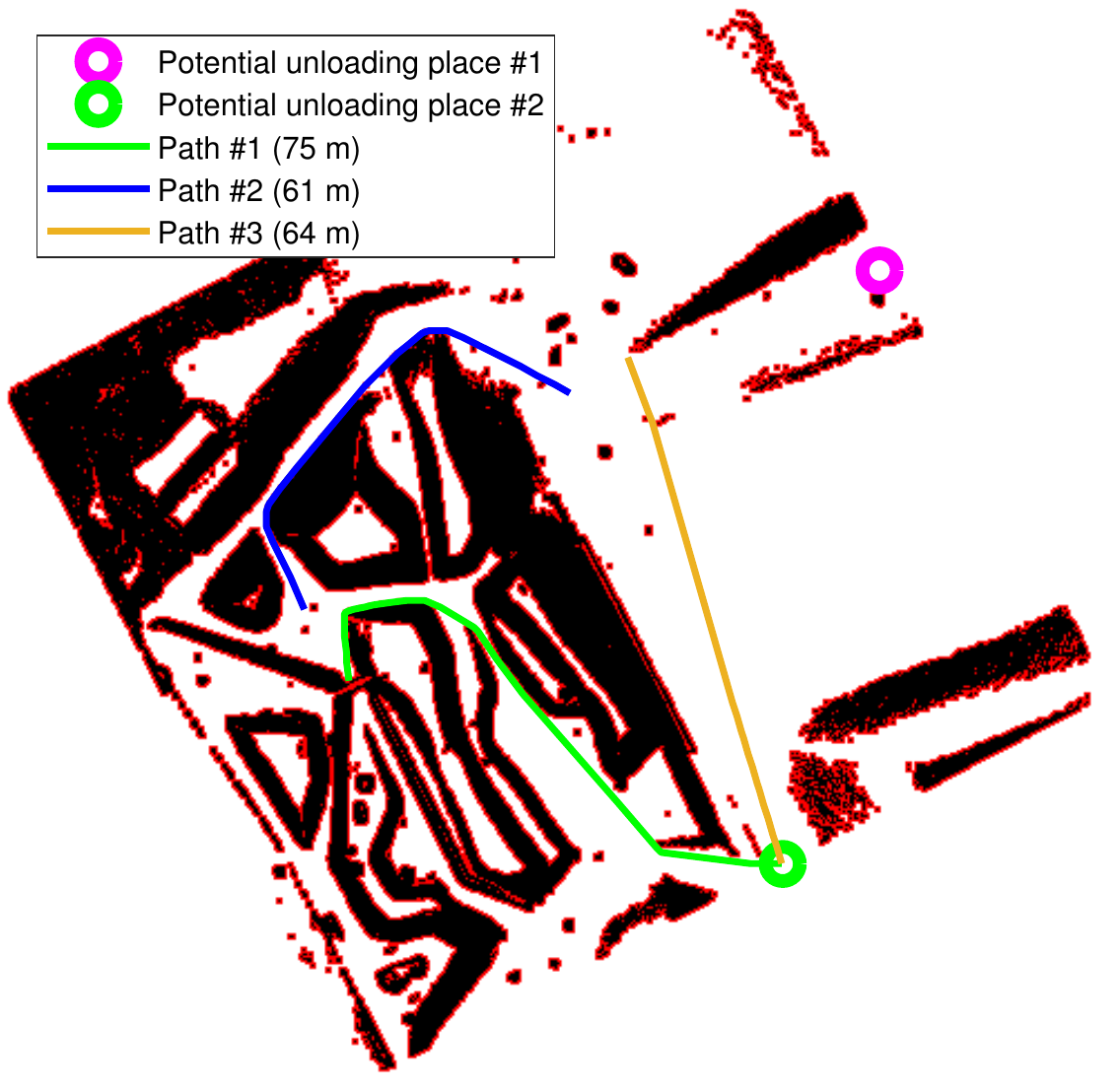}
	\caption{The A* planned trajectories between the unloading zone and the regions of interest. [1-column figure]}
	\label{fig:planned-paths-astar}
\end{figure}

The last path planning task interconnects the regions of interest and the zone most convenient for unloading the UGV. The operator selects suitable points to start the mission; we chose two spots (the green and pink circles in Figure~\ref{fig:planned-paths-astar}) on the road at the edge of the mapped area, where the contamination level is within the safe limits. The start and end points of the planned trajectory inside the ROI are fixed and cannot be altered during this phase. Using the A* algorithm implemented in the project presented in \cite{astar-implementation}, three paths were planned: from the unloading zone to a ROI, from this ROI to the next ROI, and from this last ROI back to the unloading zone. The sums of the path lengths are evaluated to select the lowest value. To reduce the UGV collision probability, all of the obstacles are expanded with an enclosing pixel (the red areas in Figure~\ref{fig:planned-paths-astar}). The resulting shortest sequence of the three paths is shown in a modified obstacle map (Figure~\ref{fig:planned-paths-astar}). The paths are 200 m long in total, and the UGV completed them in 6 minutes and 20 seconds (the speed varied from 0.4 to 0.6~m/s).

%% file: text/results-radiation-ugv.tex
\subsection{Radiation Data Processing and Source Localization}


After completing the path planning phases, we employed the UGV to acquire the radiation data in both zones. The robot's minimal turning radius was set to 0.6~m; thus, the shape of the actual trajectory differed from that of the pre-generated one. With the maximal forward speed equaling 0.6~m/s (0.4~m/s while turning), the measurement took 15 minutes and 10 seconds in zone 1, while the time relevant to zone 2 was 7 minutes and 35 seconds.

\begin{figure*}[!tbp]
	\centering
	\subfloat[]{\includegraphics[width=0.45\textwidth]{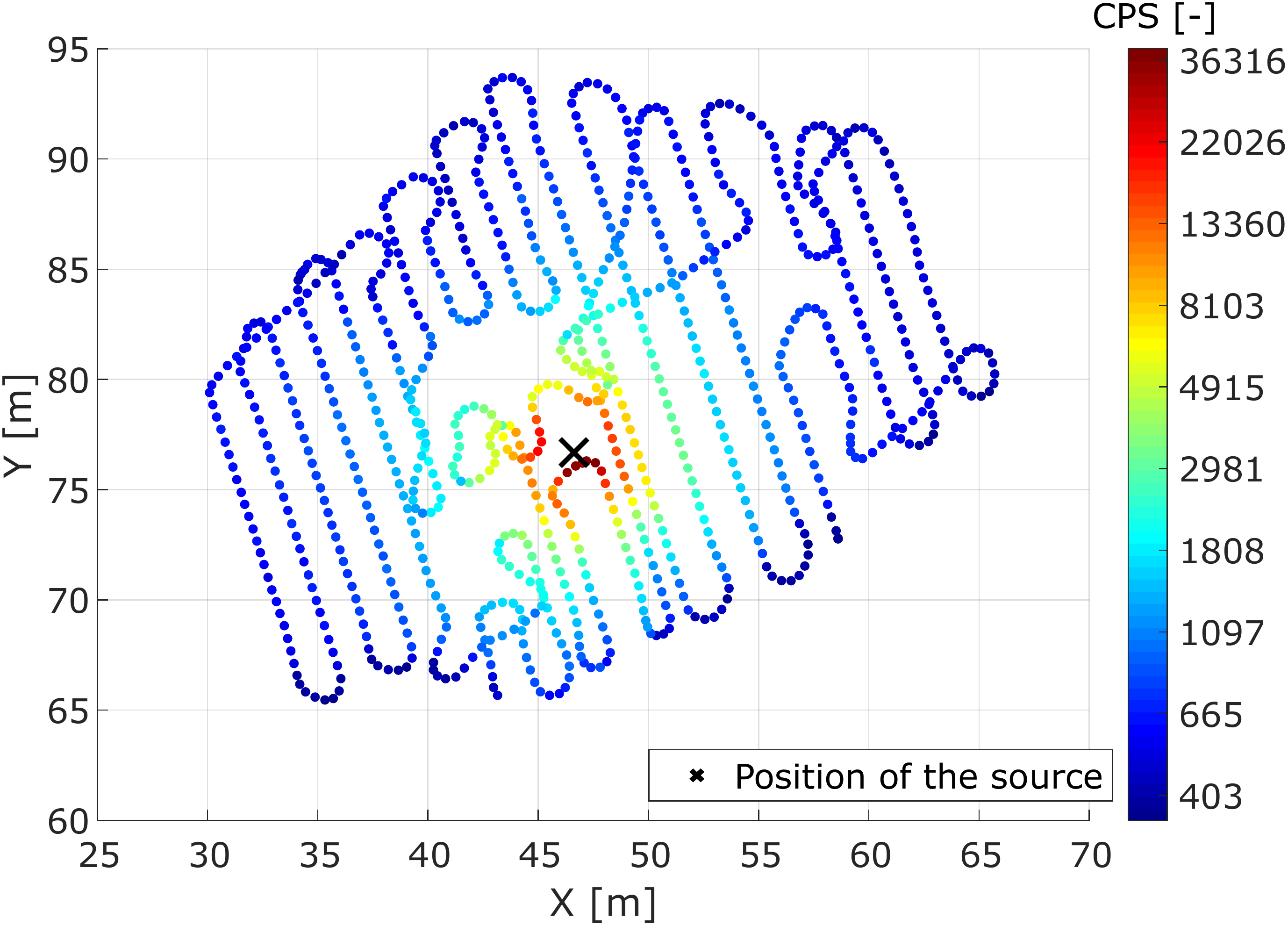}\label{fig:datapoints1}} \hspace{0.05\textwidth}
	\subfloat[]{\includegraphics[width=0.45\textwidth]{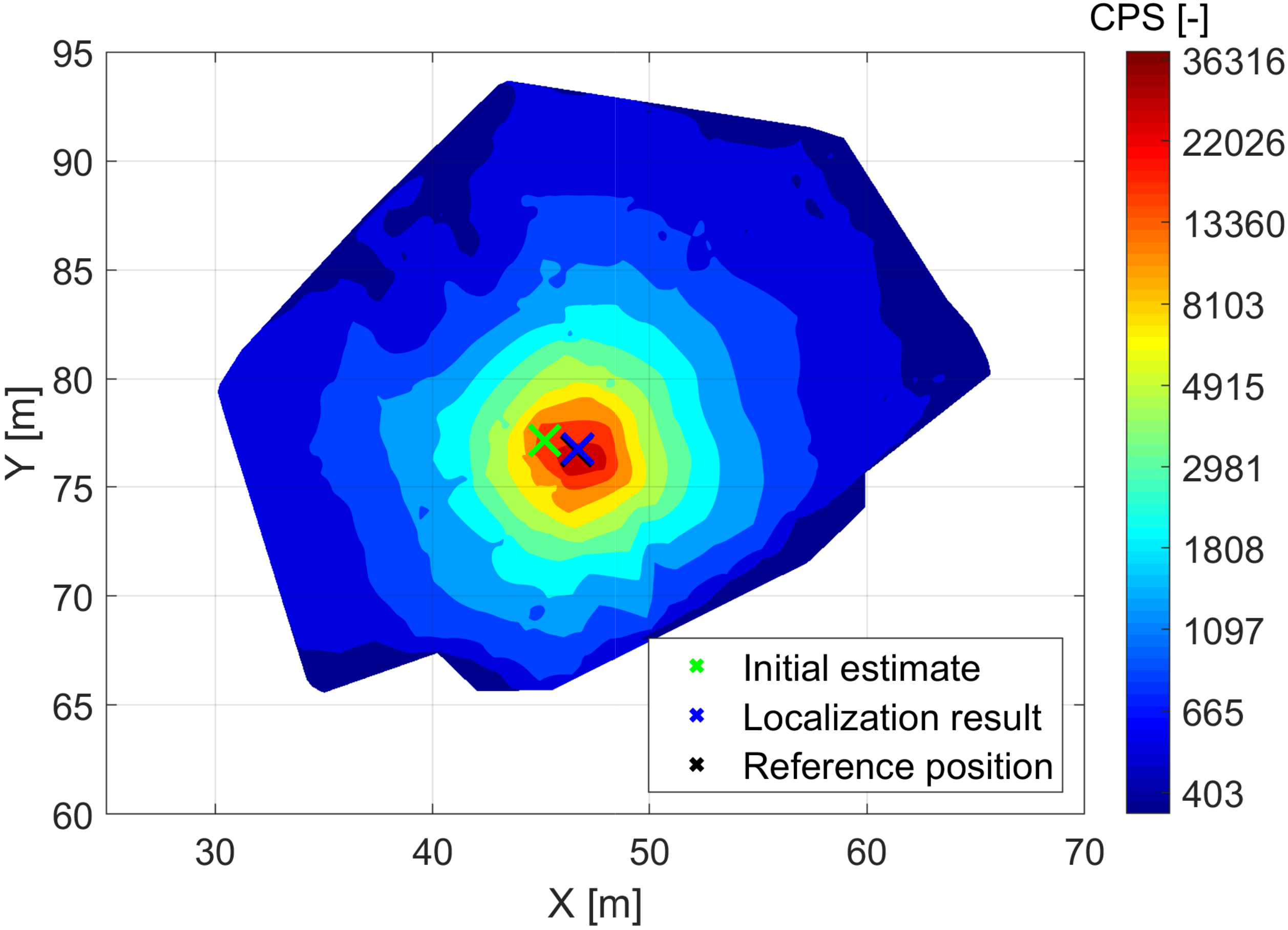}\label{fig:radmap1}}
	\caption{The individual data points measured along the planned trajectory; the points capture the total count in the first ROI (a). The interpolated radiation map highlighting the result of the source localization procedure (b). CPS: counts per second; ROI: region of interest. [2-column figure]}
	\label{fig:maps-area1}
\end{figure*}

The individual radiation data points comprise geographical coordinates and the total count of one-second measurement cycles delivered by the two detectors. The actual result is embodied in radiation spectra; at this stage, however, the spectra are relatively unimportant, and we therefore sum them into the TC. The measurement outcomes for zone 1 are presented in Figure~\ref{fig:datapoints1}; the relevant path was 495 m long. Subsequently, the data were interpolated and the background removed (the background and hotspot threshold exhibited the values of 1645 CPS and 2124 CPS, respectively). As the zone included merely a single source, the thresholding left a sole peak, and the parameter matrix was initialized smoothly. The initial and the improved estimates, the latter one based on the Gauss-Newton method, are indicated in Figure~\ref{fig:radmap1}. The localization error equaled 0.123 m (Table~\ref{tab:loc-result}).

\begin{figure*}[!tbp]
	\centering
	\subfloat[]{\includegraphics[width=0.4\textwidth]{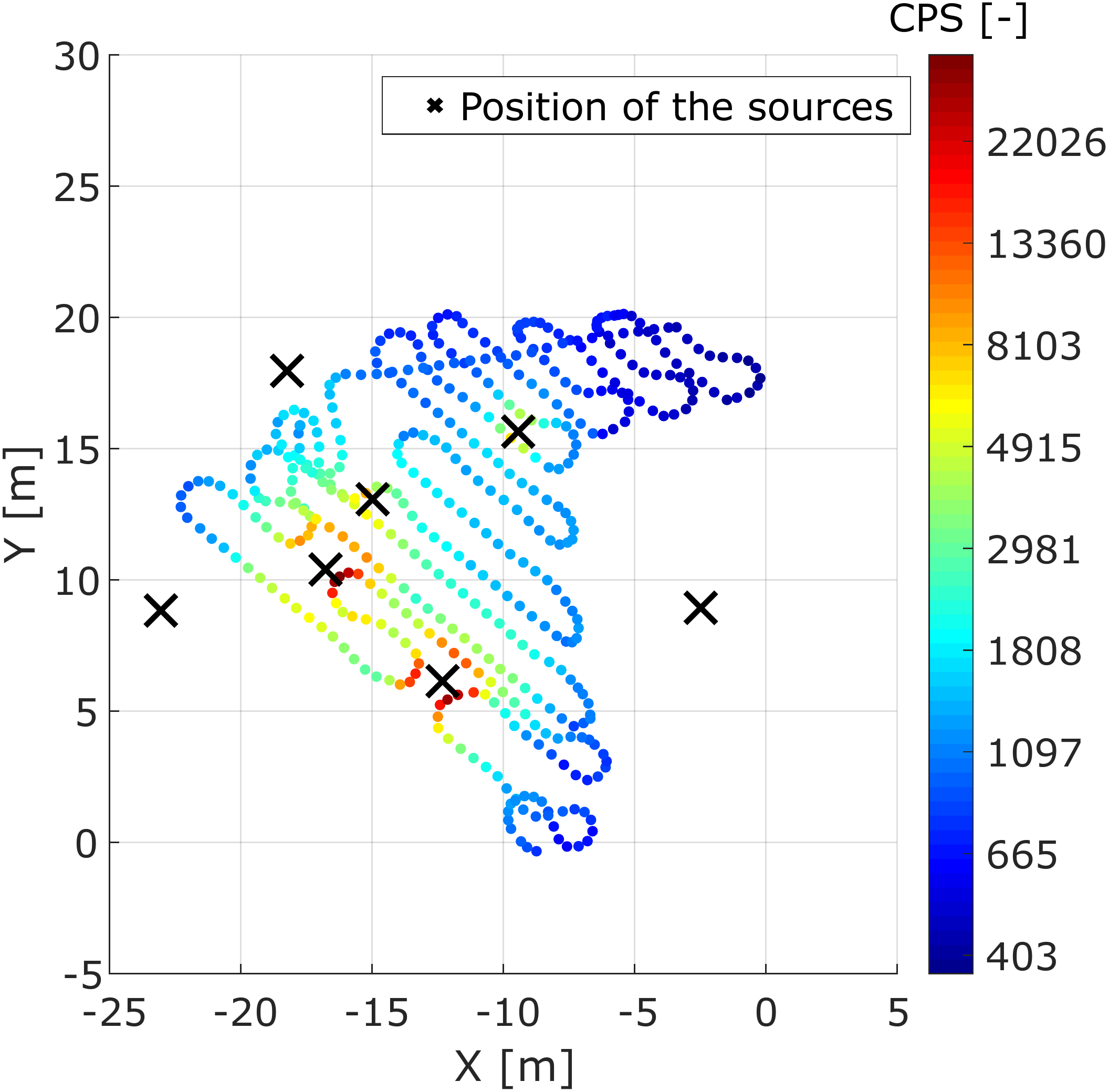}\label{fig:datapoints2}} \hspace{0.05\textwidth}
	\subfloat[]{\includegraphics[width=0.4\textwidth]{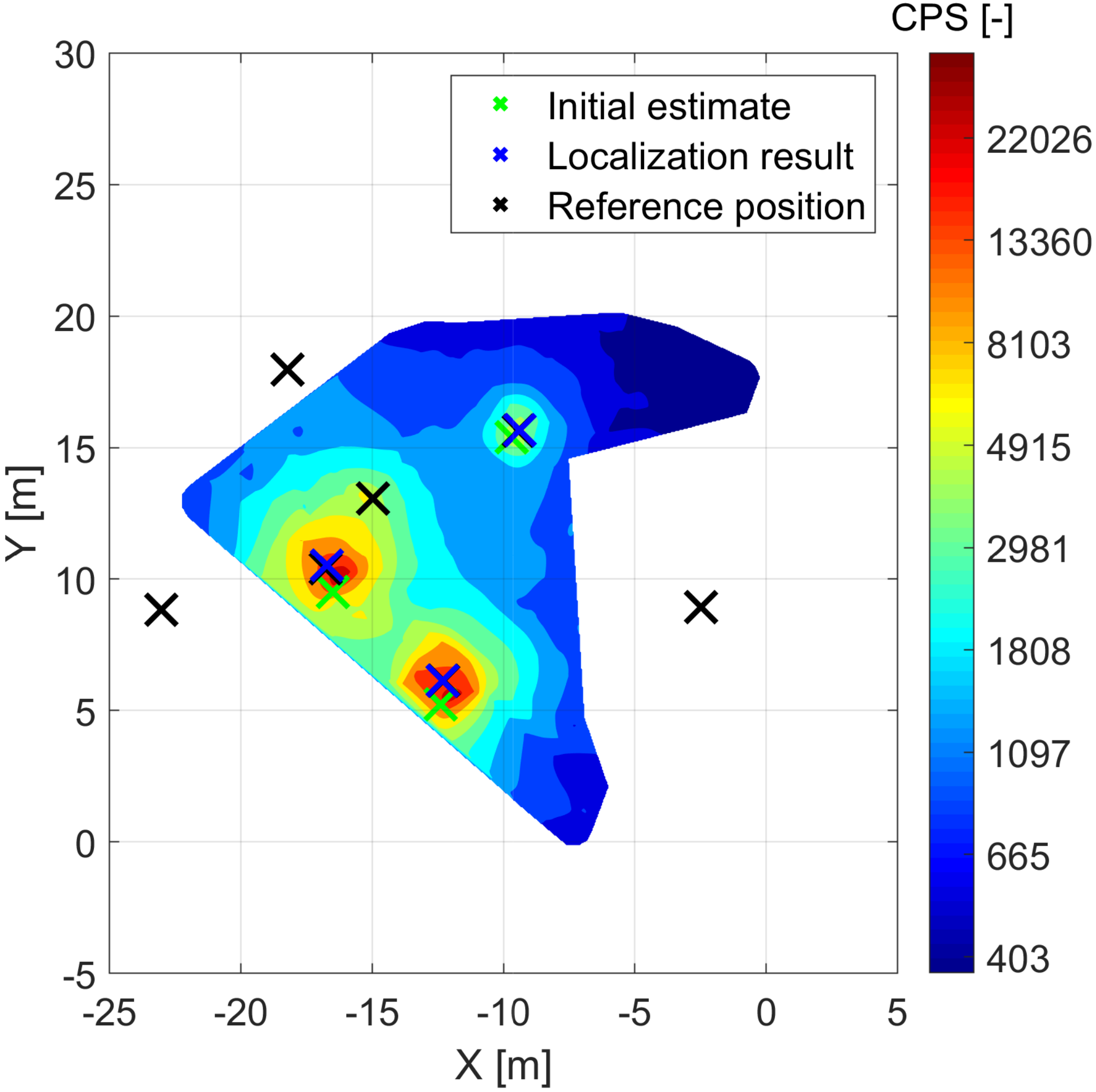}\label{fig:radmap2}}
	\caption{The individual data points measured along the planned trajectory; the points capture the total count in the second ROI (a). The interpolated radiation map highlighting the result of the source localization procedure (b). CPS: counts per second; ROI: region of interest. [2-column figure]}
	\label{fig:maps-area2}
\end{figure*}

The situation was more problematic in zone 2, where we placed 7 sources in total. The individual data points captured are shown in Figure~\ref{fig:datapoints2}; the length of the actual trajectory corresponded to 221 m. Three sources, namely, radionuclides s1, s4, and s7,  were located outside the surveyed area. In terms of further description, s7 was positioned on a steep slope and thus remained inaccessible to the UGV; s1 and s4 then lay in a free space and were omitted by the planning algorithm because the primary ROI borders had not been broad enough (these radionuclides ranked among the weakest ones and thus did not leave a noticeable signature on the aerial data). The threshold levels for the background and the hotspots equaled 2707 CPS and 4684 CPS, respectively; note that the values are greater than those relating to zone 1, as the major portion of the data points lay in the vicinity of the sources. The adaptive thresholding yielded three distinct peaks, correspondingly to sources s3, s5, and s6; the last peak (s2, weak caesium 137) was overshadowed by the strong Cs-137 in its close proximity. Consequently, only 3 out of the 7 sources were localized successfully, as is obvious from the detailed results in Table~\ref{tab:loc-result}. The average localization error in both of the zones (considering only sources whose parameters were found) equals approximately 0.10 cm RMS.

To quantify the benefits of employing the UGV in more detailed measurement, the localization algorithm was also applied to the aerial data. The thresholding result remained the same as in the ROI selection (Figure~\ref{fig:areaselection}), yielding two source estimates. Clearly, the localization error in zone 2 cannot be computed, because the 7 sources present there appear as a single one in the aerial radiation map. However, we can compare the results obtained within zone 1, where the UAS localization error equals 2.82 m after application of the Gauss-Newton algorithm (Table~\ref{tab:loc-result}).

\begin{table*}[!tbp]
	\caption{The source localization results: N-q stands for non-quantifiable, as the UAS localization error for zone 2 cannot be expressed in the usual manner. UGV: unmanned ground vehicle; UAS: unmanned aircraft system; ROI: region of interest.}
	\begin{center}
		{\def\arraystretch{1.3}
			\begin{tabular}{llccccc} 
				\hline
				\bf{Source} & \bf{Zone} & \bf{Error} & \bf{Error} & \bf{Isotope} & \bf{Activity} & \bf{Comment}\\ 
				& & \bf{UGV [m]} & \bf{UAS [m]} & & \bf{[MBq]} & \\
				\hline
				s1 & 2 & -- &  & Co-60 & 2.85 & Outside the ROI \\
				s2 & 2 & -- &  & Cs-137 & 7.53 & -- \\
				s3 & 2 & 0.067 &  & Co-60 & 2.95 & -- \\
				s4 & 2 & -- & N-q & Cs-137 & 7.53 & Outside the ROI \\
				s5 & 2 & 0.138 &  & Cs-137 & 79.82 & -- \\
				s6 & 2 & 0.018 &  & Co-60 & 24.56 & -- \\ 
				s7 & 2 & -- &  & Co-60 & 24.76 & Inaccessible to the UGV \\
				\hline
				s8 & 1 & 0.123 & 2.82 &  Co-60 & 123.78 & -- \\
				\hline
		\end{tabular}}
	\end{center}
	\label{tab:loc-result}
\end{table*}

%% file: text/discussion.tex

\begin{figure}[!tbp]
	\centering
	\includegraphics[width=0.4\textwidth]{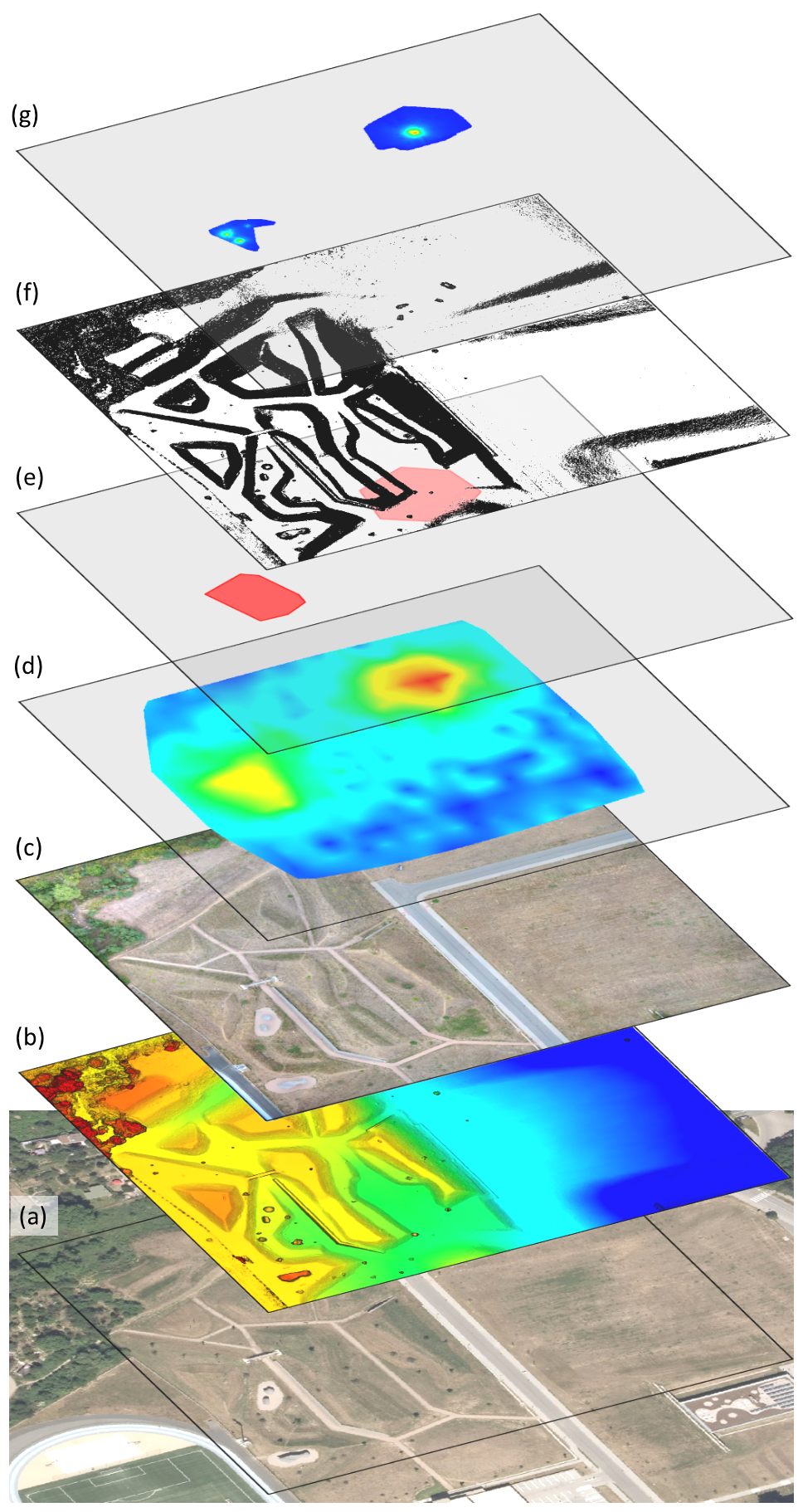}
	\caption{The most significant map layers assembled during the mapping and processing. The layers are arranged according to their times of origin, from the bottom upwards: the primary orthophoto (a); the UAS-based, shaded DEM (b); the UAS-made orthophoto (c); the UAS-delivered radiation map (d); the detected regions of interest (e); the DEM-based UGV traversability map (f); and the UGV-made radiation map (g). UAS: unmanned aircraft system; DEM: digital elevation model; UGV: unmanned ground vehicle. [1-column figure]}
	\label{fig:map-layers}
\end{figure}
\begin{figure*}[!tbp]
	\centering
	\includegraphics[width=0.95\textwidth]{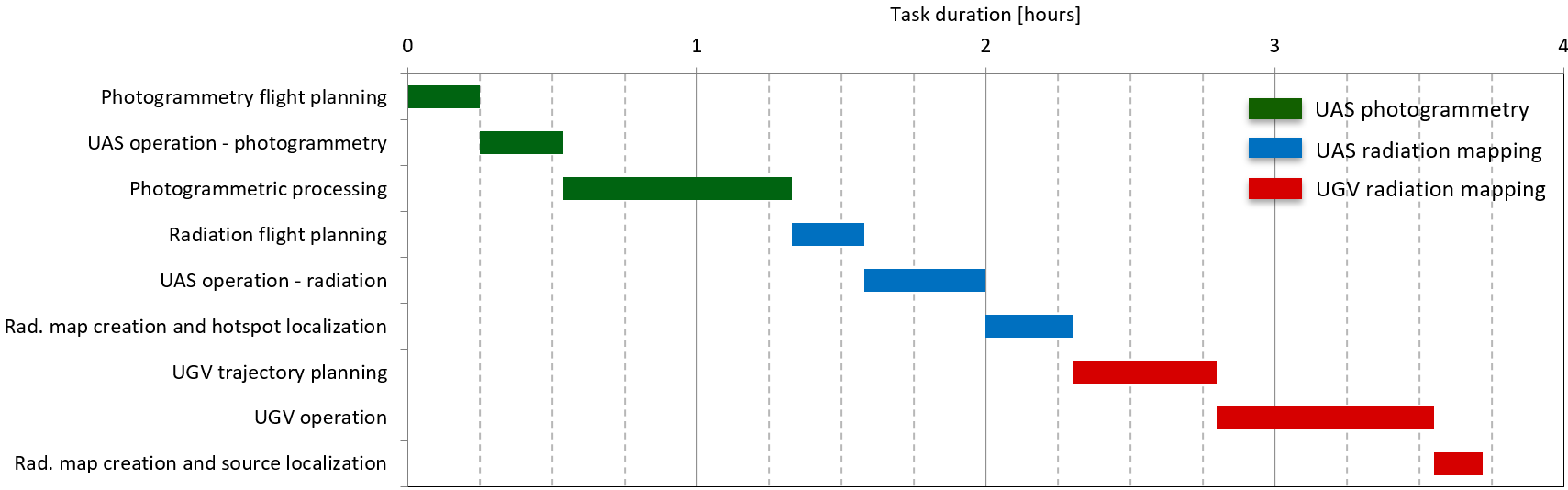}
	\caption{The approximate times of the individual tasks during the experiment. UAS: unmanned aircraft system; UGV: unmanned ground vehicle. [2-column figure]}
	\label{fig:gantt}
\end{figure*}

Within the presented experiment, we introduced and successfully tested a multi-robot radiation mapping method consisting of numerous steps (the essential mapping outputs are summarized in Figure~\ref{fig:map-layers}). The entire operation lasted 24 hours; this continuous time interval comprised not only the necessary tasks, namely, the data gathering and processing, but also the site preparation and cleanup, safety-related steps, and activities not directly associated with the experiment. The time intensity of the operations relevant to the mapping and processing are summarized within the Gantt chart in Figure~\ref{fig:gantt}. The individual items include the time spent on the automatic tasks (data processing, robot operation), operator interventions, and robot preparation and manipulation. The most time-intensive stages are the UGV operation and the photogrammetric processing. Theoretically, an ideal mission takes less than 4~hours; in reality, however, we had to face numerous minor issues that eventually prolonged the whole process, mainly as the mission marked the first time the systems were deployed together.

The UAS photogrammetry survey involved the use of our custom-built multi-sensor system and was carried out repeatedly; during the procedures, we thoroughly evaluated the achievable accuracy. Despite this, the attained values did not meet our expectations: As described in section~\ref{ssec:res-aerial-photo}, the RMS object error determined by using the six test points lay within the order of decimeters in the horizontal coordinates and rose slightly above a meter in the vertical one. According to our investigation and data analysis, all systems performed properly (including the RTK correction transmission); however, the signals on the GNSS receiver's antennas were rather weak, caused insufficient conditions for the carrier phase tracking during the entire flight. This problem resulted in RTK-fixed solution outages and made the INS exclude the GNSS data from the position and orientation estimates for a moment; the issue affected the beginning of the third survey line (Figure~\ref{fig:rtk-quality}). Since the multi-sensor system was combined with the BRUS UAS for the first time, the problem may have been generated by interferences from the UAS's electronic systems. Fortunately, the lower georeferencing quality did not manifest itself in the subsequent phases, and we still consider direct georeferencing crucial with respect to radiation-related missions.


Aerial radiation mapping proved to be a very effective tool for hotspot localization. The innovative approach involving flying at a constant AGL height regardless of the surface character allowed us to collect homogeneous data. Outside this scenario, the distance separating the ground and the detector would vary between 15 and 30~meters in a flight 15~meters above the highest location (at a fixed MSL altitude); such a diversity would certainly mean inconsistent data, and lower-positioned hotspots would be localized inaccurately or not at all. However, the DEM-based trajectory adjustment algorithm needs to be improved in several respects, of which the two most prominent ones are as follows: First, the method does not deliver the desired distance from the surface at high gradient locations, as it modifies the vertical coordinates of the waypoints only (as shown in Figure~\ref{fig:rad-traj-profile}); second, the algorithm should consider the UAS dynamics because some UAS control units fully ensure the horizontal speed while providing merely limited vertical speed, thus causing inaccurate waypoint following in steep parts. Based on the UAS-collected data, two regions of interests were automatically defined; this action reduced the original area to less than 10~\%, with only 1,500~m$^2$ left for the terrestrial mapping. 


Using only a DEM to select regions inaccessible to UGVs within the mapped area cannot yield 100\% reliable outputs. Deformable objects (such as blades of grass and light bushes) satisfy the definition of an obstacle in terms of the height and gradient, despite being effectively bypassable by a UGV; moreover, such objects cannot be separated from non-deformable obstacles, because in a DEM they are represented by the same data. Although the decision-making can utilize an orthophoto (automatically or manually), this approach produces only probable bypassability, which does not constitute a reliable option. Other issues arise from the actual capabilities of a DEM, one of the main limitations being that some free spaces, such as those under bridges, are not covered by the model. If no safe path for a UGV is found, we can follow that with the highest passability rate, albeit exclusively in the operator-assisted mode.

For many reasons, autonomous UGVs designed to participate in diverse missions require real-time obstacle avoidance. In view of this parameter, the DEM-based method is markedly limited in that the model captures only the situation existing at the time the source data were acquired, and thus the technique's applicability remains solely within the representation of fixed obstacles. including hills and mountains. Another set of incorrectly evaluated obstacles comprises objects undetected due to inaccuracies stemming from either the low resolution (e.g., in thin items such as columns and fences) provided by a DEM or poor object texture (e.g., the light being outside the usable sensor range). Such collisions can be prevented by a real-time obstacle avoidance system installed on board the UGV. In the context of our mission, it is important to emphasize that objects inside the mapped area are very likely to occur or change unexpectedly, and this type of system would significantly increase the efficiency of the entire reconnaissance process. 


Considering the requirement for short overall mission time, an adequate DEM resolution has to be selected. For this purpose, we tested higher resolutions (up to 16x) to determine that while they did not improve the resulting obstacle map, the processing time and noise level increased significantly. Based on the attempts to fine-tune the whole task, we may conclude that computing a DEM with resolutions above 5 cm/pix does not bring any substantial benefits. Regarding the UAS path planning for the second flight, which also embodies the second task employing a DEM, it is possible to point out the lower sensitivity to DEM accuracy, an aspect that enables us to achieve satisfactory results even at values below 5 cm/pixel.

\begin{figure}[!tbp]
	\centering
	\includegraphics[width=0.48\textwidth]{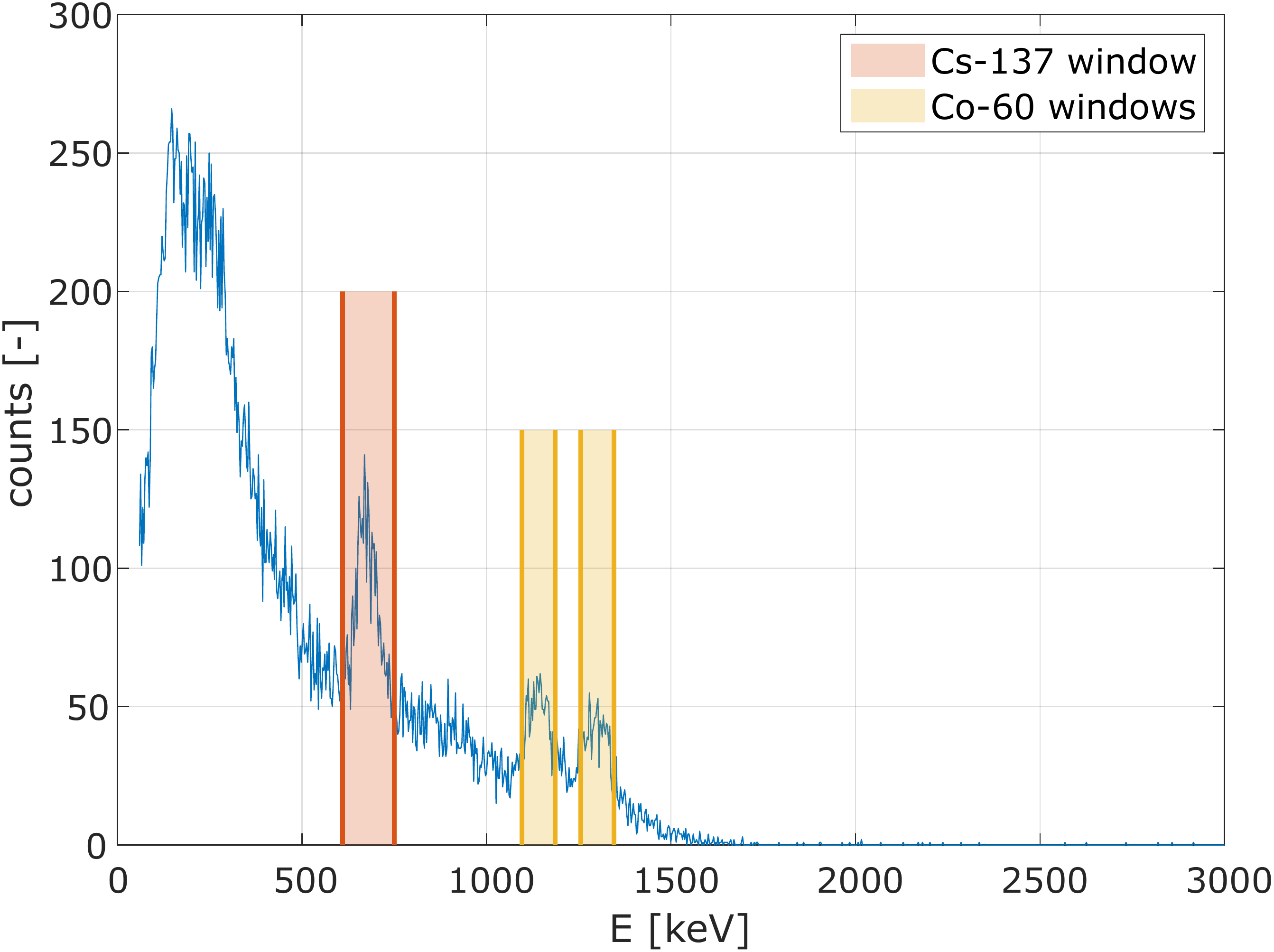}
	\caption{A radiation spectrum measured by the UGV's on-board detector; the graph indicates the energy windows of the applied radionuclides. UGV: unmanned ground vehicle [1-column figure]}
	\label{fig:spectral-windows}
\end{figure}
\begin{figure*}[!tbp]
	\centering
	\subfloat[]{\includegraphics[width=0.4\textwidth]{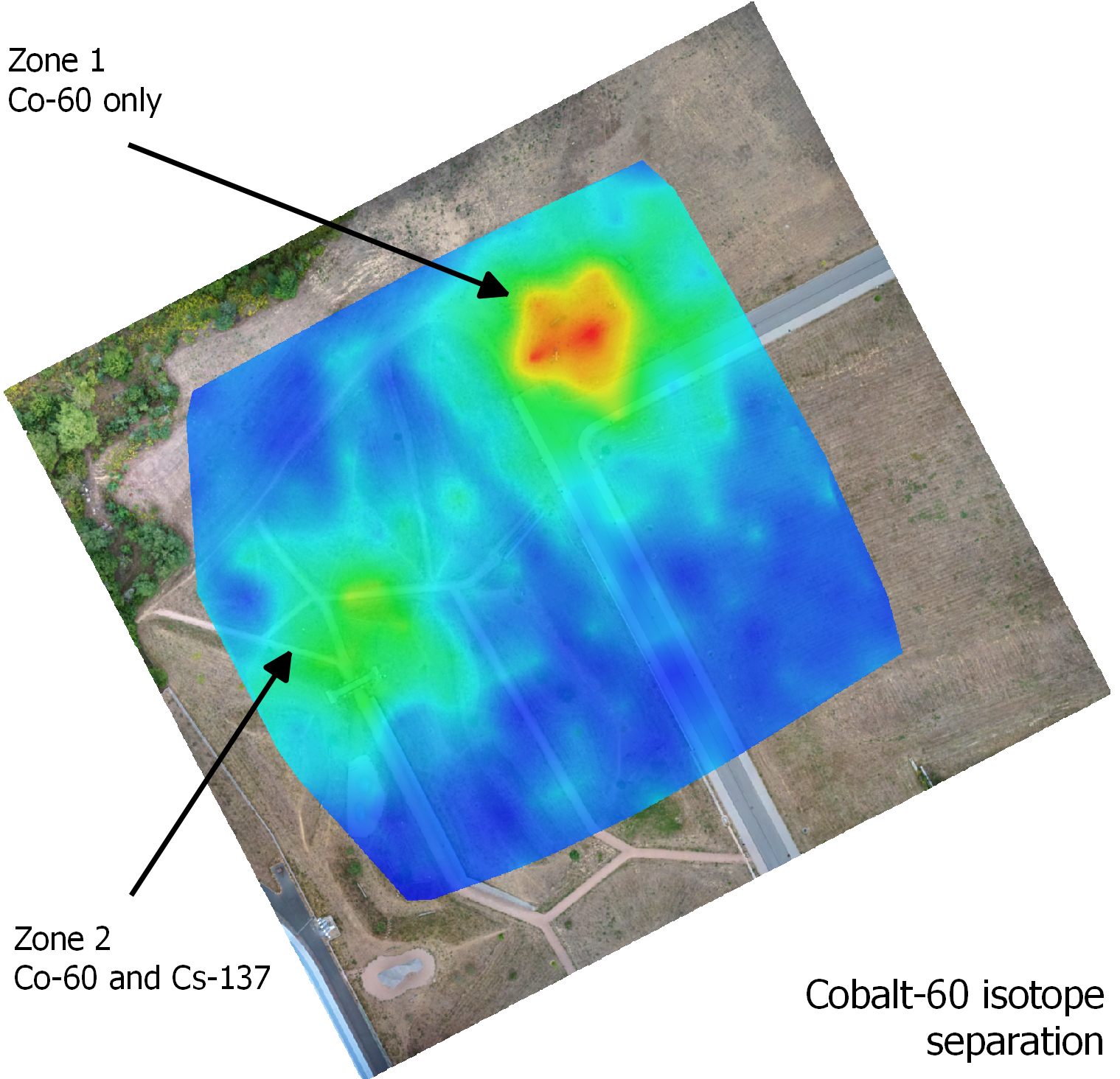}\label{fig:map-co60}} \hspace{0.05\textwidth}
	\subfloat[]{\includegraphics[width=0.4\textwidth]{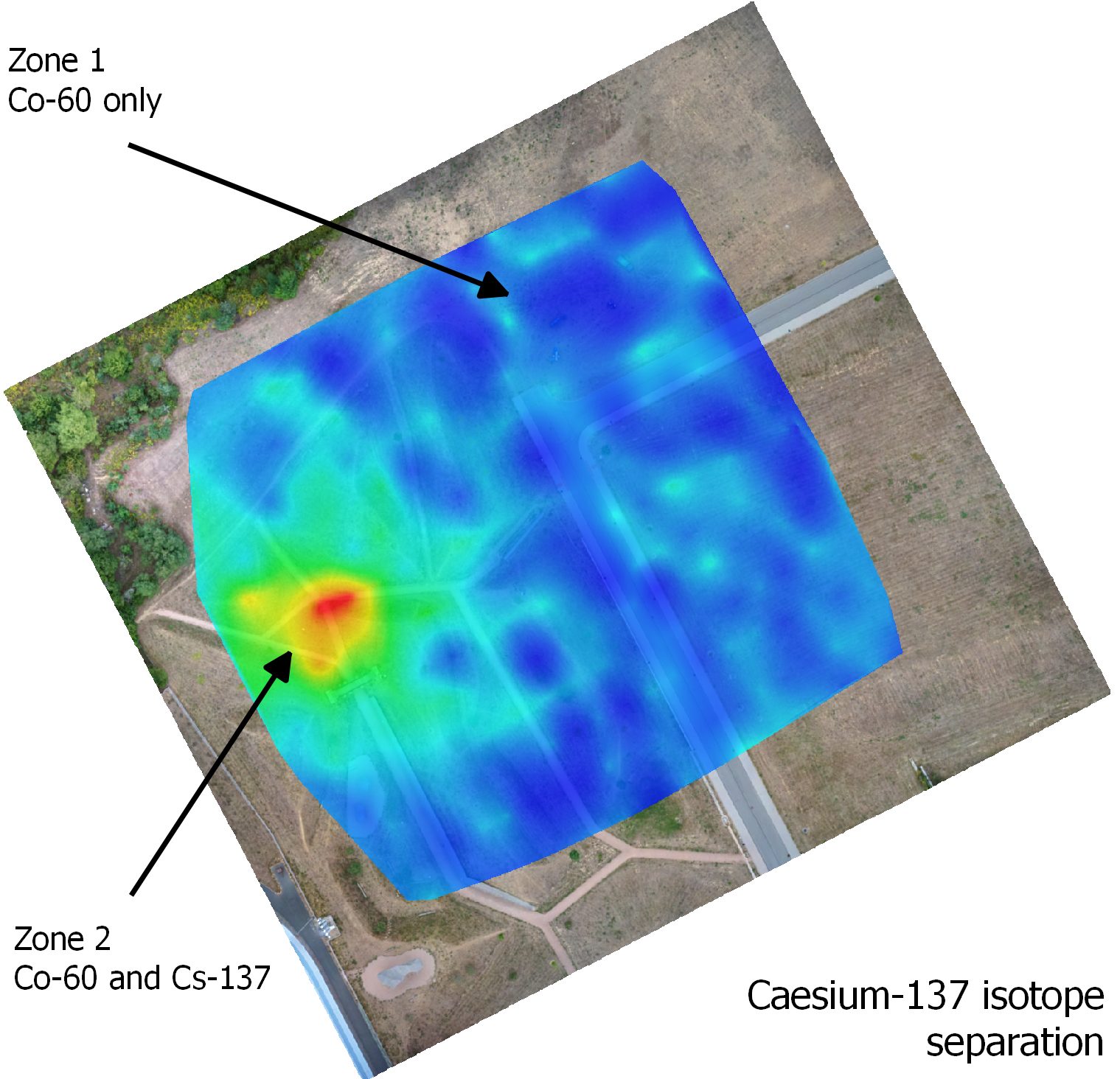}\label{fig:map-cs137}}
	\caption{The maps with separated radiation intensities for the cobalt 60 (a) and the caesium 137 (b). [2-column figure]}
	\label{fig:isotope-sep}
\end{figure*}

Although all of the algorithms worked only with either the dose rate or the raw total count during the entire source localization procedure, the use of spectrometric detectors in the experiment enabled further processing of the acquired data. Figure~\ref{fig:spectral-windows} shows the sample spectrum integrated over the period of 10 s along the trajectory between the distinct radionuclides. The graph visualizes three photopeaks, which essentially embody the 'fingerprints' of the incident photons, namely, the photons' energy that is unique for each radioactive element. The net counts in the energy windows are proportional to the contribution of the relevant isotopes towards the overall measured intensity; note that the width of a window depends on the energy resolution of the detector, usually expressed by full width at half maximum (FWHM). To compute the net value, it is necessary to subtract the average background level and also the counts yielded through the impact of the higher-energy photons (in our experiment, the cobalt 60 affects the caesium 137 window via Compton scattering). 
The photons' influence can be quantified via the stripping coefficient, acquired from those measurements where the cobalt is present while the caesium is not; such a scenario was performed in area 1. As an example of the spectral isotope separation, maps relating to the two radionuclides are presented in Figure~\ref{fig:isotope-sep}; the images clearly show that Cs-137 sources were located in only one of the hotspots. Note that this result was not intended to be part of the experiment and was supplied additionally.

The experiment indicated that both aerial and terrestrial radiation mapping procedures involve specific drawbacks, as follows: The information density of the data acquired by the UAS suffices for localizing a single isolated source (s8), providing a result that could be accurate enough in practice; however, given the coarse aerial radiation map, it is virtually impossible to distinguish between a strong source, multiple radionuclides, and non-point areal contamination, as demonstrated in zone 2. By contrast, the UGV-based measurements characterized the actual radiological situation in a better manner, yet still not precisely enough; the reason lay in that the hypothetical 'center of radiation' (an analogy to a center of mass) in zone 2 was shifted towards the east by the relatively strong source s7, causing the weak radionuclides to be left outside the region of interest. In the future, this problem could be easily eliminated by enlarging the ROI prior to executing the UGV path planning phase. Obviously, a terrestrial robot is incapable of localizing sources positioned in a space classified as an obstacle (s7), and this deficiency, in general terms, requires further application of a UAS to explore such portions of the ROI that remain inaccessible to other robots. Using a UAS in this scenario nevertheless also invokes the question of safety, as the aerial vehicle needs to be brought closer to the terrain. Regarding the ground inspection, another disadvantage consisted in that the procedure failed to separate the overshadowed weak source (s2); however, performing a measurement detailed enough to localize this source would probably be more time-intensive than repeating the entire survey after other sources had been removed from the area. Yet, despite the difficulties, the UGV has proved to be a significant component of the system because it provides a more accurate overview of the radiological situation within the hotspots.


Contrary to our previously published research, we did not attempt to employ information driven localization, i.e., real-time UGV trajectory adaptation according to continuously acquired data. Instead, the goal was to compile a radiation map as precise as possible to cover also sources that are generally difficult to detect. With some prior information, such as that only one radionuclide is sought, we could utilize the partial directional sensitivity provided by the two-detector system to head towards the radiation source immediately after its presence has been indicated. To achieve this purpose, it would be necessary to assure obstacle avoidance, fusing the source direction estimation with the obstacle map via exploiting the potential field algorithm if feasible. 


If we compare the results achieved within our research with those presented in articles focused on the same or similar topics, namely, \cite{christie_radiation_2017} and \cite{peterson_experiments_2019}, several key differences stand out. The former paper offers semantic classification of the surface type, providing useful information for navigating a terrestrial robot. Importantly, the applied UGV is equipped with an obstacle avoidance system that can be especially helpful in environments with dynamically occurring obstacles. By contrast, however, the authors do not utilize any sophisticated aerial data processing method to recognize multiple points of interest (POI) on the ground. The latter article introduces algorithms that exploit the measured spectra in selecting the POIs to perform information-driven localization of a single source;  advantageously, the authors also compare multiple methods applicable for the given purpose. Considering the outcomes of these two research projects, we can stress that the novelty and benefit of our concept consist in other aspects, defined as follows: the terrain- following capability and directly georeferenced photogrammetry delivered by the UAS; automatic selection of the ROIs; and higher-accuracy, isotope-independent localization of multiple sources, performed with a UGV whose navigation and trajectory planning are fully autonomous (except for the necessity to validate the obstacle map by an operator). Finally, it is worth mentioning in the given context that the whole experiment was completed in a single day.

%% file: text/conclusion.tex
Using relevant experiments, this paper verified a concept of exploiting aerial and terrestrial robotic platforms to localize uncontrolled radiation sources in a previously unknown outdoor area. After completing the three phases of the designed survey process, we found four of the eight radionuclides (or three of the four significant ones); the achieved accuracy was below 0.2~m, a value sufficient to support subsequent steps such as the removal of the sources from the area. The experiment was implemented in 24 hours, including the elimination of various technical issues. Theoretically, the area of 20,000~m$^2$ can be explored in only 4 hours, assuming conditions similar to those presented herein. To complete the entire task smoothly, however, the system would require further modifications. In this context, there remain major constraints as related to the weather, environment, radiological situation, and other relevant aspects: The systems must operate in adequate flight conditions, and satisfactory GNSS reception as well as the accessibility of a significant part of the area to the UGV need to be ensured. Moreover, the radiation intensities have to be well detectable yet not hazardous for the electronics. At this point, it is also vital to emphasize that the cooperation between aerial and terrestrial robots should be promoted because the same results cannot be achieved with one of the variants only; a UAS, for example, is incapable of ensuring either conclusive localization accuracy or differentiation between sources concentrated within an area of hundreds of square meters. By contrast, a UGV, if operated without the aerial data, has to explore the inspected area globally, and the lack of an obstacle map causes serious navigation problems, especially where the applied vehicle is not equipped with an evasion module. Our future research will be directed towards employing information-driven localization and fitting the UGV with an obstacle avoidance system.